\documentclass[10pt,a4paper,oneside,notitlepage,fleqn,reqno,english,table]{article}

\usepackage[hidelinks]{hyperref} 

\usepackage{multicol} 

\usepackage{longtable} 
\usepackage{multirow} 
\usepackage{rotating} 
\usepackage[cp1252]{inputenc} 
\usepackage{lscape} 
\usepackage{url}    
\usepackage[section]{placeins}  
\usepackage{afterpage} 
\usepackage{capt-of}        
\usepackage{dirtytalk} 

\usepackage[T1]{fontenc}
\usepackage{tabularx, booktabs}   
\usepackage{makecell} 
\usepackage{array} 
\usepackage{xcolor} 
\usepackage{xfrac} 

\usepackage{tikz}  
\usetikzlibrary{shapes,arrows,arrows.meta,calc,shadows,fit,intersections,datavisualization.formats.functions}
\tikzstyle{block}        = [draw, fill=blue!30, rectangle, text centered, minimum height=3em, minimum width=6em] 
\tikzstyle{blockred}     = [draw, fill=red!30, rectangle, text centered, minimum height=3em, minimum width=6em] 
\tikzstyle{blockyellow}  = [draw, fill=yellow!30, rectangle, text centered, minimum height=3em, minimum width=6em] 
\tikzstyle{blockyellow1} = [draw, fill=yellow!15, rectangle, text centered, minimum height=3em, minimum width=6em] 
\tikzstyle{blockyellow2} = [draw, fill=yellow!30, rectangle, text centered, minimum height=3em, minimum width=6em] 
\tikzstyle{blockyellow3} = [draw, fill=yellow!45, rectangle, text centered, minimum height=3em, minimum width=6em] 
\tikzstyle{blockyellow4} = [draw, fill=yellow!60, rectangle, text centered, minimum height=3em, minimum width=6em] 
\tikzstyle{blockgreen}   = [draw, fill=green!30, rectangle, text centered, minimum height=3em, minimum width=6em] 
\tikzstyle{blockgreen1}  = [draw, fill=green!15, rectangle, text centered, minimum height=3em, minimum width=6em] 
\tikzstyle{blockgreen2}  = [draw, fill=green!30, rectangle, text centered, minimum height=3em, minimum width=6em] 
\tikzstyle{blockgreen3}  = [draw, fill=green!45, rectangle, text centered, minimum height=3em, minimum width=6em] 
\tikzstyle{blockgreen4}  = [draw, fill=green!60, rectangle, text centered, minimum height=3em, minimum width=6em] 
\tikzstyle{blockgrey1}   = [draw, fill=black!8,  rectangle, text centered, minimum height=3em, minimum width=6em] 
\tikzstyle{blockgrey2}   = [draw, fill=black!16, rectangle, text centered, minimum height=3em, minimum width=6em] 
\tikzstyle{blockgrey3}   = [draw, fill=black!24, rectangle, text centered, minimum height=3em, minimum width=6em] 
\tikzstyle{blockgrey4}   = [draw, fill=black!32, rectangle, text centered, minimum height=3em, minimum width=6em] 
\tikzstyle{blockgrey5}   = [draw, fill=black!40, rectangle, text centered, minimum height=3em, minimum width=6em] 
\tikzstyle{blocknofill}  = [draw=black!50, line width=1.5pt, rectangle, rounded corners, text centered, minimum height=3em, minimum width=6em]
\tikzstyle{blockhigh}    = [draw, fill=blue!20, rectangle, text centered, minimum height=6em, minimum width=6em]
\tikzstyle{noblock}      = [draw=black!50, line width=1.5pt, rectangle, rounded corners, text centered, minimum height=3em, minimum width=6em]
\tikzstyle{sum}          = [draw, fill=blue!20, circle, node distance=1cm]
\tikzstyle{sumrest}      = [draw, fill=black, circle, radius=0.5cm]
\tikzstyle{pinstyle}     = [pin edge={to-,thin,black}]
\tikzstyle{title}        = [text centered]
\pgfdeclarelayer{background}
\pgfdeclarelayer{foreground}
\pgfsetlayers{background,main,foreground}

\usepackage{pgfplots}
\pgfplotsset{compat=1.13}
\pgfplotsset{colormap/bluered}
\usepackage{pgfplotstable}
\usepgfplotslibrary{polar}
\usepgfplotslibrary{colormaps}
\usepgfplotslibrary{external} 

\graphicspath{{figs/}} 					
\usepackage[parfill]{parskip} 					

\usepackage{amsmath} 			
\usepackage{amsfonts}			
\usepackage{mathtools}			
\everymath{\displaystyle}		
\usepackage{amssymb}  			
\usepackage{nicefrac}  			

\usepackage{tikz}  
\usetikzlibrary{shapes,arrows,calc,shadows,fit,intersections,datavisualization.formats.functions}
\tikzstyle{block}        = [draw, fill=blue!30, rectangle, text centered, minimum height=3em, minimum width=6em] 
\tikzstyle{title}        = [text centered]
\pgfdeclarelayer{background}
\pgfdeclarelayer{foreground}
\pgfsetlayers{background,main,foreground}

\usepackage{pgfplots} 
\pgfplotsset{compat=1.13}
\pgfplotsset{colormap/bluered}
\usepackage{pgfplotstable}
\usepgfplotslibrary{polar}
\usepgfplotslibrary{colormaps}
\usepgfplotslibrary{external}

\usepackage{graphicx, color} 	

\allowdisplaybreaks 

\newcommand{\sss}   {\scriptscriptstyle}    

\newcommand{\nm}   		[1] {\ensuremath{\mathrm{#1}}} 									
\newcommand{\neweq}     [2] {\begin{equation} \mathrm{#1}\label{#2} \end{equation}} 	

\newcommand{\CC}		{{C\nolinebreak[4]\hspace{-.05em}\raisebox{.4ex}{\tiny\bf ++}}}

\renewcommand{\vec}		[1] {\mbox{\boldmath{\ensuremath{\mathrm{#1}}}}}	
\newcommand{\lrp}       [1] {\left(#1\right)}								
\newcommand{\lrsb}      [1] {\left[#1\right]}								
\newcommand{\lrb}       [1] {\left\{#1\right\}}								

\newcommand{\Tzeroone}			{\vec T_{01}}
\newcommand{\Tzerotwo}			{\vec T_{02}}
\newcommand{\Tonetwo}			{\vec T_{12}}
\newcommand{\Tzeroonezero}		{\vec T_{01}^{0}}
\newcommand{\Tzerotwozero}		{\vec T_{02}^{0}}
\newcommand{\Tonetwozero}		{\vec T_{12}^{0}}
\newcommand{\Tonetwoone}		{\vec T_{12}^{1}}
\newcommand{\Tzerotwozerodot}	{\vec {\dot T}_{02}^{0}}
\newcommand{\Tonetwoonedot}		{\vec {\dot T}_{12}^{1}}
\newcommand{\Tzeroonezerodot}	{\vec {\dot T}_{01}^{0}}

\newcommand{\FC}		  {F_{\sss C}}

\newcommand{\Deltat}     	{\Delta t}
\newcommand{\DeltatTRUTH}	{\Delta t_{\sss TRUTH}}
\newcommand{\DeltatSENSED}  {\Delta t_{\sss SENSED}}

\newcommand{\DeltatGNSS}  	{\Delta t_{\sss GNSS}}
\newcommand{\DeltatION}  	{\Delta t_{\sss ION}}
\newcommand{\DeltatIMG}  	{\Delta t_{\sss IMG}}

\newcommand{\xvec}   			{\vec x}

\newcommand{\xvectilde} 		{\widetilde{\vec x}}

\newcommand{\xTRUTH}			{\vec x_{\sss TRUTH}}
\newcommand{\xSENSED}  			{\vec x_{\sss SENSED}}

\newcommand{\Sh}            {S_{\sss H}}
\newcommand{\Sv}            {S_{\sss V}}

\newcommand{\cIMGi}         {c_{\sss1}^{\sss IMG}}
\newcommand{\cIMGii}        {c_{\sss2}^{\sss IMG}}
\newcommand{\sPX}           {s_{\sss PX}}

\newcommand{\fovh}          {\Theta_{\sss H}}
\newcommand{\fovv}          {\Theta_{\sss V}}

\newcommand{\phiBCest}		  	{\hat{\vec \phi}^{\sss BC}}
\newcommand{\phiBC}			  	{\vec \phi_{\sss BC}}

\newcommand{\psiCest}			{\hat{\psi}^{\sss C}}
\newcommand{\thetaCest}	 		{\hat{\theta}^{\sss C}}
\newcommand{\xiCest}			{\hat{\xi}^{\sss C}}

\newcommand{\sigmapsiC}         {\nm{\sigma_{\psi^{\sss C}}}}
\newcommand{\sigmathetaC}  		{\nm{\sigma_{\theta^{\sss C}}}}
\newcommand{\sigmaxiC}     		{\nm{\sigma_{\xi^{\sss C}}}}
\newcommand{\sigmaphiBCest}		{\nm{\sigma_{\hat{\phi}^{\sss BC}}}}
\newcommand{\sigmaTBCBest}      {\nm{\sigma_{\hat{T}_{\sss BC}^{\sss B}}}}

\newcommand{\NpsiC}				{N_{\psi^{\sss C}}}
\newcommand{\NthetaC}			{N_{\theta^{\sss C}}}
\newcommand{\NxiC}				{N_{\xi^{\sss C}}}
\newcommand{\NpsiCest}			{N_{\hat{\psi}^{\sss C}}}
\newcommand{\NthetaCest}		{N_{\hat{\theta}^{\sss C}}}
\newcommand{\NxiCest}			{N_{\hat{\xi}^{\sss C}}}
\newcommand{\NTBCBiest}			{N_{\hat{T}_{\sss BC,1}^{\sss B}}}
\newcommand{\NTBCBiiest}		{N_{\hat{T}_{\sss BC,2}^{\sss B}}}
\newcommand{\NTBCBiiiest}		{N_{\hat{T}_{\sss BC,3}^{\sss B}}}

\newcommand{\TBCB}			{\vec T_{\sss BC}^{\sss B}}
\newcommand{\TBCBest}		{\hat{\vec T}_{\sss BC}^{\sss B}}
\newcommand{\TBCBfull}		{\vec T_{{\sss BC},full}^{\sss B}}
\newcommand{\TBCBempty}		{\vec T_{{\sss BC},empty}^{\sss B}}

\newcommand{\xEgdt}      {\vec x_{\sss GDT}}
\newcommand{\xEgdttilde} {\widetilde{\vec x}_{\sss GDT}}

\newcommand{\FN} 		  {F_{\sss N}}

\newcommand{\FB}		  {F_{\sss B}}

\newcommand{\FP}		  {F_{\sss P}}
\newcommand{\OP}          {O_{\sss P}}                            

\newcommand{\iPi}         {\vec i_{\sss1}^{\sss P}}
\newcommand{\iPii}        {\vec i_{\sss2}^{\sss P}}
\newcommand{\iPiii}       {\vec i_{\sss3}^{\sss P}}

\newcommand{\FA}		  {F_{\sss A}}							
\newcommand{\OA}          {O_{\sss A}}

\newcommand{\iAi}         {\vec i_{\sss1}^{\sss A}}
\newcommand{\iAii}        {\vec i_{\sss2}^{\sss A}}
\newcommand{\iAiii}       {\vec i_{\sss3}^{\sss A}}

\newcommand{\FY}		  {F_{\sss Y}}							
\newcommand{\OY}          {O_{\sss Y}}

\newcommand{\iYi}         {\vec i_{\sss1}^{\sss Y}}
\newcommand{\iYii}        {\vec i_{\sss2}^{\sss Y}}
\newcommand{\iYiii}       {\vec i_{\sss3}^{\sss Y}}

\newcommand{\FI}		 {F_{\sss I}}							

\newcommand{\gc}     	  {\vec g_c}

\newcommand{\vP} 	   			{\vec v^{\sss P}}

\newcommand{\vA}	    		{\vec v^{\sss A}}
\newcommand{\vY}	    		{\vec v^{\sss Y}}
\newcommand{\vN}	    		{\vec v^{\sss N}}

\newcommand{\vIB}           	{\vec v_{\sss IB}}
\newcommand{\vIP}           	{\vec v_{\sss IP}}
\newcommand{\vBP}           	{\vec v_{\sss BP}}

\newcommand{\vzeroone}			{\vec v_{\sss01}}
\newcommand{\vzerotwo}			{\vec v_{\sss02}}
\newcommand{\vonetwo}			{\vec v_{\sss12}}
\newcommand{\vzeroonezero}		{\vec v_{\sss01}^{\sss0}}
\newcommand{\vzerotwozero}		{\vec v_{\sss02}^{\sss0}}
\newcommand{\vonetwozero}		{\vec v_{\sss12}^{\sss0}}
\newcommand{\vonetwoone}		{\vec v_{\sss12}^{\sss1}}

\newcommand{\vzerotwozerodot}	{\dot {\vec v}_{\sss02}^{\sss0}}
\newcommand{\vonetwoonedot}		{\dot {\vec v}_{\sss12}^{\sss1}}
\newcommand{\vzeroonezerodot}	{\dot {\vec v}_{\sss01}^{\sss0}}

\newcommand{\vNtilde}			{\widetilde{\vec v}^{\sss N}}

\newcommand{\vtas}          {v_{\sss TAS}}

\newcommand{\vtastilde}     {\widetilde{v}_{\sss TAS}}

\newcommand{\wIB}				{\vec \omega_{\sss IB}}

\newcommand{\wIBB}				{\vec \omega_{\sss IB}^{\sss B}}

\newcommand{\wIBBskew}			{\widehat{\vec \omega}_{\sss IB}^{\sss B}}
\newcommand{\wIBBest}			{\hat{\vec \omega}_{\sss IB}^{\sss B}}
\newcommand{\wIBBestskew}		{\widehat{\hat{\vec \omega}}_{\sss IB}^{\sss B}}
\newcommand{\wIBBtildeskew}		{\widehat{\widetilde{\vec \omega}}_{\sss IB}^{\sss B}}
\newcommand{\wIBBtilde}			{\widetilde{\vec \omega}_{\sss IB}^{\sss B}}

\newcommand{\wIBskew}			{\widehat{\vec \omega}_{\sss IB}}
\newcommand{\alphaIBskew}		{\widehat{\vec \alpha}_{\sss IB}}
\newcommand{\alphaIBBskew}		{\widehat{\vec \alpha}_{\sss IB}^{\sss B}}
\newcommand{\alphaIBBest}		{\hat{\vec \alpha}_{\sss IB}^{\sss B}}
\newcommand{\alphaIBBestskew}	{\widehat{\hat{\vec \alpha}}_{\sss IB}^{\sss B}}
\newcommand{\alphaIBBtildeskew}	{\widehat{\widetilde{\vec \alpha}}_{\sss IB}^{\sss B}}
\newcommand{\alphaIBBtilde}  	{\widetilde{\vec \alpha}_{\sss IB}^{\sss B}}

\newcommand{\wIP}				{\vec \omega_{\sss IP}}

\newcommand{\wIPP}				{\vec \omega_{\sss IP}^{\sss P}}

\newcommand{\wIPPtilde}			{\widetilde{\vec \omega}_{\sss IP}^{\sss P}}
\newcommand{\wIPPtildetilde}	{\widetilde{\widetilde{\vec \omega}}_{\sss IP}^{\sss P}}

\newcommand{\wBP}			{\vec \omega_{\sss BP}}

\newcommand{\wzeroone}				{\vec \omega_{\sss01}}
\newcommand{\wzerotwo}				{\vec \omega_{\sss02}}
\newcommand{\wonetwo}				{\vec \omega_{\sss12}}
\newcommand{\wonetwozero}			{\vec \omega_{\sss12}^{\sss0}}
\newcommand{\wzerotwozero}			{\vec \omega_{\sss02}^{\sss0}}
\newcommand{\wzeroonezero}			{\vec \omega_{\sss01}^{\sss0}}
\newcommand{\wonetwoone}			{\vec \omega_{\sss12}^{\sss1}}
\newcommand{\wzerotwozerodot}		{\dot {\vec \omega}_{\sss02}^{\sss0}}
\newcommand{\wonetwoonedot}			{\dot{\vec \omega}_{\sss12}^{\sss1}}
\newcommand{\wzeroonezerodot}		{\dot{\vec \omega}_{\sss01}^{\sss0}}

\newcommand{\wzerooneskew}			{\widehat{\vec \omega}_{\sss01}}
\newcommand{\wzerooneoneskew}		{\widehat{\vec \omega}_{\sss01}^{\sss1}}
\newcommand{\wzeroonezeroskew}		{\widehat{\vec \omega}_{\sss01}^{\sss0}}
\newcommand{\wzerotwotwoskew}		{\widehat{\vec \omega}_{\sss02}^{\sss2}}
\newcommand{\wonetwotwoskew}		{\widehat{\vec \omega}_{\sss12}^{\sss2}}
\newcommand{\wzeroonetwoskew}		{\widehat{\vec \omega}_{\sss01}^{\sss2}}
\newcommand{\wzerooneoneskewdot}	{\dot{\widehat{\vec \omega}}_{\sss01}^{\sss1}}

\newcommand{\RBN}			{\vec R_{\sss BN}}

\newcommand{\RBP}			{\vec R_{\sss BP}}
\newcommand{\RBPest}		{\hat{\vec R}_{\sss BP}}

\newcommand{\RPA}			{\vec R_{\sss PA}^{\star}}
\newcommand{\RAP}			{\vec R_{\sss AP}^{\star}}
\newcommand{\RPY}			{\vec R_{\sss PY}^{\star}}
\newcommand{\RYP}			{\vec R_{\sss YP}^{\star}}

\newcommand{\Rzeroone}		{\vec R_{01}}
\newcommand{\Rzerotwo}		{\vec R_{02}}
\newcommand{\Ronetwo}		{\vec R_{12}}

\newcommand{\Rzeroonedot}	{\dot{\vec R}_{01}}
\newcommand{\Rzerotwodot}	{\dot{\vec R}_{02}}
\newcommand{\Ronetwodot}	{\dot{\vec R}_{12}}

\newcommand{\TBP}			{\vec T_{\sss BP}}

\newcommand{\TBPB}			{\vec T_{\sss BP}^{\sss B}}
\newcommand{\TBPBest}		{\hat{\vec T}_{\sss BP}^{\sss B}}
\newcommand{\TBPBfull}		{\vec T_{{\sss BP},full}^{\sss B}}
\newcommand{\TBPBempty}		{\vec T_{{\sss BP},empty}^{\sss B}}

\newcommand{\phiBP}			  	{\vec \phi^{\sss BP}}
\newcommand{\phiBPest}		  	{\hat{\vec \phi}^{\sss BP}}

\newcommand{\psiP}				{\psi^{\sss P}}
\newcommand{\thetaP}			{\theta^{\sss P}}
\newcommand{\xiP}				{\xi^{\sss P}}

\newcommand{\alphatilde}        {\widetilde{\alpha}}
\newcommand{\betatilde}         {\widetilde{\beta}}

\newcommand{\sigmapsiP}         {\nm{\sigma_{\psi^{\sss P}}}}
\newcommand{\sigmathetaP}  		{\nm{\sigma_{\theta^{\sss P}}}}
\newcommand{\sigmaxiP}     		{\nm{\sigma_{\xi^{\sss P}}}}
\newcommand{\sigmaphiBPest}		{\nm{\sigma_{\hat{\phi}^{\sss BP}}}}
\newcommand{\sigmaTBPBest}      {\nm{\sigma_{\hat{T}_{\sss BP}^{\sss B}}}}

\newcommand{\Hp}            {H_{\sss P}}

\newcommand{\fIB}				{\vec f_{\sss {IB}}}
\newcommand{\fIBB}				{\vec f_{\sss IB}^{\sss B}}

\newcommand{\fIBBtilde}			{\widetilde{\vec f}_{\sss IB}^{\sss B}}

\newcommand{\fIPP}				{\vec f_{\sss IP}^{\sss P}}
\newcommand{\fIPPtilde}			{\widetilde{\vec f}_{\sss IP}^{\sss P}}
\newcommand{\fIPPtildetilde}	{\widetilde{\widetilde{\vec f}}_{\sss IP}^{\sss P}}

\newcommand{\fIAA}				{\vec f_{\sss IA}^{\sss A}}
\newcommand{\fIAAtilde}			{\widetilde{\vec f}_{\sss IA}^{\sss A}}

\newcommand{\Nu}				{N_u}
\newcommand{\NuACC}				{\vec N_{u,\sss ACC}}

\newcommand{\Nuzero}			{N_{u0}}
\newcommand{\NuzeroACC}			{\vec N_{u0,\sss ACC}}

\newcommand{\NuzeroACCiest}		{\hat N_{u0,\sss{ACC,i}}}
\newcommand{\NuzeroGYR}			{\vec N_{u0,\sss GYR}}

\newcommand{\NuzeroGYRiest}		{\hat N_{u0,\sss{GYR,i}}}
\newcommand{\NuzeroMAG}			{\vec N_{u0,\sss MAG}}

\newcommand{\NuzeroMAGiest}		{\hat N_{u0,\sss{MAG,i}}}
\newcommand{\NhiMAG}			{\vec N_{\sss{HI},\sss MAG}}
\newcommand{\NzeroAOA}			{N_{0,\sss AOA}}
\newcommand{\NzeroAOS}			{N_{0,\sss AOS}}
\newcommand{\NzeroOSP}			{N_{0,\sss OSP}}
\newcommand{\NzeroOAT}			{N_{0,\sss OAT}}
\newcommand{\NzeroTAS}			{N_{0,\sss TAS}}
\newcommand{\Nui}				{N_{ui}}
\newcommand{\NuiACC}			{\vec N_{ui,\sss ACC}}
\newcommand{\NuiGYR}			{\vec N_{ui,\sss GYR}}
\newcommand{\Nv}				{N_v}
\newcommand{\Nvs}				{N_{vs}}
\newcommand{\Nvi}				{N_{vi}}
\newcommand{\NvACC}				{\vec N_{v,\sss ACC}}
\newcommand{\NvsACC}			{\vec N_{vs,\sss ACC}}
\newcommand{\NvGYR}				{\vec N_{v,\sss GYR}}
\newcommand{\NvsGYR}			{\vec N_{vs,\sss GYR}}
\newcommand{\NvsMAG}			{\vec N_{vs,\sss MAG}}
\newcommand{\NsAOA}				{N_{s,\sss AOA}}
\newcommand{\NsAOS}				{N_{s,\sss AOS}}
\newcommand{\NsTAS}				{N_{s,\sss TAS}}
\newcommand{\NsOSP}				{N_{s,\sss OSP}}
\newcommand{\NsOAT}				{N_{s,\sss OAT}}
\newcommand{\NgGNSSPOS}   		{\vec N_{g,\sss GNSS,POS}}
\newcommand{\NgGNSSVEL}			{\vec N_{g,\sss GNSS,VEL}}
\newcommand{\NzeroGNSSION}		{\vec N_{u0,\sss GNSS,ION}}
\newcommand{\NjGNSSION}			{\vec N_{j,\sss GNSS,ION}}
\newcommand{\NpsiP}				{N_{\psi^{\sss P}}}
\newcommand{\NthetaP}			{N_{\theta^{\sss P}}}
\newcommand{\NxiP}				{N_{\xi^{\sss P}}}
\newcommand{\NpsiPest}			{N_{\hat{\psi}^{\sss P}}}
\newcommand{\NthetaPest}		{N_{\hat{\theta}^{\sss P}}}
\newcommand{\NxiPest}			{N_{\hat{\xi}^{\sss P}}}
\newcommand{\NTBPBiest}			{N_{\hat{T}_{\sss BP,1}^{\sss P}}}
\newcommand{\NTBPBiiest}		{N_{\hat{T}_{\sss BP,2}^{\sss P}}}
\newcommand{\NTBPBiiiest}		{N_{\hat{T}_{\sss BP,3}^{\sss P}}}

\newcommand{\EACC}		    {\vec E_{\sss ACC}}

\newcommand{\BzeroACC}		{B_{0\sss{ACC}}}
\newcommand{\BzeroACCest}	{\hat{\vec B}_{0\sss{ACC}}}

\newcommand{\sigmauACC}		{\sigma_{u\sss{ACC}}}
\newcommand{\sigmavACC}		{\sigma_{v\sss{ACC}}}
\newcommand{\EGYR}			{\vec E_{\sss GYR}}

\newcommand{\BzeroGYR}		{B_{0\sss{GYR}}}
\newcommand{\BzeroGYRest}	{\hat{\vec B}_{0\sss{GYR}}}

\newcommand{\sigmauGYR}		{\sigma_{u\sss{GYR}}}
\newcommand{\sigmavGYR}		{\sigma_{v\sss{GYR}}}

\newcommand{\SACC}			{\vec S_{\sss ACC}}
\newcommand{\sACC}			{s_{\sss ACC}}
\newcommand{\sACCXi}		{s_{\sss {ACC,i}}}
\newcommand{\sACCXiest}		{\hat{s}_{\sss {ACC,i}}}
\newcommand{\sACCi}			{s_{\sss {ACC,1}}}
\newcommand{\sACCii}		{s_{\sss {ACC,2}}}
\newcommand{\sACCiii}		{s_{\sss {ACC,3}}}
\newcommand{\SGYR}			{\vec S_{\sss GYR}}
\newcommand{\sGYR}			{s_{\sss GYR}}

\newcommand{\sGYRXiest}		{\hat{s}_{\sss {GYR,i}}}
\newcommand{\sGYRi}			{s_{\sss {GYR,1}}}
\newcommand{\sGYRii}		{s_{\sss {GYR,2}}}
\newcommand{\sGYRiii}		{s_{\sss {GYR,3}}}

\newcommand{\alphaACC}			{\alpha_{\sss ACC}}
\newcommand{\alphaACCXi}		{\alpha_{\sss {ACC,i}}}

\newcommand{\alphaACCXj}		{\alpha_{\sss {ACC,j}}}
\newcommand{\alphaACCi}			{\alpha_{\sss {ACC,1}}}
\newcommand{\alphaACCii}		{\alpha_{\sss {ACC,2}}}
\newcommand{\alphaACCiii}		{\alpha_{\sss {ACC,3}}}

\newcommand{\alphaGYR}			{\alpha_{\sss GYR}}

\newcommand{\alphaGYRXij}		{\alpha_{\sss {GYR,ij}}}

\newcommand{\alphaGYRiXii}		{\alpha_{\sss {GYR,12}}}
\newcommand{\alphaGYRiiXi}		{\alpha_{\sss {GYR,21}}}
\newcommand{\alphaGYRiXiii}		{\alpha_{\sss {GYR,13}}}
\newcommand{\alphaGYRiiiXi}		{\alpha_{\sss {GYR,31}}}
\newcommand{\alphaGYRiiXiii}	{\alpha_{\sss {GYR,23}}}
\newcommand{\alphaGYRiiiXii}	{\alpha_{\sss {GYR,32}}}

\newcommand{\MACC}			{\vec M_{\sss ACC}}
\newcommand{\mACC}			{m_{\sss ACC}}
\newcommand{\MACCest}		{\hat{\vec M}_{\sss ACC}}
\newcommand{\mACCXij}		{m_{\sss {ACC,ij}}}
\newcommand{\mACCXijest}	{\hat{m}_{\sss {ACC,ij}}}
\newcommand{\NACC}			{\vec N_{\sss ACC}}
\newcommand{\nACCXij}		{n_{\sss {ACC,ij}}}
\newcommand{\MGYR}			{\vec M_{\sss GYR}}
\newcommand{\mGYR}			{m_{\sss GYR}}
\newcommand{\MGYRest}		{\hat{\vec M}_{\sss GYR}}
\newcommand{\mGYRXij}		{m_{\sss {GYR,ij}}}
\newcommand{\mGYRXijest}	{\hat{m}_{\sss {GYR,ij}}}

\newcommand{\sigmaAOA}		{\sigma_{\sss AOA}}
\newcommand{\sigmaAOS}		{\sigma_{\sss AOS}}
\newcommand{\sigmaOSP}		{\sigma_{\sss OSP}}
\newcommand{\sigmaOAT}		{\sigma_{\sss OAT}}
\newcommand{\sigmaTAS}		{\sigma_{\sss TAS}}
\newcommand{\BzeroAOA}		{B_{0\sss{AOA}}}
\newcommand{\BzeroAOS}		{B_{0\sss{AOS}}}
\newcommand{\BzeroOSP}		{B_{0\sss{OSP}}}
\newcommand{\BzeroOAT}		{B_{0\sss{OAT}}}
\newcommand{\BzeroTAS}		{B_{0\sss{TAS}}}

\newcommand{\sigmaGNSSPOS}		{\sigma_{\sss GNSS,POS}}
\newcommand{\sigmaGNSSPOSHOR}	{\sigma_{\sss GNSS,POS,HOR}}
\newcommand{\sigmaGNSSPOSVER}	{\sigma_{\sss GNSS,POS,VER}}
\newcommand{\sigmaGNSSVEL}		{\sigma_{\sss GNSS,VEL}}
\newcommand{\sigmaGNSSION}		{\sigma_{\sss GNSS,ION}}
\newcommand{\BzeroGNSSION}		{B_{0,\sss GNSS,ION}}

\newcommand{\BN}	    	{\vec B^{\sss N}}

\newcommand{\BNREAL}    	{\vec B_{\sss REAL}^{\sss N}}

\newcommand{\BB}	    	{\vec B^{\sss B}}
\newcommand{\BBtilde}		{\widetilde{\vec B}^{\sss B}}
\newcommand{\BBtildetilde}	{\widetilde{\widetilde{\vec B}}^{\sss B}}

\newcommand{\EMAG}	    		{\vec E_{\sss MAG}}

\newcommand{\BzeroMAG}			{B_{0,\sss MAG}}
\newcommand{\BzeroMAGvec}		{\vec B_{0,\sss MAG}}
\newcommand{\BzeroMAGvecest}	{\hat{\vec B}_{0,\sss MAG}}

\newcommand{\BhiMAG}			{B_{\sss{HI,MAG}}}
\newcommand{\BhiMAGvec}		 	{\vec B_{\sss{HI,MAG}}}
\newcommand{\BhiMAGvecest}		{\hat{\vec B}_{\sss{HI,MAG}}}

\newcommand{\BhiMAGXiest}		{\hat{B}_{\sss {HI,MAG,i}}}

\newcommand{\MMAG}			{\vec M_{\sss MAG}}
\newcommand{\MMAGest}		{\hat{\vec M}_{\sss MAG}}

\newcommand{\sigmavMAG}		{\sigma_{v,\sss MAG}}
\newcommand{\sMAG}			{s_{\sss MAG}}
\newcommand{\mMAG}			{m_{\sss MAG}}

\newcommand{\sMAGXiest}		{\hat{s}_{\sss {MAG,i}}}

\newcommand{\mMAGXijest}	{\hat{m}_{\sss {MAG,ij}}}

\newcommand{\seedA} 		{\Upsilon_{i,\sss A}}
\newcommand{\seedAACC} 		{\upsilon_{i,\sss A,\sss ACC}}
\newcommand{\seedAGYR}		{\upsilon_{i,\sss A,\sss GYR}}
\newcommand{\seedAMAG}		{\upsilon_{i,\sss A,\sss MAG}}
\newcommand{\seedAPLAT}		{\upsilon_{i,\sss A,\sss PLAT}}
\newcommand{\seedACAM}		{\upsilon_{i,\sss A,\sss CAM}}

\newcommand{\seedF} 		{\Upsilon_{j,\sss F}}
\newcommand{\seedFACC} 		{\upsilon_{j,\sss F,\sss ACC}}
\newcommand{\seedFGYR}		{\upsilon_{j,\sss F,\sss GYR}}
\newcommand{\seedFMAG}		{\upsilon_{j,\sss F,\sss MAG}}
\newcommand{\seedFOSP}		{\upsilon_{j,\sss F,\sss OSP}}
\newcommand{\seedFOAT}		{\upsilon_{j,\sss F,\sss OAT}}
\newcommand{\seedFTAS}		{\upsilon_{j,\sss F,\sss TAS}}
\newcommand{\seedFAOA}		{\upsilon_{j,\sss F,\sss AOA}}
\newcommand{\seedFAOS}		{\upsilon_{j,\sss F,\sss AOS}}
\newcommand{\seedFGNSS}		{\upsilon_{j,\sss F,\sss GNSS}}

\newcommand{\azeroone}		{\vec a_{\sss01}}
\newcommand{\azerotwo}		{\vec a_{\sss02}}
\newcommand{\aonetwo}		{\vec a_{\sss12}}
\newcommand{\azeroonezero}	{\vec a_{\sss01}^{\sss0}}
\newcommand{\azerotwozero}	{\vec a_{\sss02}^{\sss0}}
\newcommand{\aonetwoone}	{\vec a_{\sss12}^{\sss1}}
\newcommand{\aonetwozero}	{\vec a_{\sss12}^{\sss0}}

\newcommand{\alphazeroone}				{\vec \alpha_{\sss01}}
\newcommand{\alphazerotwo}				{\vec \alpha_{\sss02}}
\newcommand{\alphaonetwo}				{\vec \alpha_{\sss12}}
\newcommand{\alphazerotwozero}			{\vec \alpha_{\sss02}^{\sss0}}
\newcommand{\alphaonetwoone}			{\vec \alpha_{\sss12}^{\sss1}}
\newcommand{\alphazeroonezero}			{\vec \alpha_{\sss01}^{\sss0}}
\newcommand{\alphaonetwozero}			{\vec \alpha_{\sss12}^{\sss0}}
\newcommand{\alphazerooneskew}			{\widehat{\vec \alpha}_{\sss01}}
\newcommand{\alphazerooneoneskew}		{\widehat{\vec \alpha}_{\sss01}^{\sss1}}
\newcommand{\alphazeroonezeroskew}		{\widehat{\vec \alpha}_{\sss01}^{\sss0}}

\newcommand{\Bvec}						{\vec B}

\begin{document}

\title{Customizable Stochastic High Fidelity Model of the Sensors and Camera onboard a Low SWaP Fixed Wing Autonomous Aircraft}
\author{Eduardo Gallo\footnote{The author holds a MSc in Aerospace Engineering by the Polytechnic University of Madrid and has twenty-two years of experience working in aircraft trajectory prediction, modeling, and flight simulation. He is currently a Senior Trajectory Prediction and Aircraft Performance Engineer at Boeing Research \& Technology Europe (BR\&TE), although he is publishing this article in his individual capacity and time as part of his PhD thesis titled ``Autonomous Unmanned Air Vehicle GNSS-Denied Navigation'', advised by Dr. Antonio Barrientos within the Centre for Automation and Robotics of the Polytechnic University of Madrid.} \footnote{Contact: edugallo@yahoo.com, \url{https://orcid.org/0000-0002-7397-0425}}}
\date{February 2021}
\maketitle


\section*{Abstract}

The navigation systems of autonomous aircraft rely on the readings provided by a suite of onboard sensors to estimate the aircraft state. In the case of fixed wing vehicles, the sensor suite is composed by triads of accelerometers, gyroscopes, and magnetometers, a Global Navigation Satellite System (GNSS) receiver, and an air data system (Pitot tube, air vanes, thermometer, and barometer), and is often complemented by one or more digital cameras. An accurate representation of the behavior and error sources of each of these sensors, together with the images generated by the cameras, in indispensable for flight simulation and the evaluation of novel inertial or visual navigation algorithms, and more so in the case of low SWaP (size, weight, and power) aircraft, in which the quality and price of the sensors is limited. This article presents realistic and customizable models for each of these sensors, which have been implemented as an open-source \nm{\CC} simulation, available at \cite{Gallo2021_sensors}. Provided with the true variation of the aircraft state with time, the simulation provides a time stamped series of the errors generated by all sensors, as well as realistic images of the Earth surface that resemble those taken from a real camera flying along the indicated state positions and attitudes.


\section{Introduction and Outline}\label{sec:Introduction}

The sensors onboard an autonomous aircraft measure various aspects of the aircraft \emph{real} or \emph{actual trajectory} \nm{\xvec = \xTRUTH} and provide these measurements to the aircraft guidance, navigation and control (GNC) system. The outputs of these sensors, collectively known as the \emph{sensed trajectory} \nm{\xvectilde = \xSENSED}, represent the only link between the real but unknown actual trajectory and the GNC system in charge of achieving an actual trajectory that deviates as little as possible from the guidance targets.

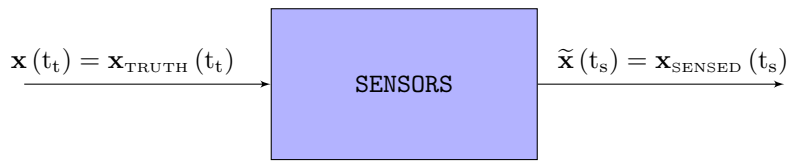
\begin{figure}[h]
\centering
\begin{tikzpicture}[auto, node distance=2cm,>=latex']
	\node [coordinate](input) {};
	\node [block, right of=input, minimum width=3.5cm, node distance=5.0cm, align=center, minimum height=2.0cm] (SENSORS) {\texttt{SENSORS}};	
	\node [coordinate, right of=SENSORS, node distance=5.0cm] (output) {};	
	\draw [->] (input) -- node[pos=0.4] {\nm{\xvec\lrp{t_t} = \xTRUTH\lrp{t_t}}} (SENSORS.west);
	\draw [->] (SENSORS.east) -- node[pos=0.55] {\nm{\xvectilde\lrp{t_s} = \xSENSED\lrp{t_s}}} (output);
\end{tikzpicture}
\caption{Sensors flow diagram}
\label{fig:Sensors_flow_diagram_generic}
\end{figure}

Note that although the actual aircraft state varies continuously in the real world, in simulation it is usually the outcome of a high frequency discrete integration process \cite{SIMULATION} that results in \nm{\xvec\lrp{t_t} = \xTRUTH\lrp{t_t}}, where \nm{t_t = t \cdot \DeltatTRUTH}. The sensed trajectory \nm{\xvectilde\lrp{t_s} = \xSENSED\lrp{t_s}}, where \nm{t_s = s \cdot \DeltatSENSED}, is however intrinsically discrete, although the working frequency of the different sensors may vary. This article considers that all sensors operate at the same rate of \nm{\DeltatSENSED}, with the exception of the GNSS receiver and the onboard camera, which work at \nm{\DeltatGNSS} and \nm{\DeltatIMG}, respectively. It is also assumed that all sensors are fixed to the aircraft structure in a strapdown configuration and that their measurement processes are instantaneous and time synchronized with each other at their respective frequencies. Table \ref{tab:Intro_trajectory_frequencies} provides the default frequencies employed in this article as well as in the corresponding software implementation \cite{Gallo2021_sensors}, where all values can be adjusted by the user.
\begin{center}
\begin{tabular}{lrr}
	\hline
	\multicolumn{1}{c}{\textbf{Discrete Time}} & \textbf{Frequency} & \textbf{Rate} \\
	\hline
	\nm{t_t = t \cdot \DeltatTRUTH}            & \nm{500 \lrsb{hz}} & \nm{0.002 \lrsb{sec}} \\
	\nm{t_s = s \cdot \DeltatSENSED}           & \nm{100 \lrsb{hz}} & \nm{0.01 \lrsb{sec}}  \\ 
	\nm{t_i = i \cdot \DeltatIMG}              & \nm{ 10 \lrsb{hz}} & \nm{0.1 \lrsb{sec}}   \\
	\nm{t_g = g \cdot \DeltatGNSS}             & \nm{  1 \lrsb{hz}} & \nm{1 \lrsb{sec}}     \\	
	\hline
\end{tabular}
\end{center}
\captionof{table}{Default frequencies of the states in the actual and sensed trajectories} \label{tab:Intro_trajectory_frequencies}

The sensed trajectory can be defined as a time stamped series of state vectors \nm{\xvectilde = \xSENSED} that groups the measurements provided by the different onboard sensors, comprising the only view of the actual trajectory at the disposal of the GNC system: 
\neweq{\xvectilde = \xSENSED = \lrsb{\fIBBtilde, \, \wIBBtilde, \, \BBtilde, \, \widetilde{\xvec}_{\sss GDT}, \, \vNtilde, \,  \widetilde{p}, \, \widetilde{T}, \, \vtastilde, \, \alphatilde, \, \betatilde, \, \, \vec I}^T} {eq:Sensors_state_vector_gnssactive}

Note that when present, the super index represents the frame or reference system in which a certain variable is viewed; if two sub indexes are present, it implies that the vector goes from the first frame to the second. Table \ref{tab:Sensors_ST} makes use of the body frame \nm{\FB}, rigidly attached to the aircraft structure with origin in its center of mass \cite{Farrell2008}, the NED frame \nm{\FN} also centered on the aircraft center of mass with axes in the North-East-Down directions \cite{Farrell2008}, and the inertial frame \nm{\FI}, usually considered as centered in the Sun with axes fixed with respect to other stars \cite{Etkin1972}. Additionally, the \emph{specific force} \nm{\fIB} is defined as the non gravitational acceleration experienced by the aircraft body with respect to an inertial frame \cite{Groves2008}.
\begin{center}
\begin{tabular}{lclcc}
	\hline
	\textbf{Components} & \textbf{Variable} & \textbf{Measured by} & \textbf{Acronym} & \textbf{Rate} \\
	\hline
	Specific Force            & \nm{\fIBBtilde}                     & Accelerometers    & \texttt{ACC}		& \nm{\DeltatSENSED} \\
	Inertial Angular Velocity & \nm{\wIBBtilde}                     & Gyroscopes        & \texttt{GYR}		& \nm{\DeltatSENSED} \\
	Magnetic Field            & \nm{\BBtilde}                       & Magnetometers     & \texttt{MAG}		& \nm{\DeltatSENSED} \\
	Geodetic Coordinates	  & \nm{\widetilde{\vec x}_{\sss GDT}}	& GNSS Receiver     & \texttt{GNSS}		& \nm{\DeltatGNSS} \\
	Ground Velocity			  & \nm{\vNtilde}                       & GNSS Receiver     & \texttt{GNSS}		& \nm{\DeltatGNSS} \\
	Air Pressure              & \nm{\widetilde{p}}                  & Barometer         & \texttt{OSP}\footnotemark		& \nm{\DeltatSENSED} \\
	Air Temperature           & \nm{\widetilde{T}}                  & Thermometer       & \texttt{OAT}\footnotemark		& \nm{\DeltatSENSED} \\
	Airspeed 				  & \nm{\vtastilde}            		    & Pitot Tube		& \texttt{TAS}\footnotemark		& \nm{\DeltatSENSED} \\
	Angle of Attack           & \nm{\alphatilde}					& Air Vanes			& \texttt{AOA}\footnotemark		& \nm{\DeltatSENSED} \\
	Angle of Sideslip         & \nm{\betatilde}						& Air Vanes			& \texttt{AOS}\footnotemark		& \nm{\DeltatSENSED} \\
	Image                     & \nm{\vec I}                         & Digital Camera    & \texttt{CAM}\footnotemark		& \nm{\DeltatIMG}  \\
	\hline
\end{tabular}
\end{center}
\captionof{table}{Components of sensed trajectory} \label{tab:Sensors_ST}
\addtocounter{footnote}{-5}	
\footnotetext{\texttt{OSP} stands for Outside Static Pressure.}
\addtocounter{footnote}{1}	
\footnotetext{\texttt{OAT} stands for Outside Static Temperature.}
\addtocounter{footnote}{1}	
\footnotetext{\texttt{TAS} stands for True Air Speed.}	
\addtocounter{footnote}{1}	
\footnotetext{\texttt{AOA} stands for Angle Of Attack.}	
\addtocounter{footnote}{1}	
\footnotetext{\texttt{AOS} stands for Angle of Sideslip.}	
\addtocounter{footnote}{1}	
\footnotetext{\texttt{CAM} stands for Camera.}	

The main objective of this article is to properly model the sources of error incurred by the different sensors onboard the aircraft, with special emphasis on the inertial sensors (accelerometers and gyroscopes) that provide the essential information on which the navigation system of low SWaP vehicles relies. The model parameters need to be \emph{customizable} so the user can employ the values that better resemble the performances of the specific equipment being modeled. They also need to be \emph{stochastic} to properly represent the nature of the different random processes involved, while ensuring that the time variation of the errors generated by each sensor can be repeated if so desired. Finally, the primary use of the sensor models developed in this article is to be part of a \emph{high fidelity} flight simulation such as that described in \cite{SIMULATION}, so the models need to be realistic and include as few simplifications as possible.

Section \ref{sec:Sensors_Simulation} describes the objectives of the simulation with special emphasis on its stochastic properties and how it can be customized for different sensors. It is followed by detailed explanations of the models representing the errors present in the measurements of the different sensors. Section \ref{sec:Sensors_Inertial} focuses on the inertial sensors and explains their characteristics and error sources, resulting in a model that captures their stochastic properties. A similar but less detailed approach is taken in section \ref{sec:Sensors_NonInertial} with the magnetometers, air data system, and GNSS receiver. Section \ref{sec:Sensors_camera} describes the characteristics of the digital camera employed for visual navigation, and the tool employed to generate realistic images that resemble what a real camera would view if located at the same position and attitude; although the camera differs from all other sensors in that it does not provide a measurement or reading but a digital image, in this article it is indeed considered a sensor as it provides the GNC system with information about its surroundings that can be employed for navigation. Last, section \ref{sec:Calibration} describes sensor calibration activities that are indispensable for the determination of various of the parameters present in the sensor models.


\section{Customizable and Stochastic Simulation}\label{sec:Sensors_Simulation}

The different sensor models described in this article have been implemented in a \texttt{C++} simulation (the open-source code is available at \cite{Gallo2021_sensors}) with the objective of properly simulating their stochastic nature while ensuring that the simulated variation with time of the errors can be repeated at a later time if so desired. The implementation hence follows these steps:
\begin{enumerate}

\item Initialize a discrete uniform distribution with any seed (any value is valid, so 1 was employed by the author), which produces random integers where each possible value has an equal likelihood of being produced. Call this distribution a number of times equal or higher than twice the maximum number of runs to be executed (each run provides the variation of time of all sensors for an unlimited amount of time, and corresponds to a single aircraft flight), divide them into two groups of the same size, and store the results for later use. These values, called \nm{\seedA} and \nm{\seedF}, are respectively the \emph{aircraft seeds} and the \emph{flight seeds}, where \say{i} is the aircraft number representing given fixed error realizations\footnote{Fixed error contributions vary from aircraft to aircraft but are constant for all flights of that aircraft.}, and \say{j} is the trajectory number representing run-to-run and in-run error realizations\footnote{Run-to-run and in-run error contributions vary from one flight of a given aircraft to the next.}. The stored aircraft and flight seeds become the initialization seeds for each of the simulation executions or runs, so this step does not need to be repeated.

\item Every time the simulator needs to obtain the errors generated by the different sensors (which usually correspond to a given flight), it is initialized with a given aircraft seed \nm{\seedA} together with a flight seed \nm{\seedF}. As these seeds are the only inputs required for all the stochastic processes within the sensors, the results of a given run can always be repeated by employing the same seeds. 
\begin{center}
\begin{tabular}{lccc}
	\hline
	\textbf{Type} & & \textbf{Error Sources} & \textbf{Seeds} \\
	\hline
	Aircraft	                & i 				& fixed & \nm{\seedAACC, \, \seedAGYR, \, \seedAMAG, \, \seedAPLAT, \, \seedACAM} \\
	\multirow{2}{*}{Flight}	& \multirow{2}{*}{j}& \multirow{2}{*}{run-to-run \& in-run}	& \nm{\seedFACC, \, \seedFGYR, \, \seedFMAG, \, \seedFOSP, \, \seedFOAT} \\
	            				& 					& & \nm{\seedFTAS, \, \seedFAOA, \, \seedFAOS, \, \seedFGNSS}  \\
	\hline
\end{tabular}
\end{center}
\captionof{table}{Sensor seeds} \label{tab:Sensors_seeds}

The selected aircraft and flight seeds are then employed to initialize two different discrete uniform distributions. One is executed five times to provide the \emph{fixed sensor seeds} (\nm{\seedAACC}, \nm{\seedAGYR}, \nm{\seedAMAG}, \nm{\seedAPLAT}, and \nm{\seedACAM}), while the other is realized nine times to obtain the \emph{run sensor seeds} (\nm{\seedFACC}, \nm{\seedFGYR}, \nm{\seedFMAG}, \nm{\seedFOSP}, \nm{\seedFOAT}, \nm{\seedFTAS}, \nm{\seedFAOA}, \nm{\seedFAOS}, and \nm{\seedFGNSS}). These seeds hence become the initialization seeds for each of the different sensors described throughout this article. 

\item Each sensor relies on either one or two standard normal distributions \nm{N\lrp{0, \, 1}}, depending on whether its error model is based exclusively on run-to-run and in-run error contributions or it also contains fixed error sources. The normal distributions of every sensor are initialized with the corresponding seeds (\nm{\upsilon_{i,\sss F}} and \nm{\upsilon_{j,\sss R}}) for that sensor.

\item Upon initialization, the \emph{fixed} normal distribution of every sensor is employed to generate all the values corresponding to scale factors, cross couplings, hard iron magnetism, and mounting errors. The \emph{run} normal distribution in turn is employed to generate the required bias offsets.

\item Once the simulation has been initialized, it is able to estimate the errors generated by each sensor working at the required sensor rate. As the simulation time advances, every time a sensor is called to provide a measurement, its already initialized and used run normal distribution is called to generate the corresponding random walk increments and white noises.

\item Additionally, when provided with a time series of the aircraft position and attitude, the simulation generates realistic images of the Earth surface that resemble those taken with an onboard camera.
\end{enumerate}

All the error models described in the following sections rely on a series of parameters whose value varies depending on the specific sensors represented in the simulation. Even though the user is expected to specify the value of these parameters, this article suggests the values used by the author to represent the sensors that could nowadays be installed onboard a small size fixed wing low SWaP autonomous aircraft, and which from this point on are referred as the ``\emph{Baseline}'' configuration. 


\section{Inertial Sensors}\label{sec:Sensors_Inertial}

Inertial sensors comprise \emph{accelerometers} and \emph{gyroscopes}, which measure specific force and inertial angular velocity about a single axis, respectively \cite{Hibbeler2015}. An \emph{inertial measurement unit} (IMU) encompasses multiple accelerometers and gyroscopes, usually three of each, obtaining three dimensional measurements of the specific force and angular rate \cite{Farrell2008} viewed in the platform frame \nm{\FP} (section \ref{subsec:Sensors_Inertial_Frames}). The individual accelerometers and gyroscopes however are not aligned with the \nm{\FP} axes, but with those of the non orthogonal accelerometers \nm{\FA} and gyroscopes \nm{\FY} frames, also defined in section \ref{subsec:Sensors_Inertial_Frames}. The output of the inertial sensors must hence first be transformed from the \nm{\FA} and \nm{\FY} frames to \nm{\FP}, as described in sections \ref{subsec:Sensors_Accelerometer_Triad_ErrorModel} and \ref{subsec:Sensors_Gyroscope_Triad_ErrorModel}, and then from the \nm{\FP} frame to the body frame \nm{\FB} as explained in section \ref{subsec:Sensors_Inertial_Mounting}, where they can be employed by the navigation system. The accelerometers and gyroscopes are assumed to be infinitesimally small and located at the IMU reference point (section \ref{subsec:Sensors_Inertial_Mounting}), which coincides with the origin of these three frames \nm{\lrp{\OP = \OA = \OY}}.

The IMU is physically attached to the aircraft structure in a strapdown configuration, so both the displacement \nm{\TBPB} and the Euler angles \nm{\phiBP = \lrsb{\psiP, \ \thetaP, \ \xiP}^T} that describe the relative position and rotation between the body \nm{\FB} and platform \nm{\FP} frames are constant. Accelerometers can be divided by their underlying technology into pendulous and vibrating beams, while gyroscopes are classified into spinning mass, optical (ring laser or fiber optic), and vibratory \cite{Groves2008}. Current inertial sensor development is mostly focused on \emph{micro machined electromechanical system} (MEMS) sensors\footnote{There exist both pendulous and vibrating beam MEMS accelerometers, but all MEMS gyroscopes are vibratory.}, which makes direct use of the chemical etching and batch processing techniques used by the electronics integrated circuit industry to obtain sensors with small size, low weight, rugged construction, low power consumption, low price, high reliability, and low maintenance \cite{Titterton2004}. On the negative side, the accuracy of MEMS sensors is still low, although tremendous progress has been achieved in the last two decades and more is expected in the future.

There is no universal classification of inertial sensors according to their performance, although they can be broadly assigned into five different categories or grades: marine (submarines and spacecraft), aviation (commercial and military), intermediate (small aircraft and helicopters), tactical (unmanned air vehicles and guided weapons), and automotive (consumer) \cite{Groves2008}. The full range of grades covers approximately six orders of magnitude of gyroscope performance and only three for the accelerometers, but higher performance is always associated with  bigger size, weight, and cost. Tactical grade IMUs cover a wide range of performance values, but can only provide a stand-alone navigation solution for a few minutes, while automotive grade IMUs are unsuitable for navigation.

The different errors that appear in the measurements provided by accelerometers and gyroscopes are described in section \ref{subsec:Sensors_Inertial_ErrorSources}. Section \ref{subsec:Sensors_Inertial_ErrorModelSingleAxis} presents a model for the measurements of a single inertial sensor, while sections \ref{subsec:Sensors_Inertial_ObtainmentSystemNoise} and \ref{subsec:Sensors_Inertial_ObtainmentBiasDrift} focus on how to obtain the specific values for white noise and bias on which the model relies from the documentation. Section \ref{subsec:Sensors_Inertial_Frames} describes the reference systems required to represent the IMU measurements. Additional errors appear when three accelerometers or gyroscopes are employed together, and these are modeled in section \ref{subsec:Sensors_Accelerometer_Triad_ErrorModel} for accelerometers and section \ref{subsec:Sensors_Gyroscope_Triad_ErrorModel} for gyroscopes. The analysis of the inertial sensors concludes with sections \ref{subsec:Sensors_Inertial_ErrorModel} and \ref{subsec:Sensors_Inertial_Selected_gyr_acc}, which provide a comprehensive error model for the IMU measurements, as well as the specifications of the baseline configuration. The final model also depends on the relative position of the IMU with respect to the body frame, which is described in section \ref{subsec:Sensors_Inertial_Mounting}.


\subsection{Inertial Sensor Error Sources}\label{subsec:Sensors_Inertial_ErrorSources}

In addition to the accelerometers and gyroscopes, an IMU also contains a processor, storage for the calibration parameters, one or more temperature sensors, and a power supply. As described below, each sensor has several error sources, but each of them has four components: \emph{fixed} contribution, \emph{temperature} dependent variation, \emph{run-to-run} variation, and \emph{in-run} variation \cite{Groves2008,Grewal2010}. The first two can be measured at the laboratory (at different temperatures) and the calibration results stored in the IMU so the processor can later compensate the sensor outputs based on the reading provided by the temperature sensor. Calibration however increases manufacturing costs so it may be absent in the case of inexpensive sensors. The run-to-run variation results in a contribution to a given error source that varies every time the sensor is employed but remains constant within a given run. It can not be compensated by the IMU processor but can be calibrated by the navigation system every time it is turned on with a process known as fine-alignment \cite{Farrell2008, Groves2008, Chatfield1997}. The in-run contribution to the error sources slowly varies during execution and can not be calibrated in the laboratory nor by the navigation system.
\begin{center}
\begin{tabular}{lcc}
	\hline
	\textbf{IMU Grade} & \textbf{Accelerometer Bias} \nm{\lrsb{mg}} & \textbf{Gyroscope Bias} \nm{\lrsb{deg/hr}} \\
	\hline
	Marine			& 0.01			& 0.001		\\
	Aviation		& 0.03 - 0.1	& 0.01		\\
	Intermediate	& 0.1 - 1		& 0.1		\\
	Tactical		& 1 - 10		& 1 - 100	\\
	Automotive		& > 10			& > 100		\\
	\hline
\end{tabular}
\end{center}
\captionof{table}{Typical inertial sensor biases according to IMU grade}\label{tab:Sensors_Inertial_bias}

Let's now discuss the different sources of error that influence an inertial sensor \cite{Groves2008,Trusov2011,KVH2014,Chow2011}:
\begin{itemize}
\item The \emph{bias} is an error exhibited by all accelerometers and gyroscopes that is independent of the underlying specific force or angular rate being measured, and comprises the dominant contribution to the overall sensor error. It can be defined as any nonzero output when the sensor input is zero \cite{Grewal2010}, and can be divided into its static and dynamic components.  The static part, also known as fixed bias, \emph{bias offset}, turn-on bias, or bias repeatability, comprises the run-to-run variation, while the dynamic component, known as in-run bias variation, \emph{bias drift}, bias instability (or stability), is typically about ten percent of the static part and slowly varies over periods of order one minute. As the bias is the main contributor to the overall sensor error, its value can be understood as a sensor quality measure. Table \ref{tab:Sensors_Inertial_bias} provides approximate values for the inertial sensor biases according to the IMU grade \cite{Groves2008}:

While the bias offset can be greatly reduced through fine-alignment \cite{Farrell2008, Groves2008, Chatfield1997}, the bias drift can not be determined and needs to be modeled as a stochastic process. It is mostly a warm up effect that should be almost non existent after a few minutes of operation, and corresponds to the minimum point of the sensor Allan curve \cite{KVH2014,IEEE1998}. It is generally modeled as as a random walk process obtained by the integration of a white noise signal, coupled with limits that represent the conclusion of the warm up process.

\item The \emph{scale factor error} is the departure of the input output gradient of the instrument from unity following unit conversion at the IMU processor. It represents a varying relationship between sensor input and output caused by aging and manufacturing tolerances. As it is a combination of fixed contribution plus temperature dependent variation, most of it can be eliminated through calibration (section \ref{subsec:Inertial_Calibration}).

\item The \emph{cross coupling error} or non orthogonality error is a fixed contribution that arises from the misalignment of the sensitive axes of the inertial sensors with respect to the orthogonal axes of the platform frame due to manufacturing limitations, and can also be highly reduced through calibration. The scale factor and cross coupling errors errors are in the order of \nm{10^{-4}} and \nm{10^{-3} \lrsb{-}} for most inertial sensors, although they can be higher for some low grade gyroscopes. The cross coupling error is equal to the sine of the misalignment, which is listed by some manufacturers.

\item \emph{System noise} or random noise is inherent to all inertial sensors and can combine electrical, mechanical, resonance, and quantization sources. It can originate at the sensor itself or at any other electronic equipment that interferes with it. System noise is a stochastic process usually modeled as white noise because its noise spectrum is approximately white, and can not be calibrated as there is no correlation between past and future values. A white noise process is characterized by its spectral power density PSD, which is constant as it does not depend on the signal frequency. It corresponds to the 1 [\nm{sec}] crossing of the sensor Allan curve \cite{KVH2014,IEEE1998}.

System noise is sometimes referred to as random walk, which can generate confusion with the bias. The reason is that the inertial sensor outputs are always integrated to obtain ground velocity in the case of accelerometers and aircraft attitude in the case of gyroscopes. As the integration of a white noise process is indeed a random walk, the later term is commonly employed to refer to system noise.

Table \ref{tab:Sensors_Inertial_noise} contains typical values for accelerometer and gyroscope root PSD according to sensor grade \cite{Groves2008}:
\begin{center}
\begin{tabular}{lcc}
	\hline
	\textbf{IMU Grade} & \textbf{Accelerometer Root PSD} \nm{\lrsb{m/sec/hr^{0.5}}} & \textbf{Gyroscope Root PSD} \nm{\lrsb{deg/hr^{0.5}}} \\
	\hline
	Aviation		& 0.012		& 0.002			\\
	Tactical		& 0.06		& 0.03 - 0.1	\\
	Automotive		& 0.6		& 1				\\
	\hline
\end{tabular}
\end{center}
\captionof{table}{Typical inertial sensor system noise according to IMU grade}\label{tab:Sensors_Inertial_noise}

\item Other minor error sources not considered in this article are g-dependent bias (sensitivity of spinning mass and vibratory gyroscopes to specific force), scale factor nonlinearity, and higher order errors (spinning mass gyroscopes and pendulous accelerometers).
\end{itemize}


\subsection{Single Axis Inertial Sensor Error Model}\label{subsec:Sensors_Inertial_ErrorModelSingleAxis}

Since all sensors modeled in this article are required to provide measurements at equispaced discrete times \nm{t_s = s \cdot \DeltatSENSED = s \cdot \Deltat}, this section focuses on obtaining a discrete model for the bias and white noise errors of a single axis inertial sensor. The results obtained here will be employed in the following sections to generate a comprehensive IMU model.

Let's consider a sensor in which the difference between its measurement at any given time \nm{\widetilde{x}\lrp{t}} and the real value of the physical magnitude being measured at that same time \nm{x\lrp{t}} can be represented by a zero mean white noise Gaussian process \nm{\eta_v\lrp{t}} with spectral density \nm{\sigma_v^2}:
\neweq{\widetilde{x}\lrp{t} = x\lrp{t} + \eta_v\lrp{t}}{eq:Sensor_SensorModel_wn01}

Dividing (\ref{eq:Sensor_SensorModel_wn01}) by \nm{\DeltatSENSED = \Deltat} and integrating results in:
\neweq{\dfrac{1}{\Deltat} \int_{t_0}^{t_0 + \Deltat} \widetilde{x}\lrp{t} \, dt = \dfrac{1}{\Deltat} \int_{t_0}^{t_0 + \Deltat} \lrsb{x\lrp{t} + \eta_v\lrp{t}} \, dt}{eq:Sensor_SensorModel_wn02}

Assuming that the measurement and real value are both constant over the integration interval\footnote{Note that the stochastic process \nm{\eta_v} can not be considered constant over any interval.} \cite{Crassidis2006} yields
\neweq{\widetilde{x}\lrp{t_0 + \Deltat} = x\lrp{t_0 + \Deltat} + \dfrac{1}{\Deltat} \int_{t_0}^{t_0 + \Deltat} \eta_v\lrp{t} \, dt}{eq:Sensor_SensorModel_wn03}

This expression results in the white noise sensor error \nm{w\lrp{t}}, which is the difference between the sensor measurement \nm{\widetilde{x}\lrp{t}} and the true value \nm{x\lrp{t}}. Its mean and variance can be readily computed:
\begin{eqnarray}
\nm{w\lrp{t_0 + \Deltat}} & = & \nm{\dfrac{1}{\Deltat} \int_{t_0}^{t_0 + \Deltat} \eta_v\lrp{t} \, dt} \label{eq:Sensor_SensorModel_wnerror_cont} \\
\nm{E\lrsb{w\lrp{t_0 + \Deltat}}} & = & \nm{0} \label{eq:Sensor_SensorModel_wnerror_mean} \\
\nm{Var\lrsb{w\lrp{t_0 + \Deltat}}} & = & \nm{\dfrac{\sigma_v^2}{\Deltat}} \label{eq:Sensor_SensorModel_wnerror_variance}
\end{eqnarray}

Based on these results, the white noise error can be modeled by a discrete random variable identically distributed to the above continuous white noise error, this is, one that results in the same mean and variance, where \nm{\Nv \sim N\lrp{0, \, 1}} is a standard normal random variable:
\neweq{w\lrp{s \, \DeltatSENSED} = w\lrp{s \, \Deltat} = \dfrac{\sigma_v}{\Deltat^{1/2}} \, \Nvs}{eq:Sensor_SensorModel_wnerror}

Let's now consider a second model in which the measurement error or bias is given by a first order random walk process or integration of a zero mean white noise Gaussian process \nm{\eta_u\lrp{t}} with spectral density \nm{\sigma_u^2}:
\neweq{\dot b\lrp{t} = \eta_u\lrp{t} \ \longrightarrow \ b\lrp{t_0 + \Deltat} = b\lrp{t_0} + \int_{t_0}^{t_0 + \Deltat} \eta_u\lrp{t} \, dt}{eq:Sensor_SensorModel_bias01}

Its mean and variance can be quickly computed:
\begin{eqnarray}
\nm{E\lrsb{b\lrp{t_0 + \Deltat}}} & = & \nm{E\lrsb{b\lrp{t_0}}} \label{eq:Sensor_SensorModel_bias_mean} \\
\nm{Var\lrsb{b\lrp{t_0 + \Deltat}}} & = & \nm{\sigma_u^2 \, \Deltat} \label{eq:Sensor_SensorModel_bias_variance}
\end{eqnarray}

These results indicate that the bias can be modeled by a discrete random variable identically distributed (this is, resulting in the same expected value and variance) to the continuous random walk above:
\neweq{b\lrp{t_0 + \Deltat} = b\lrp{t_0} + \sigma_u \, \Deltat^{1/2} \, \Nu} {eq:Sensor_SensorModel_bias_disc1}

where \nm{\Nu \sim N\lrp{0, \, 1}} is a standard normal random variable. Operating with the above expression results in the final expression for the discrete bias, as well as its mean and variance:
\begin{eqnarray}
\nm{b\lrp{s \, \Deltat}} & = & \nm{B_0 \, \Nuzero + \sigma_u \, \Deltat^{1/2} \, \sum_{i=1}^s \Nui} \label{eq:Sensor_SensorModel_bias_disc} \\
\nm{E\lrsb{b\lrp{s \, \Deltat}}} & = & \nm{0} \label{eq:Sensor_SensorModel_bias_disc_mean} \\
\nm{Var\lrsb{b\lrp{s \, \Deltat}}} & = & \nm{B_0^2 + \sigma_u^2 \, s \, \Deltat} \label{eq:Sensor_SensorModel_bias_discvariance}
\end{eqnarray}

A comprehensive single axis sensor error model without scale factor can hence be constructed by adding together the influence of the system noise provided by (\ref{eq:Sensor_SensorModel_wnerror}) and the bias given by (\ref{eq:Sensor_SensorModel_bias_disc}) \cite{Woodman2007}, while assuming that the standard normal random variables \nm{\Nu} and \nm{\Nv} are uncorrelated\footnote{Note that the expected value and variance of each of the two discrete components of this sensor model coincide with those of their continuous counterparts, but their combined mean and variance provided by expressions (\ref{eq:Sensor_SensorModel_error_mean}) and (\ref{eq:Sensor_SensorModel_error_variance}) differ from that of the combination of the two continuous error models given by (\ref{eq:Sensor_SensorModel_wnerror_cont}) and (\ref{eq:Sensor_SensorModel_bias01}). This is the case even if considering that the two zero mean white noise Gaussian processes \nm{\eta_u} and \nm{\eta_v} are independent and hence uncorrelated. It is however possible to obtain a discrete model whose discrete bias and white noise components are not only identically distributed that those of their continuous counterparts \cite{Crassidis2006}, even adding the equivalence of covariance between the bias and the sensor error, but this results in a significantly more complex model that behaves similarly to the one above at all but the shortest time samples after sensor initialization. The author has decided not to do so in the model described in this article, reducing complexity with little or no loss of realism.}:
\begin{eqnarray}
\nm{e_{\sss BW}\lrp{s \, \Deltat}} & = & \nm{\widetilde{x}\lrp{s \, \Deltat} - x\lrp{s \, \Deltat} = B_0 \, \Nuzero + \sigma_u \, \Deltat^{1/2} \, \sum_{i=1}^s \Nui + \dfrac{\sigma_v}{\Deltat^{1/2}} \, \Nvs} \label{eq:Sensor_SensorModel_error} \\
\nm{E\lrsb{e_{\sss BW}\lrp{s \, \Deltat}}} & = & \nm{0} \label{eq:Sensor_SensorModel_error_mean} \\
\nm{Var\lrsb{e_{\sss BW}\lrp{s \, \Deltat}}} & = & \nm{B_0^2 + \sigma_u^2 \, s \, \Deltat + \dfrac{\sigma_v^2}{\Deltat}} \label{eq:Sensor_SensorModel_error_variance}
\end{eqnarray}

The \emph{discrete sensor error} or difference between the measurement provided by the sensor \nm{\widetilde{x}\lrp{s \, \DeltatSENSED} = \widetilde{x}\lrp{s \, \Deltat}} at any given discrete time \nm{s \, \Deltat} and the real value of the physical variable being measured at that same discrete time \nm{x\lrp{s \, \Deltat}} is the combination of a bias or first order random walk and a white noise process, and depends on three parameters: the bias offset \nm{B_0}, the bias instability \nm{\sigma_u}, and the white noise \nm{\sigma_v}. The contributions of these three different sources to the sensor error as well as to its first and second integrals\footnote{Gyroscopes measure angular velocity and their output needs to be integrated once to obtain attitude, while accelerometers measure specific force and are integrated once to obtain velocity and twice to obtain position.} are very different and inherent to many of the challenges encountered when employing accelerometers and gyroscopes for inertial navigation, as explained below.

\begin{figure}[h]
\centering
\begin{tikzpicture}
\begin{axis}[
cycle list={{red,no markers,very thick},{blue,no markers,very thick},{orange!50!yellow, no markers},{violet, no markers},{green, no markers},{magenta, no markers},{olive,no markers}},
width=8.0cm, 
xmin=0, xmax=1000, xtick={0,200,400,600,800,1000},
xlabel={\nm{t \left[sec\right]}},
xmajorgrids,
ytick={-2,-1,0,1,2},
ylabel={\nm{E\left[e_{\sss BW}\left(s \,  \Deltat\right)\right] \, \left[10^{-1} \cdot unit\right]}},
ymajorgrids,
axis lines=left,
axis line style={-stealth},
legend entries={average 50 runs, theory},
legend pos=north west,
legend style={font=\footnotesize},
legend cell align=left,
]
\pgfplotstableread{figs/error_mean.txt}\mytable
\addplot table [header=false, x index=0,y index=1] {\mytable};
\addplot table [header=false, x index=0,y index=2] {\mytable};
\addplot table [header=false, x index=0,y index=3] {\mytable};
\addplot table [header=false, x index=0,y index=4] {\mytable};
\addplot table [header=false, x index=0,y index=5] {\mytable};
\addplot table [header=false, x index=0,y index=6] {\mytable};
\addplot table [header=false, x index=0,y index=7] {\mytable};
\pgfplotsset{cycle list shift = 2}
\addplot table [header=false, x index=0,y index=8] {\mytable};
\addplot table [header=false, x index=0,y index=9] {\mytable};
\addplot table [header=false, x index=0,y index=10] {\mytable};
\addplot table [header=false, x index=0,y index=11] {\mytable};
\addplot table [header=false, x index=0,y index=12] {\mytable};
\end{axis}	
\end{tikzpicture}%
\hskip 1pt
\begin{tikzpicture}
\begin{axis}[
cycle list={{red,no markers,very thick},{blue,no markers,very thick},{orange!50!yellow, no markers},{violet, no markers},{green, no markers}},
width=8.0cm,
xmin=0, xmax=1000, xtick={0,200,400,600,800,1000},
xlabel={\nm{t \left[sec\right]}},
xmajorgrids,
ymin=0, ymax=1.8, ytick={0,0.2,0.4,0.6,0.8,1,1.2,1.4,1.6},
ylabel={\nm{Var\left[e_{\sss BW}\left(s \,  \Deltat\right)\right]^{1/2} \, \left[10^{-1} \cdot unit\right]}},
ymajorgrids,
axis lines=left,
axis line style={-stealth},
legend entries={average 50 runs, theory: total, theory: bias offset, theory: bias instability, theory: white noise},
legend pos=north west,
legend style={font=\footnotesize},
legend cell align=left,
]
\pgfplotstableread{figs/error_std.txt}\mytable
\addplot table [header=false, x index=0,y index=1] {\mytable};
\addplot table [header=false, x index=0,y index=2] {\mytable};
\addplot table [header=false, x index=0,y index=3] {\mytable};
\addplot table [header=false, x index=0,y index=4] {\mytable};
\addplot table [header=false, x index=0,y index=5] {\mytable};
\end{axis}		
\end{tikzpicture}%
\caption{Propagation with time of sensor error mean and standard deviation}
\label{fig:sensors_error}
\end{figure}
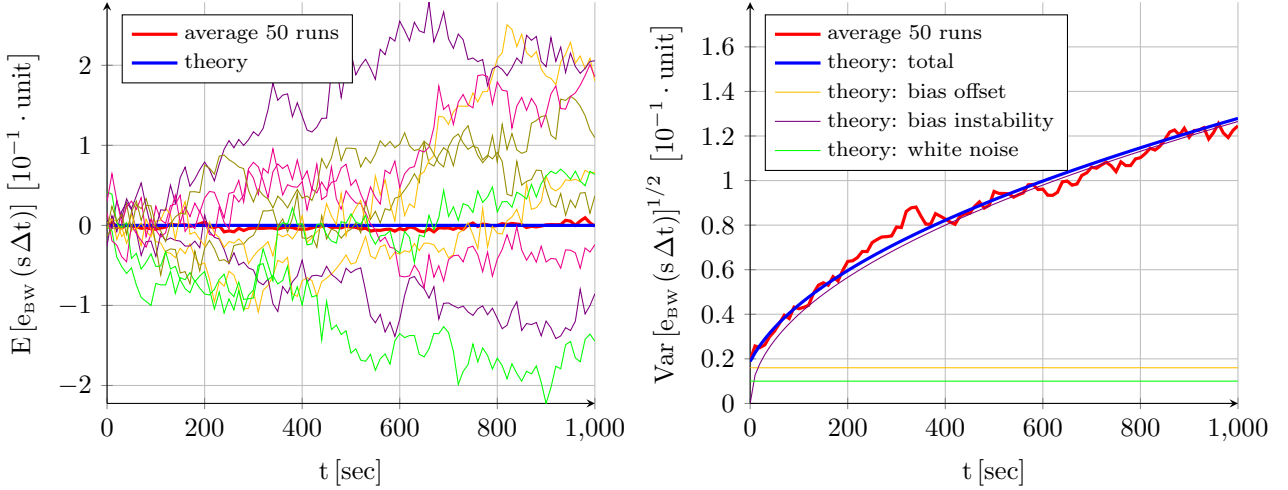

Figure \ref{fig:sensors_error} represents the performance of a fictitious sensor of \nm{B_0 = 1.6 \cdot 10^{-2}}, \nm{\sigma_u = 4 \cdot 10^{-3}}, and \nm{\sigma_v = 1 \cdot 10^{-3}} working at a frequency of \nm{100 \lrsb{hz} \ \lrp{\Deltat = 0.01 \lrsb{sec}}}, and is intended to showcase the different behavior and relative influence on the total error of each of its three components. The figures show the theoretical variation with time of the sensor error mean (left side) and standard deviation (right side) given by (\ref{eq:Sensor_SensorModel_error_mean}) and (\ref{eq:Sensor_SensorModel_error_variance}) together with the average of fifty different runs. In addition, the left figure also includes ten of those runs to showcase the variability in results implicit to the random variables\footnote{Although the data is generated at \nm{100 \lrsb{hz}}, the figure only employs 1 out of every 1000 points, so it appears far less noisy than the real data.}, while the right side shows the theoretical contribution to the standard deviation of each of the three components. In addition to the near equivalence between the theory and the average of several runs, the figure shows that the bias instability is the commanding long term factor in the deviation between the sensor measurement and its zero mean (the standard deviation of the bias instability grows with the square root of time while the other two components are constant). As discussed in section \ref{subsec:Sensors_Inertial_ErrorSources}, the bias drift or bias instability is indeed the most important quality parameter of an inertial sensor. This is also the case when the sensor output is integrated, as discussed below.

Let's integrate the sensor error over a timespan \nm{s \, \Deltat} to evaluate the growth with time of both its expected value and its variance\footnote{As the interest lies primarily in \nm{s \gg 1}, a simple integration method such as the rectangular rule is employed.}:
\begin{eqnarray}
\nm{f_{\sss BW}\lrp{s \, \Deltat}} & = & \nm{f_{\sss BW}\lrp{0} + \int_0^{s \, \Deltat} e_{\sss BW}\lrp{\tau} \, d\tau = f_{\sss BW}\lrp{0} + \Deltat \sum_{i=1}^s e_{\sss BW}\lrp{i \, \Deltat}} \nonumber \\
& = & \nm{f_{\sss BW}\lrp{0} + B_0 \, \Nuzero \, s \, \Deltat + \sigma_u \, \Deltat^{3/2} \sum_{i=1}^s \lrp{s-i+1} \, \Nui + \sigma_v \, \Deltat^{1/2} \sum_{i=1}^s \Nvi} \label{eq:Sensor_SensorModel_1st} \\
\nm{E\lrsb{f_{\sss BW}\lrp{s \, \Deltat}}} & = & \nm{f_{\sss BW}\lrp{0}} \label{eq:Sensor_SensorModel_1st_mean} \\
\nm{Var\lrsb{f_{\sss BW}\lrp{s \, \Deltat}}} & = & \nm{B_0^2 \, \lrp{s \, \Deltat}^2 + \dfrac{\sigma_u^2}{6} \, \Deltat^3 \, s \, \lrp{s + 1} \, \lrp{2 \, s + 1} + \sigma_v^2 \, \lrp{s \, \Deltat}} \nonumber \\
& \nm{\approx} & \nm{B_0^2 \, \lrp{s \, \Deltat}^2 + \dfrac{\sigma_u^2}{3} \, \lrp{s \, \Deltat}^3 + \sigma_v^2 \, \lrp{s \, \Deltat}} \label{eq:Sensor_SensorModel_1st_variance}
\end{eqnarray}

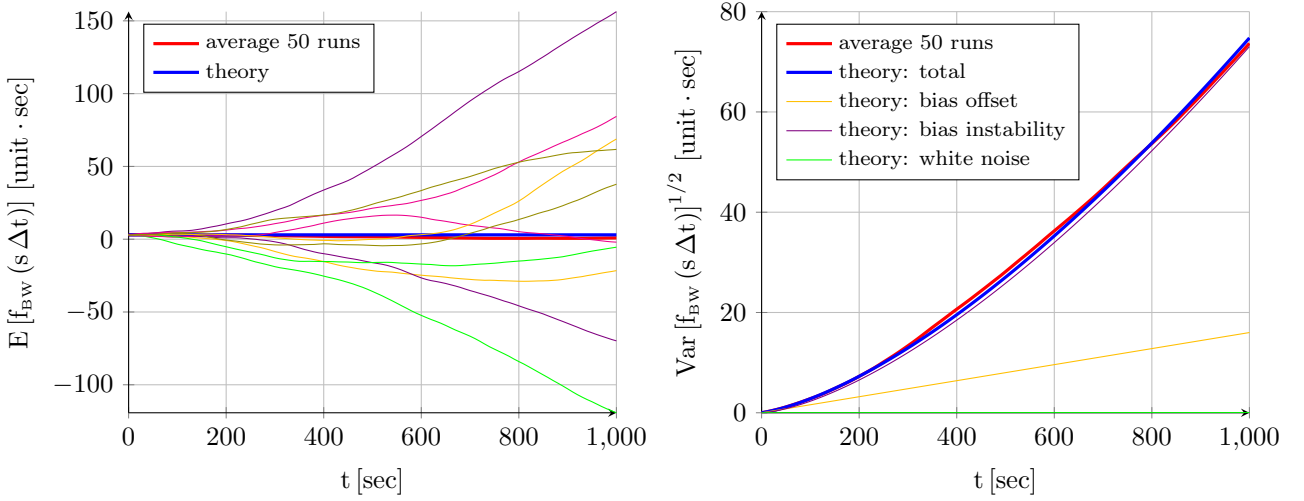
\begin{figure}[h]
\centering
\begin{tikzpicture}
\begin{axis}[
cycle list={{red,no markers,very thick},{blue,no markers,very thick},{orange!50!yellow, no markers},{violet, no markers},{green, no markers},{magenta, no markers},{olive,no markers}},
width=8.0cm, 
xmin=0, xmax=1000, xtick={0,200,400,600,800,1000},
xlabel={\nm{t \left[sec\right]}},
ytick={-100,-50,0,50,100,150},
xmajorgrids,
ylabel={\nm{E\left[f_{\sss BW}\left(s \,  \Deltat\right)\right] \, \left[unit \cdot sec\right]}},
ymajorgrids,
axis lines=left,
axis line style={-stealth},
legend entries={average 50 runs, theory},
legend pos=north west,
legend style={font=\footnotesize},
legend cell align=left,
]
\pgfplotstableread{figs/error_1st_integral_mean.txt}\mytable
\addplot table [header=false, x index=0,y index=1] {\mytable};
\addplot table [header=false, x index=0,y index=2] {\mytable};
\addplot table [header=false, x index=0,y index=3] {\mytable};
\addplot table [header=false, x index=0,y index=4] {\mytable};
\addplot table [header=false, x index=0,y index=5] {\mytable};
\addplot table [header=false, x index=0,y index=6] {\mytable};
\addplot table [header=false, x index=0,y index=7] {\mytable};
\pgfplotsset{cycle list shift = 2}
\addplot table [header=false, x index=0,y index=8] {\mytable};
\addplot table [header=false, x index=0,y index=9] {\mytable};
\addplot table [header=false, x index=0,y index=10] {\mytable};
\addplot table [header=false, x index=0,y index=11] {\mytable};
\addplot table [header=false, x index=0,y index=12] {\mytable};
\end{axis}	
\end{tikzpicture}%
\hskip 1pt
\begin{tikzpicture}
\begin{axis}[
cycle list={{red,no markers,very thick},{blue,no markers,very thick},{orange!50!yellow, no markers},{violet, no markers},{green, no markers}},
width=8.0cm,
xmin=0, xmax=1000, xtick={0,200,400,600,800,1000},
xlabel={\nm{t \left[sec\right]}},
xmajorgrids,
ymin=0, ymax=80, ytick={0,20,40,60,80},
ylabel={\nm{Var\left[f_{\sss BW}\left(s \, \Deltat\right)\right]^{1/2} \, \left[unit \cdot sec\right]}},
ymajorgrids,
axis lines=left,
axis line style={-stealth},
legend entries={average 50 runs, theory: total, theory: bias offset, theory: bias instability, theory: white noise},
legend pos=north west,
legend style={font=\footnotesize},
legend cell align=left,
]
\pgfplotstableread{figs/error_1st_integral_std.txt}\mytable
\addplot table [header=false, x index=0,y index=1] {\mytable};
\addplot table [header=false, x index=0,y index=2] {\mytable};
\addplot table [header=false, x index=0,y index=3] {\mytable};
\addplot table [header=false, x index=0,y index=4] {\mytable};
\addplot table [header=false, x index=0,y index=5] {\mytable};
\end{axis}		
\end{tikzpicture}%
\caption{Propagation with time of first integral of sensor error mean and standard deviation}
\label{fig:sensors_1st_integral}
\end{figure}

Figure \ref{fig:sensors_1st_integral} follows the same pattern as figure \ref{fig:sensors_error} but applied to the error integral instead of to the error itself. It would represent the attitude error resulting from integrating the gyroscope output, or the velocity error expected when integrating the specific force measured by an accelerometer. The conclusions are the same as before but significantly more accentuated. Not only is the expected value of the error constant instead of zero (\nm{f_{\sss BW}\lrp{0} = 3} has been employed in the experiment), but the growth in the standard deviation (over a non zero mean) is much quicker than before. The bias instability continues to be the dominating factor but now increases with a power of \nm{t^{3/2}}, while the bias offset and white noise contributions also increase with time, although with powers of \nm{t} and \nm{t^{1/2}} respectively. Let's continue the process and integrate the error a second time:
\begin{eqnarray}
\nm{g_{\sss BW}\lrp{s \, \Deltat}} & = & \nm{g_{\sss BW}\lrp{0} + \int_0^{s \, \Deltat} f_{\sss BW}\lrp{\tau} \, d\tau = g_{\sss BW}\lrp{0} + \Deltat \sum_{i=1}^s f_{\sss BW}\lrp{i \, \Deltat} = g_{\sss BW}\lrp{0} + f_{\sss BW}\lrp{0} \, s \, \Deltat +} \nonumber \\
& & \nm{+ \dfrac{B_0}{2} \, \Nuzero \, \lrp{s \, \Deltat}^2 + \sigma_u \, \Deltat^{5/2} \sum_{i=1}^s \sum_{j=1}^{s-i+1} j \, \Nui + \sigma_v \, \Deltat^{3/2} \sum_{i=1}{s} \lrp{s-i+1} \, \Nvi} \label{eq:Sensor_SensorModel_2nd} \\
\nm{E\lrsb{g_{\sss BW}\lrp{s \, \Deltat}}} & = & \nm{g_{\sss BW}\lrp{0} + f_{\sss BW}\lrp{0} \, \lrp{s \, \Deltat}} \label{eq:Sensor_SensorModel_2nd_mean} \\
\nm{Var\lrsb{g_{\sss BW}\lrp{s \,  \Deltat}}} & = & \nm{\dfrac{B_0^2}{4} \, \lrp{s \, \Deltat}^4 + \sigma_u^2 \, \Deltat^5 \, \sum_{i=1}^s \lrp{\sum_{j=1}^{s-i+1} j}^2 + \dfrac{\sigma_v^2}{6} \, \Deltat^3 \, s \, \lrp{s + 1} \, \lrp{2 \, s + 1}} \nonumber \\
& \nm{\approx} & \nm{\dfrac{B_0^2}{4} \, \lrp{s \, \Deltat}^4 + \dfrac{\sigma_u^2}{20} \, \lrp{s \, \Deltat}^5 + \dfrac{\sigma_v^2}{3} \, \lrp{s \, \Deltat}^3} \label{eq:Sensor_SensorModel_2nd_variance}
\end{eqnarray}

\begin{figure}[h]
\centering
\begin{tikzpicture}
\begin{axis}[
cycle list={{red,no markers,very thick},{blue,no markers,very thick},{orange!50!yellow, no markers},{violet, no markers},{green, no markers},{magenta, no markers},{olive,no markers}},
width=8.0cm, 
xmin=0, xmax=1000, xtick={0,200,400,600,800,1000},
xlabel={\nm{t \left[sec\right]}},
xmajorgrids,
ytick={-4,-2,0,2,4,6},
ylabel={\nm{E\left[g_{\sss BW}\left(s \, \Deltat\right)\right] \, \left[10^4 \cdot unit \cdot sec^2\right]}},
ymajorgrids,
axis lines=left,
axis line style={-stealth},
legend entries={average 50 runs, theory},
legend pos=north west,
legend style={font=\footnotesize},
legend cell align=left,
]
\pgfplotstableread{figs/error_2nd_integral_mean.txt}\mytable
\addplot table [header=false, x index=0,y index=1] {\mytable};
\addplot table [header=false, x index=0,y index=2] {\mytable};
\addplot table [header=false, x index=0,y index=3] {\mytable};
\addplot table [header=false, x index=0,y index=4] {\mytable};
\addplot table [header=false, x index=0,y index=5] {\mytable};
\addplot table [header=false, x index=0,y index=6] {\mytable};
\addplot table [header=false, x index=0,y index=7] {\mytable};
\pgfplotsset{cycle list shift = 2}
\addplot table [header=false, x index=0,y index=8] {\mytable};
\addplot table [header=false, x index=0,y index=9] {\mytable};
\addplot table [header=false, x index=0,y index=10] {\mytable};
\addplot table [header=false, x index=0,y index=11] {\mytable};
\addplot table [header=false, x index=0,y index=12] {\mytable};
\end{axis}	
\end{tikzpicture}%
\hskip 1pt
\begin{tikzpicture}
\begin{axis}[
cycle list={{red,no markers,very thick},{blue,no markers,very thick},{orange!50!yellow, no markers},{violet, no markers},{green, no markers}},
width=8.0cm,
xmin=0, xmax=1000, xtick={0,200,400,600,800,1000},
xlabel={\nm{t \left[sec\right]}},
xmajorgrids,
ymin=0, ymax=3.0, ytick={0,0.5,1,1.5,2,2.5,3},
ylabel={\nm{Var\left[g_{\sss BW}\left(s \, \Deltat\right)\right]^{1/2} \, \left[10^4 \cdot unit \cdot sec^2\right]}},
ymajorgrids,
axis lines=left,
axis line style={-stealth},
legend entries={average 50 runs, theory: total, theory: bias offset, theory: bias instability, theory: white noise},
legend pos=north west,
legend style={font=\footnotesize},
legend cell align=left,
]
\pgfplotstableread{figs/error_2nd_integral_std.txt}\mytable
\addplot table [header=false, x index=0,y index=1] {\mytable};
\addplot table [header=false, x index=0,y index=2] {\mytable};
\addplot table [header=false, x index=0,y index=3] {\mytable};
\addplot table [header=false, x index=0,y index=4] {\mytable};
\addplot table [header=false, x index=0,y index=5] {\mytable};
\end{axis}		
\end{tikzpicture}%
\caption{Propagation with time of second integral of sensor error mean and standard deviation}
\label{fig:sensors_2nd_integral}
\end{figure}
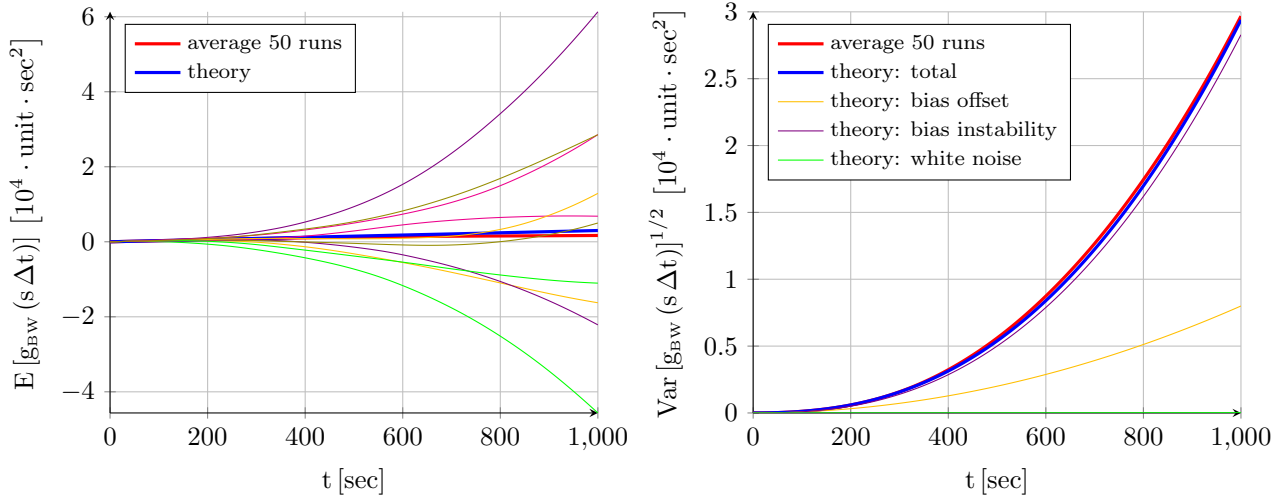

Figure \ref{fig:sensors_2nd_integral} shows the same type of figures but applied to the second integral of the error (\nm{g_{\sss BW}\lrp{0} = 1.5 } has been employed in the experiment). In this case the degradation of the results with time is even more intense to the point were the measurements are useless after a very short period of time. Unless corrected by the navigation system, this is equivalent to the error in position obtained by double integrating the output of the accelerometers.

Let's summarize the main points of the single axis inertial sensor discrete error model developed in this section, that includes the influence of the bias and the system error but not that of the scale factor and cross coupling errors included in the three dimensional error model of section \ref{subsec:Sensors_Inertial_ErrorModel}. The error \nm{e_{\sss BW}\lrp{s \, \Deltat}}, which applies to specific force for accelerometers and inertial angular velocity in the case of gyroscopes, depends on three factors: bias offset \nm{B_0}, bias drift \nm{\sigma_u}, and white noise \nm{\sigma_v}. Its mean is always zero, but the error standard deviation grows with time (\nm{\propto \, t^{1/2}}) due to the bias drift with constant contributions from the bias offset and the white noise. When integrating the error to obtain \nm{f_{\sss BW}\lrp{s \, \Deltat}}, equivalent to ground velocity for accelerometers and attitude for gyroscopes, the initial speed error or initial attitude error \nm{f_{\sss BW}\lrp{0}} becomes the fourth contributor, and an important one indeed, as it becomes the mean of the first integral error at any time. The standard deviation, which measures the spread over the non zero mean, increases very quickly with time because of the bias instability (\nm{\propto \, t^{3/2}}), with contributions from the offset (\nm{\propto \, t}) and the white noise (\nm{\propto \, t^{1/2}}). If integrating a second time to obtain \nm{g_{\sss BW}\lrp{s \, \Deltat}}, or position in case of the accelerometer, the initial position error \nm{g_{\sss BW}\lrp{0}} turns into the fifth contributor. The expected value of the position error grows linearly with time due to the initial velocity error with an additional constant contribution from the initial position error, while the position standard deviation (measuring spread over a growing average value) grows extremely quick due mostly to the bias instability (\nm{\propto \, t^{5/2}}), but also because of the bias offset (\nm{\propto \, t^{2}}) and the white noise (\nm{\propto \, t^{3/2}}). Table \ref{tab:Sensors_Inertial_units} shows the standard units of the different sources of error for both accelerometers and gyroscopes.
\begin{center}
\begin{tabular}{lccccc}
	\hline
	\textbf{Units} & \nm{B_0} & \nm{\sigma_u} & \nm{\sigma_v} & \nm{f_{\sss BW}\lrp{0}} & \nm{g_{\sss BW}\lrp{0}} \\
	\hline
	Accelerometer	& \nm{m/sec^2}	& \nm{m/sec^{2.5}}		& \nm{m/sec^{1.5}}		& \nm{m/sec}	& \nm{m}	\\
	Gyroscope		& \nm{deg/sec}	& \nm{deg/sec^{1.5}}	& \nm{deg/sec^{0.5}}	& \nm{deg}		& N/A		\\
	\hline
\end{tabular}
\end{center}
\captionof{table}{Units for single axis inertial sensor error sources}\label{tab:Sensors_Inertial_units}


\subsection{Obtainment of System Noise Values}\label{subsec:Sensors_Inertial_ObtainmentSystemNoise}

This section focuses on the significance of system or white noise error \nm{\sigma_v} and how to obtain it from sensor specifications, which often refer to the integral of the output instead of the output itself. As the integral of white noise is a random walk process, the angle random walk of a gyroscope is equivalent to white noise in the angular rate output, while velocity random walk refers to specific force white noise in accelerometers \cite{Grewal2010}. The discussion that follows applies to gyroscopes but is fully applicable to accelerometers if replacing angular rate by specific force and attitude or angle by ground velocity.

Angle random walk, measured in [\nm{rad/sec^{1/2}}], [\nm{deg/hr^{1/2}}], or equivalent, describes the average deviation or error that occurs when the sensor output signal is integrated due to system noise exclusively, without considering other error sources such as bias or scale factor \cite{Stockwell}. If integrating multiple times and obtaining a distribution of end points at a given final time \nm{s \, \Deltat}, the standard deviation of this distribution, containing the final angles at the final time, scales linearly with the white noise level \nm{\sigma_v}, the square root of the integration step size \nm{\Deltat}, and the square root of the number of steps \nm{s}, as noted by the last term of (\ref{eq:Sensor_SensorModel_1st_variance}). This means that an angle random walk of 1 [\nm{deg/sec^{1/2}}] translates into an standard deviation for the error of 1 [\nm{deg}] after 1 [\nm{sec}], 10 [\nm{deg}] after 100 [\nm{sec}], and \nm{1000^{1/2} \approx 31.6} [\nm{deg}] after 1000 [\nm{sec}].

Manufacturers often provide this information as the power spectral density PSD of the white noise process in [\nm{deg^2/hr^2/hz}] or equivalent, where it is necessary to take its square root to obtain \nm{\sigma_v}, or as the root PSD in [\nm{deg/hr/hz^{1/2}}] that is equivalent to \nm{\sigma_v}. Sometimes it is even provided as the PSD of the random walk process, not the white noise, in units [\nm{deg/hr}] or equivalent. It is then necessary to multiply this number by the square root of the sampling interval \nm{\Deltat} or divide it by the square root of the sampling frequency to obtain the desired \nm{\sigma_v} value.


\subsection{Obtainment of Bias Drift Values}\label{subsec:Sensors_Inertial_ObtainmentBiasDrift}

This section describes the meaning of bias instability \nm{\sigma_u} (also known as bias stability or bias drift) and how to obtain it from sensor specifications. As in the previous section, the discussion is centered on gyroscopes but it is fully applicable to accelerometers as well. Bias instability can be defined as the potential of the sensor error to stay within a certain range for a certain time \cite{Renaut2013}.  A small number of manufacturers directly provide sensor output changes over time, which directly relates with the bias instability (also known as in-run bias variation, bias drift, or rate random walk) per the second term of (\ref{eq:Sensor_SensorModel_error_variance}). If provided with an angular rate change of x [\nm{deg/sec}] (\nm{1\sigma}) in t [\nm{sec}], then \nm{\sigma_u} can be obtained as follows \cite{Crassidis2006,Farrenkopf1974}:
\neweq{\sigma_u = \dfrac{x}{t^{1/2}}}{eq:Sensor_SensorModel_sigmau}

As the bias drift is responsible for the growth of sensor error with time (figure \ref{fig:sensors_error}), manufacturers more commonly provide bias stability measurements that describe how the bias of a device may change over a specified period of time \cite{Woodman2007}, typically around 100 [\nm{sec}]. Bias stability is usually specified as a \nm{1\sigma} value with units [\nm{deg/hr}] or [\nm{deg/sec}], which can be interpreted as follows according to (\ref{eq:Sensor_SensorModel_error}, \ref{eq:Sensor_SensorModel_error_mean}, \ref{eq:Sensor_SensorModel_error_variance}). If the sensor error (or bias) is known at a given time t, then a \nm{1\sigma} bias stability of 0.01 [\nm{deg/hr}] over 100 [\nm{sec}] means that the bias at time \nm{t + 100} [\nm{sec}] is a random variable with mean the error at time t and standard deviation  0.01 [\nm{deg/hr}], and expression (\ref{eq:Sensor_SensorModel_sigmau}) can be used to obtain \nm{\sigma_u}. As the bias behaves as a random walk over time whose standard deviation grows proportionally to the square root of time, the bias stability is sometimes referred as a bias random walk.

In reality bias fluctuations do not really behave as a random walk. If they did, the uncertainty in the bias of a device would grow without bound as the timespan increased, which is not the case. In practice the bias is constrained to be within some range, and therefore the random walk model is only a good approximation to the true process for short periods of time \cite{Woodman2007}.


\subsection{Platform, Accelerometers, and Gyroscopes Frames}\label{subsec:Sensors_Inertial_Frames}

The following sections make use of three different reference frames to describe the readings of accelerometers and gyroscopes:
\begin{itemize} 

\item The \emph{platform frame} \nm{\FP} is a Cartesian reference system with its origin located at the IMU reference point (section \ref{subsec:Sensors_Inertial_Mounting}) and its three axes \nm{\lrb{\iPi,\,\iPii,\,\iPiii}} forming a right hand system that is loosely aligned with the aircraft body axes, so they point in the general directions of the aircraft fuselage (forward), aircraft wings (rightwards), and downward, respectively \cite{Farrell2008,Chatfield1997,Rogers2007}.

A proper definition of the platform frame is indispensable for navigation, as the calibrated outputs of the accelerometers and gyroscopes are based on it (sections \ref{subsec:Sensors_Accelerometer_Triad_ErrorModel} and \ref{subsec:Sensors_Gyroscope_Triad_ErrorModel}). The \nm{\FP} frame can be obtained from the body frame \nm{\FB} by a rotation best described by the Euler angles \nm{\phiBP = \lrsb{\psiP, \ \thetaP, \ \xiP}^T}\footnote{These Euler angles correspond to the 3 - 2 - 1 (yaw, pitch, roll) convention employed in aeronautics.} followed by a translation \nm{\TBPB} (section \ref{subsec:Sensors_Inertial_Mounting}) from the aircraft center of mass to the IMU reference point.

\item The \emph{accelerometers frame} \nm{\FA} is a non orthogonal reference system also centered at the IMU reference point \cite{Farrell2008,Chatfield1997,Rogers2007}. The basis vectors \nm{\lrb{\iAi,\,\iAii,\,\iAiii}} are aligned with each of the three accelerometers sensing axes\footnote{Each accelerometer hence only senses the specific force component parallel to its sensing axis.} (section \ref{subsec:Sensors_Accelerometer_Triad_ErrorModel}), but they are not orthogonal among them due to manufacturing inaccuracies. This implies that the angles between the \nm{\FA} and \nm{\FP} axes are very small.  

It is always possible, with no loss of generality, to impose that \nm{\iPi} coincides with \nm{\iAi} and that \nm{\iPii} is located in the plane defined by \nm{\iAi} and \nm{\iAii}. If this is the case, \nm{\iAi  \perp \iPii}, \nm{\iAi \perp \iPiii}, and \nm{\iAii \perp \iPiii}, and the relative attitude between the \nm{\FP} and \nm{\FA} frames can be defined by three independent small rotations. 
\begin{itemize}
\item The \nm{\iAii} axis can be obtained from \nm{\iPii} by means of a small rotation \nm{\alphaACCiii} about \nm{\iPiii}.
\item The \nm{\iAiii} axis can be obtained from \nm{\iPiii} by two small rotations: \nm{\alphaACCi} about \nm{\iPi} and \nm{\alphaACCii} about \nm{\iPii}.
\end{itemize}

Although the exact relationships can be obtained \cite{Chatfield1997}, and given that the angles are very small, it is possible to consider \nm{\cos \alphaACCXi = 1, \ \sin \alphaACCXi = \alphaACCXi}, and \nm{\alphaACCXi \cdot \alphaACCXj = 0 \ \forall \ i,j \in \lrb{1,\ 2, \ 3}, i \neq j}, resulting in the following transformations between free vectors viewed in the platform (\nm{\vP}) and accelerometer (\nm{\vA}) frames respectively\footnote{As \nm{\FA} is not orthogonal, the transformation matrices are denoted with \nm{\star} to indicate that they are not proper rotation matrices.}: 
\begin{eqnarray}
\nm{\vP = \RPA \ \vA} & = & \nm{\begin{bmatrix} 1 & 0 & 0 \\ \nm{\alphaACCiii} & 1 & 0 \\ - \nm{\alphaACCii} & \nm{\alphaACCi} & 1 \end{bmatrix} \ \vA}\label{eq:RefSystems_P_A} \\
\nm{\vA = \RAP \ \vP} & = & \nm{\begin{bmatrix} 1 & 0 & 0 \\ - \nm{\alphaACCiii} & 1 & 0 \\ \nm{\alphaACCii} & - \nm{\alphaACCi} & 1 \end{bmatrix} \ \vP}\label{eq:RefSystems_A_P}
\end{eqnarray}

\item The \emph{gyroscopes frame} \nm{\FY} is similar to the accelerometers frame \nm{\FA} defined above, but aligned with the gyroscopes sensing axes instead of those of the accelerometers  \cite{Farrell2008,Chatfield1997,Rogers2007}. It is also a non orthogonal reference system centered at the IMU reference point, but no simplifications can be made about the relative orientation of its axes \nm{\lrb{\iYi,\,\iYii,\,\iYiii}} with respect to those of \nm{\FP}, so their relative attitude is defined by six small rotations \nm{\alphaGYRXij \ \forall \ i,j \in \lrb{1,\ 2, \ 3}, i \neq j}, where \nm{\alphaGYRXij} is the rotation of \nm{\vec i_{i}^{\sss Y}} about \nm{\vec i_{j}^{\sss P}}.

An approach similar to that employed with accelerometers leads to the following transformations between free vectors viewed in the platform (\nm{\vP}) and gyroscope (\nm{\vY}) frames: 
\begin{eqnarray}
\nm{\vP = \RPY \ \vY} & = & \nm{\begin{bmatrix} 1 & -\nm{\alphaGYRiiXiii} & \nm{\alphaGYRiiiXii} \\ \nm{\alphaGYRiXiii} & 1 & -\nm{\alphaGYRiiiXi} \\ - \nm{\alphaGYRiXii} & \nm{\alphaGYRiiXi} & 1 \end{bmatrix} \ \vY}\label{eq:RefSystems_P_Y} \\
\nm{\vY = \RYP \ \vP} & = & \nm{\begin{bmatrix}  1 & \nm{\alphaGYRiiXiii} & -\nm{\alphaGYRiiiXii} \\ -\nm{\alphaGYRiXiii} & 1 & \nm{\alphaGYRiiiXi} \\ \nm{\alphaGYRiXii} & -\nm{\alphaGYRiiXi} & 1 \end{bmatrix} \ \vP}\label{eq:RefSystems_Y_P}
\end{eqnarray}
\end{itemize}


\subsection{Accelerometer Triad Sensor Error Model}\label{subsec:Sensors_Accelerometer_Triad_ErrorModel}

An IMU is equipped with an accelerometer triad composed by three individual accelerometers, each of which measures the projection of the specific force over its sensing axis as described in section \ref{subsec:Sensors_Inertial_ErrorModelSingleAxis} while incurring in an error \nm{e_{\sss BW,ACC}} that can be modeled as a combination of bias offset, bias drift, and white noise (\ref{eq:Sensor_SensorModel_error}). The three accelerometers can be considered infinitesimally small and located at the \emph{IMU reference point}, which is defined as the intersection between the sensing axes of the three accelerometers. As the accelerometer frame \nm{\FA} is centered at the IMU reference point and its three non orthogonal axes coincide with the accelerometers sensing axes, (\ref{eq:Sensor_Inertial_acc_error_FA}) joins together the measurements of the three individual accelerometers:
\neweq{\fIAAtilde = \SACC \ \lrp{\fIAA + \vec e_{\sss BW,ACC}^{\sss A}}}{eq:Sensor_Inertial_acc_error_FA}

where \nm{\fIAA} is the specific force viewed in the accelerometer frame \nm{\FA}, \nm{\fIAAtilde} represents its measurement also viewed in \nm{\FA}, \nm{\vec e_{\sss BW,ACC}^{\sss A}} is the error introduced by each accelerometer (\ref{eq:Sensor_SensorModel_error}), and \nm{\SACC} is a square diagonal matrix containing the scale factor errors \nm{\lrb{\sACCi, \ \sACCii, \ \sACCiii}} for each accelerometer (section \ref{subsec:Sensors_Inertial_ErrorSources}). It is however preferred to obtain an expression in which the specific forces are viewed in the orthogonal platform frame \nm{\FP} instead of the accelerometer frame \nm{\FA}. As both share the same origin,
\neweq{\fIPPtilde = \RPA \ \SACC \ \lrp{\RAP \ \fIPP + \vec e_{\sss BW,ACC}^{\sss A}}}{eq:Sensor_Inertial_acc_error_FP_bis}

where \nm{\RPA} and \nm{\RAP}, defined by (\ref{eq:RefSystems_P_A}) and (\ref{eq:RefSystems_A_P}), contain the cross coupling errors \nm{\lrb{\alphaACCi, \ \alphaACCii, \ \alphaACCiii}} generated by the misalignment of the accelerometer sensing axes. The scale factor and cross coupling errors contain fixed and temperature dependent error contributions (refer to section \ref{subsec:Sensors_Inertial_ErrorSources}) that can be modeled as normal random variables:  
\begin{eqnarray}
\nm{\sACCXi}     & = & \nm{N\lrp{1, \ \sACC^2}  \; \hspace{20pt} \forall \ i \in \lrb{1,\ 2, \ 3}}\label{eq:Sensors_Inertial_acc_error_scale} \\
\nm{\alphaACCXi} & = & \nm{N\lrp{0, \ \alphaACC^2} \hspace{20pt} \forall \ i \in \lrb{1,\ 2, \ 3}}\label{eq:Sensors_Inertial_acc_error_cross}
\end{eqnarray}

where \nm{\sACC} and \nm{\alphaACC} can be obtained from the sensor specifications. Equation (\ref{eq:Sensor_Inertial_acc_error_FP_bis}) can be transformed to make it more useful by defining the accelerometer scale and cross coupling error matrix \nm{\MACC}:
\begin{eqnarray}
\nm{\MACC} & = & \nm{\RPA \ \SACC \ \RAP = \begin{bmatrix} \nm{m_{\sss{ACC,11}}} & 0 & 0 \\ \nm{m_{\sss{ACC,21}}} & \nm{m_{\sss{ACC,22}}} & 0 \\ \nm{m_{\sss{ACC,31}}} & \nm{m_{\sss{ACC,32}}} & \nm{m_{\sss{ACC,33}}} \end{bmatrix}}\nonumber \\
& \nm{\approx} & \nm{\begin{bmatrix} \nm{\sACCi} & 0 & 0 \\ \nm{\alphaACCiii \ \lrp{\sACCi - \sACCii}} & \nm{\sACCii} & 0 \\ \nm{\alphaACCii \ \lrp{\sACCiii - \sACCi}} & \nm{\alphaACCi \ \lrp{\sACCii - \sACCiii}} & \nm{\sACCiii} \end{bmatrix}}\label{eq:Sensors_Inertial_acc_M}
\end{eqnarray}

Considering that the scale and cross coupling errors are uncorrelated and very small, and taking into account the expressions for the mean and variance of the sum and product of two random variables \cite{Frishman1971}, the different components \nm{\mACCXij} of \nm{\MACC} can be obtained as follows \nm{\forall \ i, \ j \in \lrb{1,\ 2, \ 3}}:
\begin{eqnarray}
\nm{\mACCXij} & = & \nm{N\lrp{1, \ \sACC^2}                                                   \hspace{146pt} i = j}\label{eq:Sensors_Inertial_acc_error_scale_cross1} \\
\nm{\mACCXij} & = & \nm{N\lrp{0, \ \lrsb{\sqrt{2} \ \alphaACC \ \sACC}^2} = N\lrp{0, \mACC^2} \hspace{20pt} i > j}\label{eq:Sensors_Inertial_acc_error_scale_cross2} \\
\nm{\mACCXij} & = & \nm{0                                                                     \hspace{191pt} i < j}\label{eq:Sensors_Inertial_acc_error_scale_cross3}
\end{eqnarray}

Let's also define the accelerometer error transformation matrix \nm{\NACC} as
\neweq{\NACC = \RPA \ \SACC = \begin{bmatrix} \nm{n_{\sss{ACC,11}}} & 0 & 0 \\ \nm{n_{\sss{ACC,21}}} & \nm{n_{\sss{ACC,22}}} & 0 \\ \nm{n_{\sss{ACC,31}}} & \nm{n_{\sss{ACC,32}}} & \nm{n_{\sss{ACC,33}}} \end{bmatrix} = \begin{bmatrix} \nm{\sACCi} & 0 & 0 \\ \nm{\alphaACCiii \ \sACCi} & \nm{\sACCii} & 0 \\ - \nm{\alphaACCii \ \sACCi} & \nm{\alphaACCi \ \sACCii} & \nm{\sACCiii} \end{bmatrix}}{eq:Sensors_Inertial_acc_N}

A process similar to that employed above leads to:
\begin{eqnarray}
\nm{\nACCXij} & = & \nm{N\lrp{1, \ \sACC^2}                                                     \hspace{132pt} i = j}\label{eq:Sensors_Inertial_acc_errorN_scale_cross1} \\
\nm{\nACCXij} & = & \nm{N\lrp{0, \ \alphaACC^2 \lrp{1 + \sACC^2}} \approx N\lrp{0, \alphaACC^2} \hspace{20pt} i > j}\label{eq:Sensors_Inertial_acc_errorN_scale_cross2} \\
\nm{\nACCXij} & = & \nm{0                                                                       \hspace{178pt} i < j}\label{eq:Sensors_Inertial_acc_errorN_scale_cross3}
\end{eqnarray}

Taking into account the expressions for the mean and variance of the sum and product of two random variables \cite{Frishman1971}, and knowing that the cross coupling errors are very small \nm{\lrp{1 + \alphaACC^2 \approx 1}}, it can be proven that the bias and white noise errors viewed in the platform frame \nm{\FP} respond to a expression similar to (\ref{eq:Sensor_SensorModel_error}):
\begin{eqnarray}
\nm{\vec e_{\sss BW,ACC}^{\sss P}} & = & \nm{\vec e_{\sss BW,ACC}^{\sss P} \lrp{s \, \DeltatSENSED} = \vec e_{\sss BW,ACC}^{\sss P} \lrp{s \, \Deltat}} \nonumber \\
& = & \nm{\NACC \ \vec e_{\sss BW,ACC}^{\sss A} = \BzeroACC \, \NuzeroACC + \sigmauACC \, \Deltat^{1/2} \, \sum_{i=1}^s \NuiACC + \dfrac{\sigmavACC}{\Deltat^{1/2}} \, \NvsACC} \label{eq:Sensors_Inertia_acc_error}
\end{eqnarray}

where each \nm{\NuACC} and \nm{\NvACC} is a random vector composed by three independent standard normal random variables \nm{N\lrp{0, \, 1}}. Note that as the bias drift is mostly a warm up process that stabilizes itself after a few minutes of operation, the random walk within (\ref{eq:Sensors_Inertia_acc_error}) is not allowed to vary freely but is restricted to within a band of width \nm{\pm \ 100 \, \sigmauACC \, \Deltat^{1/2}}. The final model for the accelerometer measurements viewed in \nm{\FP} results in
\neweq{\fIPPtilde = \MACC \ \fIPP + \vec e_{\sss BW,ACC}^{\sss P}}{eq:Sensor_Inertial_acc_error_FP}

where \nm{\MACC} is described in (\ref{eq:Sensors_Inertial_acc_M}) through (\ref{eq:Sensors_Inertial_acc_error_scale_cross3}) and \nm{\vec e_{\sss BW,ACC}^{\sss P}} is provided by (\ref{eq:Sensors_Inertia_acc_error}). This model relies on inputs for the bias offset \nm{\BzeroACC}, bias drift \nm{\sigmauACC}, white noise \nm{\sigmavACC}, scale factor error \nm{\sACC}, and cross coupling error \nm{\mACC}, which are obtained from the accelerometer specifications shown in table \ref{tab:Sensors_acc} within section \ref{subsec:Sensors_Inertial_Selected_gyr_acc}.


\subsection{Gyroscopes Triad Sensor Error Model}\label{subsec:Sensors_Gyroscope_Triad_ErrorModel}

The IMU is also equipped with a triad of gyroscopes, each of which measures the projection of the inertial angular velocity over its sensing axis as described in section \ref{subsec:Sensors_Inertial_ErrorModelSingleAxis}. The obtainment of the gyroscope triad model is fully analogous to that of the accelerometers in the previous section, with the added difficulty that the transformation between the gyroscope frame \nm{\FY} and platform frame \nm{\FP} relies on six small angles instead of three. The starting point hence is:
\neweq{\wIPPtilde = \RPY \ \SGYR \ \lrp{\RYP \ \wIPP + \vec e_{\sss BW,GYR}^{\sss Y}}}{eq:Sensor_Inertial_gyr_error_FP_bis}

where \nm{\wIPP} is the inertial angular velocity viewed in the platform frame \nm{\FP}, \nm{\wIPPtilde} represents its measurement also viewed in \nm{\FP}, \nm{\vec e_{\sss BW,GYR}^{\sss Y}} is the error introduced by each gyroscope (\ref{eq:Sensor_SensorModel_error}), \nm{\SGYR} is a square diagonal matrix containing the scale factor errors \nm{\lrb{\sGYRi, \ \sGYRii, \ \sGYRii}}, and \nm{\RPY} and \nm{\RYP}, defined by (\ref{eq:RefSystems_P_Y}) and (\ref{eq:RefSystems_Y_P}), contain the cross coupling errors \nm{\alphaGYRiXii, \ \alphaGYRiiXi, \ \alphaGYRiXiii, \ \alphaGYRiiiXi, \ \alphaGYRiiXiii, \ \alphaGYRiiiXii} generated by the misalignment of the gyroscope sensing axes.

Operating in the same way as in section \ref{subsec:Sensors_Accelerometer_Triad_ErrorModel} leads to:
\begin{eqnarray}
\nm{\vec e_{\sss BW,GYR}^{\sss P}} & = & \nm{\vec e_{\sss BW,GYR}^{\sss P} \lrp{s \, \DeltatSENSED} = \vec e_{\sss BW,GYR}^{\sss P} \lrp{s \, \Deltat}}\nonumber \\
& = & \nm{\BzeroGYR \, \NuzeroGYR + \sigmauGYR \, \Deltat^{1/2} \, \sum_{i=1}^s \NuiGYR + \dfrac{\sigmavGYR}{\Deltat^{1/2}} \, \NvsGYR}\label{eq:Sensors_Inertia_gyr_error} \\
\nm{\wIPPtilde} & = & \nm{\MGYR \ \wIPP + \vec e_{\sss BW,GYR}^{\sss P}}\label{eq:Sensor_Inertial_gyr_error_FP}
\end{eqnarray}

where each \nm{\NuiGYR} and \nm{\NvGYR} is a random vector composed by three independent standard normal random variables \nm{N\lrp{0, \, 1}}. As in the case of the accelerometers, the random walk within (\ref{eq:Sensors_Inertia_gyr_error}) representing the bias drift is not allowed to vary freely but is restricted to within a band of width \nm{\pm \ 100 \, \sigmauGYR \, \Deltat^{1/2}}. This model relies on inputs for the bias offset \nm{\BzeroGYR}, bias drift \nm{\sigmauGYR}, white noise \nm{\sigmavGYR}, scale factor error \nm{\sGYR}, and cross coupling error \nm{\mGYR}, which are obtained from the gyroscope specifications shown in table \ref{tab:Sensors_gyr} within section \ref{subsec:Sensors_Inertial_Selected_gyr_acc}. The gyroscope scale and cross coupling error matrix \nm{\MGYR} responds to:
\begin{eqnarray}
\nm{\MGYR} & = & \nm{\RPY \ \SGYR \ \RYP = \begin{bmatrix} \nm{m_{\sss{GYR,11}}} & \nm{m_{\sss{GYR,12}}} & \nm{m_{\sss{GYR,13}}} \\ \nm{m_{\sss{GYR,21}}} & \nm{m_{\sss{GYR,22}}} & \nm{m_{\sss{GYR,23}}} \\ \nm{m_{\sss{GYR,31}}} & \nm{m_{\sss{GYR,32}}} & \nm{m_{\sss{GYR,33}}} \end{bmatrix}}\nonumber \\
& \nm{\approx} & \nm{\begin{bmatrix} \nm{\sGYRi} & \nm{\alphaGYRiiXiii \ \lrp{\sGYRi - \sGYRii}} & \nm{\alphaGYRiiiXii \ \lrp{\sGYRiii - \sGYRi}} \\ \nm{\alphaGYRiXiii \ \lrp{\sGYRi - \sGYRii}} & \nm{\sGYRii} & \nm{\alphaGYRiiiXi \ \lrp{\sGYRii - \sGYRiii}} \\ \nm{\alphaGYRiXii \ \lrp{\sGYRiii - \sGYRi}} & \nm{\alphaGYRiiXi \ \lrp{\sGYRii - \sGYRiii}}  & \nm{\sGYRiii} \end{bmatrix}}\label{eq:Sensors_Inertial_gyr_M} \\
\nm{\mGYRXij} & = & \nm{N\lrp{1, \ \sGYR^2}                                         \hspace{122pt} i = j}  \label{eq:Sensors_Inertial_gyr_error_scale_cross1} \\
\nm{\mGYRXij} & = & \nm{N\lrp{0, \ 2 \ \alphaGYR^2 \ \sGYR^2} = N\lrp{0, \ \mGYR^2} \hspace{20pt} i \neq j}\label{eq:Sensors_Inertial_gyr_error_scale_cross2}
\end{eqnarray}


\subsection{Mounting of Inertial Sensors}\label{subsec:Sensors_Inertial_Mounting}

Equations (\ref{eq:Sensor_Inertial_acc_error_FP}) and (\ref{eq:Sensor_Inertial_gyr_error_FP}) contain the relationships between the specific force \nm{\fIPP} and inertial angular velocity \nm{\wIPP} and their measurements \nm{\lrp{\fIPPtilde, \ \nm{\wIPPtilde}}}, when evaluated and viewed in the platform frame \nm{\FP}. However, from the point of view of the navigation system, both magnitudes need to be evaluated and viewed in the body frame \nm{\FB} instead of \nm{\FP}. These equations thus need to be modified so they relate \nm{\fIBB} with \nm{\fIBBtilde} as well as \nm{\wIBB} with \nm{\wIBBtilde}, respectively, as described in section \ref{subsec:Sensors_Inertial_ErrorModel} below. To do that, it is necessary to define the relative pose (position plus attitude) between the \nm{\FP} and \nm{\FB} frames. Note that the IMU, represented by the platform frame \nm{\FP}, should be mounted in the aircraft as close as possible to the center of gravity (this reduces errors, as described in section \ref{subsec:Sensors_Inertial_ErrorModel}), and loosely aligned with the aircraft body axes.

To provide realism to the stochastic simulation, this article assumes that the real displacement \nm{\TBPB} between the two frames is deterministic, while the relative rotation \nm{\phiBP = \lrsb{\psiP, \ \thetaP, \ \xiP}^T} is stochastic. In this way, each simulated aircraft represented by its specific aircraft seed \nm{\seedA} exhibits a slightly different IMU platform attitude with respect to the aircraft body:
\begin{itemize}

\item As the IMU reference point is fixed with respect to the structure but the aircraft center of mass slowly moves as the fuel load diminishes, it is possible to establish a linear model that provides the displacement between the origins of both frames according to the aircraft mass\footnote{The aircraft masses \nm{m_{full}} and \nm{m_{empty}} when the fuel tank is fully loaded or empty are inputs, as are the displacements between the IMU reference point and the aircraft center of mass \nm{\TBPBfull} and \nm{\TBPBempty}.}:
\neweq{\TBPB = \vec f\lrp{m} = \TBPBfull + \dfrac{m_{full} - m}{m_{full} - m_{empty}} \; \lrp{\TBPBempty - \TBPBfull}} {eq:Sensors_Inertial_Mounting_Tbpb}

\item The platform Euler angles respond to the stochastic model provided by (\ref{eq:Sensors_Inertial_Mounting_eulerBP}), in which each specific Euler angle is obtained as the product of the standard deviations (\nm{\sigmapsiP}, \nm{\sigmathetaP}, \nm{\sigmaxiP}) by a single realization of a standard normal random variable \nm{N\lrp{0, \, 1}} (\nm{\NpsiP}, \nm{\NthetaP}, and \nm{\NxiP}). 
\neweq{\phiBP = \lrsb{\sigmapsiP \, \NpsiP, \ \sigmathetaP \, \NthetaP, \ \sigmaxiP \NxiP}^T}{eq:Sensors_Inertial_Mounting_eulerBP}
\end{itemize}
\begin{center}
\begin{tabular}{lccc}
	\hline
	\textbf{Concept} & \textbf{Variable} & \textbf{Value} & \textbf{Unit} \\
	\hline
	True Yaw Error 	   	               & \nm{\sigmapsiP}     & \nm{0.5}  & [\nm{deg}]  \\
	True Pitch Error                   & \nm{\sigmathetaP}   & \nm{2.0}  & [\nm{deg}]  \\
	True Bank Error 		           & \nm{\sigmaxiP}      & \nm{0.1}  & [\nm{deg}]  \\
	Platform Position Estimation Error & \nm{\sigmaTBPBest}  & \nm{0.01} & [\nm{m}] \\
	Platform Angular Estimation Error  & \nm{\sigmaphiBPest}	& \nm{0.03} & [\nm{deg}]  \\
	\hline
\end{tabular}
\end{center}
\captionof{table}{IMU mounting accuracy values}\label{tab:Sensor_Inertial_mounting}

Once the real pose between the \nm{\FP} and \nm{\FB} frames is established, it is necessary to specify its estimation employed by the IMU in the comprehensive model introduced in section \ref{subsec:Sensors_Inertial_ErrorModel}, which is discussed in section \ref{subsec:Platform_frame}. Stochastic models are employed in the simulation for both the translation \nm{\TBPBest} and rotation \nm{\phiBPest}, changing their values from one aircraft seed \nm{\seedA} to the next:
\begin{eqnarray}
\nm{\TBPBest} & = & \nm{\TBPB + \lrsb{\sigmaTBPBest \, \NTBPBiest, \ \sigmaTBPBest \, \NTBPBiiest, \ \sigmaTBPBest \NTBPBiiiest}^T} \label{eq:Sensors_Inertial_Mounting_TBPBest} \\
\nm{\phiBPest} & = & \nm{\phiBP + \lrsb{\sigmaphiBPest \, \NpsiPest, \ \sigmaphiBPest \, \NthetaPest, \ \sigmaphiBPest \NxiPest}^T} \label{eq:Sensors_Inertial_Mounting_eulerBPest}
\end{eqnarray}

where the default standard deviations \nm{\sigmaTBPBest} and \nm{ \sigmaphiBPest} are shown in table \ref{tab:Sensor_Inertial_mounting}, and \nm{\NpsiPest}, \nm{\NthetaPest}, \nm{\NxiPest}, \nm{\NTBPBiest}, \nm{\NTBPBiiest}, \nm{\NTBPBiiiest} are six realizations of a standard normal random variable \nm{N\lrp{0, \, 1}}. Note that table \ref{tab:Sensor_Inertial_mounting} lists the default values employed in the simulation, which can be adjusted by the user.

\nm{\TBP} can be considered quasi stationary as it slowly varies based on the aircraft mass, and the relative position of their axes \nm{\phiBP} remains constant because the IMU is rigidly attached to the aircraft structure. Although Euler angles have been employed in this section, from this point on it is more practical to employ the rotation matrix \nm{\RBP} to represent the rotation between two different frames \cite{Shuster1993}. The time derivatives of \nm{\TBP} and \nm{\RBP} are hence zero:
\neweq{\dot{\vec T}_{\sss BP} = \dot{\vec R}_{\sss BP} = \vec 0 \ \longrightarrow \ \vBP = \vec a_{\sss BP} = \wBP = \vec \alpha_{\sss BP} = \vec 0}{eq:Sensor_Inertial_BP_diff}
  

\subsection{Comprehensive Inertial Sensor Error Model}\label{subsec:Sensors_Inertial_ErrorModel}

Two considerations are required to establish the measurement equations for the inertial sensors viewed in the body frame \nm{\FB}. First, let's apply the composition rules of section \ref{subsec:MOT_Composition_Summary} considering \nm{\FI} as \nm{F_{\sss0}}, \nm{\FB} as \nm{F_{\sss1}}, and \nm{\FP} as \nm{F_{\sss2}}, which results in:
\begin{eqnarray}
\nm{\wIP} & = & \nm{\wIB}\label{eq:Sensor_Inertial_BP_omega} \\
\nm{\vec \alpha_{\sss IP}} & = & \nm{\vec \alpha_{\sss IB}}\label{eq:Sensor_Inertial_BP_alpha} \\
\nm{\vIP} & = & \nm{\vIB + \wIBskew \ \TBP}\label{eq:Sensor_Inertial_BP_v} \\
\nm{\vec a_{\sss IP}} & = & \nm{\vec a_{\sss IB} + \alphaIBskew \ \TBP + \wIBskew \ \wIBskew \ \TBP}\label{eq:Sensor_Inertial_BP_a}
\end{eqnarray}

Second, it is also necessary to consider that as \nm{\RBPest} is a rotation matrix in which all rows and columns are unitary vectors, the projection of the \nm{\FP} frame bias and white noise errors \nm{\vec e_{\sss BW,ACC}^{\sss P}} and \nm{\vec e_{\sss BW,GYR}^{\sss P}} onto the \nm{\FB} frame does not change their stochastic properties:
\begin{eqnarray}
\nm{\vec e_{\sss BW,ACC}^{\sss B} \lrp{s \, \Deltat}} & = & \nm{\RBPest \ \vec e_{\sss BW,ACC}^{\sss P} = \BzeroACC \, \NuzeroACC + \sigmauACC \, \Deltat^{1/2} \, \sum_{i=1}^s \NuiACC + \dfrac{\sigmavACC}{\Deltat^{1/2}} \, \NvsACC}\label{eq:Sensors_Inertia_acc_bw_error} \\
\nm{\vec e_{\sss BW,GYR}^{\sss B} \lrp{s \, \Deltat}} & = & \nm{\RBPest \ \vec e_{\sss BW,GYR}^{\sss P} = \BzeroGYR \, \NuzeroGYR + \sigmauGYR \, \Deltat^{1/2} \, \sum_{i=1}^s \NuiGYR + \dfrac{\sigmavGYR}{\Deltat^{1/2}} \, \NvsGYR}\label{eq:Sensors_Inertia_gyr_bw_error}
\end{eqnarray}

As the inertial angular velocity does not change when evaluated in the \nm{\FB} and \nm{\FP} frames (\ref{eq:Sensor_Inertial_BP_omega}), its measurement in the body frame can be derived from (\ref{eq:Sensor_Inertial_gyr_error_FP}) by first projecting it from \nm{\FB} to \nm{\FP} based on the real rotation matrix \nm{\RBP} and then projecting back the measurement into \nm{\FB} based on the estimated rotation matrix \nm{\RBPest}. The bias and white noise error is also projected according to (\ref{eq:Sensors_Inertia_gyr_bw_error}):
\neweq{\wIBBtilde = \RBPest \ \MGYR \ {\RBP}^T \ \wIBB + \vec e_{\sss BW,GYR}^{\sss B}}{eq:Sensor_Inertial_gyr_error_final}

The expression for the specific force measurement is significantly more complex because the back and forth transformations of the specific force between the \nm{\FB} and \nm{\FP} frames need to consider the influence of the lever arm \nm{\TBP}, as indicated in (\ref{eq:Sensor_Inertial_BP_a}). The additional terms introduce errors in the measurements, so as indicated in section \ref{subsec:Sensors_Inertial_Mounting} it is desirable to locate the IMU as close as possible to the aircraft center of mass.
\neweq{\fIBBtilde = \RBPest \ \MACC \ {\RBP}^T \ \lrp{\fIBB + \alphaIBBskew \ \TBPB + \wIBBskew \ \wIBBskew \ \TBPB} - \alphaIBBestskew \ \TBPBest - \wIBBestskew \ \wIBBestskew \ \TBPBest + \vec e_{\sss BW,ACC}^{\sss B}}{eq:Sensor_Inertial_acc_error_prefinal}

Note that this expression can not be directly evaluated as the estimated values for the inertial angular velocity and acceleration (\nm{\wIBBest, \ \alphaIBBest}) are unknown by the IMU until obtained by the navigation filter. The IMU can however rely on the gyroscope readings, directly replacing \nm{\wIBBest} with \nm{\wIBBtilde} and computing \nm{\alphaIBBtilde} based on the difference between the present and previous \nm{\wIBBtilde} readings, resulting in:
\neweq{\fIBBtilde = \RBPest \ \MACC \ {\RBP}^T \ \lrp{\fIBB + \alphaIBBskew \ \TBPB + \wIBBskew \ \wIBBskew \ \TBPB} - \alphaIBBtildeskew \ \TBPBest - \wIBBtildeskew \ \wIBBtildeskew \ \TBPBest + \vec e_{\sss BW,ACC}^{\sss B}}{eq:Sensor_Inertial_acc_error_final}

Table \ref{tab:Sensors_Inertial_error_sources} lists the error sources contained in the comprehensive inertial sensor error model represented by (\ref{eq:Sensor_Inertial_gyr_error_final}, \ref{eq:Sensor_Inertial_acc_error_final}). The first two columns list the different error sources, while the third column specifies their origin according to the criterion established in the first paragraph of section \ref{subsec:Sensors_Inertial_ErrorSources}. The section where each error is described appears on the fourth column, followed by the seeds employed to ensure the results variability for different aircraft (\nm{\seedA}) as well as different flights (\nm{\seedF}):
\begin{center}
\begin{tabular}{lccccc}
	\hline
	\multicolumn{2}{c}{\textbf{Error}} & \textbf{Main Source} & \textbf{Description} & \multicolumn{2}{c}{\textbf{Seeds}} \\
	\hline
	Bias Offset			& \nm{\BzeroACC, \ \BzeroGYR}	& run-to-run	& \ref{subsec:Sensors_Inertial_ErrorModelSingleAxis}												 	& \nm{\seedFACC}, \nm{\seedFGYR} & \nm{\seedF} \\
	Bias Drift			& \nm{\sigmauACC, \ \sigmauGYR} & in-run		& \ref{subsec:Sensors_Inertial_ErrorModelSingleAxis}													& \nm{\seedFACC}, \nm{\seedFGYR} & \nm{\seedF} \\
	System Noise		& \nm{\sigmavACC, \ \sigmavGYR}	& in-run		& \ref{subsec:Sensors_Inertial_ErrorModelSingleAxis}													& \nm{\seedFACC}, \nm{\seedFGYR} & \nm{\seedF} \\
	Scale Factor		& \nm{\sACC, \ \sGYR}			& fixed \& T	& \ref{subsec:Sensors_Accelerometer_Triad_ErrorModel}, \ref{subsec:Sensors_Gyroscope_Triad_ErrorModel}	& \nm{\seedAACC}, \nm{\seedAGYR} & \nm{\seedA} \\
	Cross Coupling		& \nm{\mACC, \ \mGYR}			& fixed			& \ref{subsec:Sensors_Accelerometer_Triad_ErrorModel}, \ref{subsec:Sensors_Gyroscope_Triad_ErrorModel}	& \nm{\seedAACC}, \nm{\seedAGYR} & \nm{\seedA} \\
	Lever Arm			& \nm{\TBP, \ \sigmaTBPBest}	& fixed			& \ref{subsec:Sensors_Inertial_Mounting}																& \nm{\seedAPLAT}                & \nm{\seedA} \\
	IMU Attitude		& \nm{\sigmapsiP, \ \sigmathetaP, \ \sigmaxiP, \ \sigmaphiBPest} & fixed & \ref{subsec:Sensors_Inertial_Mounting}										& \nm{\seedAPLAT}                & \nm{\seedA} \\
	\hline
\end{tabular}
\end{center}
\captionof{table}{Inertial sensor error sources}\label{tab:Sensors_Inertial_error_sources}

Note that \nm{\BzeroACC}, \nm{\BzeroGYR}, \nm{\sigmauACC}, \nm{\sigmauGYR}, \nm{\sigmavACC}, \nm{\sigmavGYR}, \nm{\sACC}, \nm{\sGYR}, \nm{\mACC}, and \nm{\mGYR} are taken from the right hand column of tables \ref{tab:Sensors_gyr} and \ref{tab:Sensors_acc} (section \ref{subsec:Sensors_Inertial_Selected_gyr_acc}), while the remaining error sources come from section \ref{subsec:Sensors_Inertial_Mounting}. It is worth pointing out that all errors are modeled as stochastic variables or processes (with the exception of the \nm{\TBP} displacement between the body and platform frames, which is deterministic), as expressions (\ref{eq:Sensor_Inertial_gyr_error_final}, \ref{eq:Sensor_Inertial_acc_error_final}) rely on the errors provided by (\ref{eq:Sensors_Inertia_acc_bw_error}, \ref{eq:Sensors_Inertia_gyr_bw_error}), the scale and cross coupling matrices given by (\ref{eq:Sensors_Inertial_acc_M}, \ref{eq:Sensors_Inertial_gyr_M}), and the transformations given by (\ref{eq:Sensors_Inertial_Mounting_Tbpb}, \ref{eq:Sensors_Inertial_Mounting_eulerBP}, \ref{eq:Sensors_Inertial_Mounting_TBPBest}, \ref{eq:Sensors_Inertial_Mounting_eulerBPest}).

In the case of the accelerometer triad, the stochastic nature of the fixed and run-to-run error contributions to the model relies on three realizations of normal distributions for the bias offset, three for the scale factor errors, three for the cross coupling errors, and nine for the mounting errors, while the in-run error contributions require three realizations each for the bias drift and system noise at every discrete sensor measurement. The gyroscope triad is similar but requires six realizations to model the cross coupling errors instead of three, while using the same six realizations that the accelerometer triad to model the true and estimated rotation between the \nm{\FB} and \nm{\FP} frames.

Expressions (\ref{eq:Sensor_Inertial_gyr_error_final}, \ref{eq:Sensor_Inertial_acc_error_final}) can be rewritten to show the measurements as functions of the full errors (\nm{\EACC, \, \EGYR}), which represent all the errors introduced by the inertial sensors with the exception of white noise. 
\begin{eqnarray}
\nm{\fIBBtilde \lrp{s \, \Deltat}} & = & \nm{\fIBB \lrp{s \, \Deltat} + \EACC \lrp{s \, \Deltat} + \dfrac{\sigmavACC}{\Deltat^{1/2}} \, \NvsACC} \label{eq:Sensor_Inertial_acc_error_filter} \\
\nm{\wIBBtilde \lrp{s \, \Deltat}} & = & \nm{\wIBB \lrp{s \, \Deltat} + \EGYR \lrp{s \, \Deltat} + \dfrac{\sigmavGYR}{\Deltat^{1/2}} \, \NvsGYR} \label{eq:Sensor_Inertial_gyr_error_filter}
\end{eqnarray}


\subsection{Baseline Inertial Sensors}\label{subsec:Sensors_Inertial_Selected_gyr_acc}

The value of all parameters present in the previous models can be specified by the user when employing the simulation \cite{Gallo2021_sensors, SIMULATION}. This article however provides default values that represent the inertial sensors that could nowadays be installed onboard a small size fixed wing low SWaP autonomous aircraft, and which are referred to as the baseline configuration. In the case of the gyroscopes, they correspond to the MEMS gyroscopes installed inside the Analog Devices ADIS16488A IMU \cite{ADIS16488A}. Table \ref{tab:Sensors_gyr} shows its performances, which have been taken from the data sheet when possible, and corrected when suspicious. A calibration process as that described in section \ref{subsec:Inertial_Calibration} is assumed to eliminate \nm{95\%} of the scale factor and cross coupling errors.
\begin{center}
\begin{tabular}{lrc|lllc}
	\hline
	\textbf{GYR Baseline} & \multicolumn{1}{c}{\textbf{Spec}} & \textbf{Unit} & \multicolumn{1}{c}{\textbf{Variable}} & \multicolumn{1}{c}{\textbf{Value}} & \multicolumn{1}{c}{\textbf{Calibration}} & \textbf{Unit} \\
	\hline
	In-Run Bias Stability (1 \nm{\sigma})	& 5.10			& [\nm{deg/hr}]			& \nm{\sigmauGYR}	& \nm{1.42 \cdot 10^{-4}}	& \nm{1.42 \cdot 10^{-4}}	& [\nm{deg/sec^{1.5}}] \\
	Angle Random Walk (1 \nm{\sigma})		& 0.26			& [\nm{deg/hr^{0.5}}]	& \nm{\sigmavGYR}	& \nm{4.30 \cdot 10^{-3}}	& \nm{4.30 \cdot 10^{-3}}	& [\nm{deg/sec^{0.5}}] \\
	Nonlinearity\footnotemark				& 0.01			& [\%]		 			& \nm{\sGYR}		& \nm{3.00 \cdot 10^{-4}}	& \nm{1.50 \cdot 10^{-5}}	& [-] \\
	Misalignment							& \nm{\pm} 0.05	& [\nm{deg}]			& \nm{\mGYR}		& \nm{8.70 \cdot 10^{-4}}	& \nm{4.35 \cdot 10^{-5}}	& [-] \\
	Bias Repeatability (1 \nm{\sigma})		& \nm{\pm} 0.2	& [\nm{deg/sec}]		& \nm{\BzeroGYR}	& \nm{2.00 \cdot 10^{-1}}	& \nm{2.00 \cdot 10^{-1}}	& [\nm{deg/sec}] \\
	\hline
\end{tabular}
\captionof{table}{Performance of ``Baseline'' gyroscopes} \label{tab:Sensors_gyr}
\end{center}
\footnotetext{The \nm{0.01\%} scale factor error obtained in \cite{ADIS16488A} is considered too optimistic and hence modified to \nm{0.03\% = 3.00 \cdot 10^{-4} \ [-]}.}

The previous table show the performances in three columns of data. The left most column ("Spec") corresponds to data taken directly from the sensors specifications, which get converted in the middle column ("Value") to the parameters shown in sections \ref{subsec:Sensors_Inertial_ErrorModelSingleAxis} through \ref{subsec:Sensors_Inertial_ErrorModel}\footnote{The conversion between bias instability and \nm{\sigma_u} uses a period of 100 [\nm{sec}], as noted in section \ref{subsec:Sensors_Inertial_ObtainmentBiasDrift}.}. The right column ("Calibration") contains the specifications employed in the simulation after the calibration process, which reduces the scale factor and cross coupling errors by \nm{95\%}.
\begin{center}
\begin{tabular}{lrc|lllc}
	\hline
	\textbf{ACC Baseline} & \multicolumn{1}{c}{\textbf{Spec}} & \textbf{Unit} & \multicolumn{1}{c}{\textbf{Variable}} & \multicolumn{1}{c}{\textbf{Value}} & \multicolumn{1}{c}{\textbf{Calibration}} & \textbf{Unit} \\
	\hline
	In-Run Bias Stability (1 \nm{\sigma})	& 0.07 				& [\nm{mg}]				& \nm{\sigmauACC}	& \nm{6.86 \cdot 10^{-5}}	& \nm{6.86 \cdot 10^{-5}}	& [\nm{m/sec^{2.5}}] \\
	Velocity Random Walk (1 \nm{\sigma})	& 0.029				& [\nm{m/sec/hr^{0.5}}]	& \nm{\sigmavACC}	& \nm{4.83 \cdot 10^{-4}}	& \nm{4.83 \cdot 10^{-4}}	& [\nm{m/sec^{1.5}}] \\
	Nonlinearity							& 0.1				& [\%]		 			& \nm{\sACC}		& \nm{1.00 \cdot 10^{-3}}   & \nm{5.00 \cdot 10^{-5}}	& [-] \\
	Misalignment							& \nm{\pm} 0.035	& [\nm{deg}]			& \nm{\mACC}		& \nm{6.11 \cdot 10^{-4}}	& \nm{3.05 \cdot 10^{-5}}	& [-] \\
	Bias Repeatability (1 \nm{\sigma})		& \nm{\pm} 16		& [\nm{mg}]				& \nm{\BzeroACC}	& \nm{1.57 \cdot 10^{-1}}	& \nm{1.57 \cdot 10^{-1}}	& [\nm{m/sec^2}] \\
	\hline
\end{tabular}
\captionof{table}{Performance of ``Baseline'' accelerometers} \label{tab:Sensors_acc}
\end{center}

The baseline accelerometers also correspond to the MEMS accelerometers installed inside the Analog Devices ADIS16488A IMU \cite{ADIS16488A}. All the specifications shown in table \ref{tab:Sensors_acc} have been taken from the data sheet. As in the case of the gyroscopes, a calibration process as that described in section \ref{subsec:Inertial_Calibration} is assumed to eliminate \nm{95\%} of the scale factor and cross coupling errors\footnote{The conversion between bias instability and \nm{\sigma_u} uses a period of 100 [\nm{sec}], as noted in section \ref{subsec:Sensors_Inertial_ObtainmentBiasDrift}.}.


\section{Non Inertial Sensors}\label{sec:Sensors_NonInertial}

This section describes the different non-inertial sensors usually installed onboard a fixed wind autonomous aircraft, such as a triad of \emph{magnetometers} to measure the Earth magnetic field, a \emph{GNSS receiver} that provides absolute position and velocity measurements, and the \emph{air data system}, which in addition to the pressure altitude and temperature also provides a measurement of the airspeed and the airflow angles. 


\subsection{Magnetometers} \label{subsec:Sensors_NonInertial_Magnetometers}

Magnetometers measure magnetic field intensity along a given direction and are very useful for estimating the aircraft heading. Although other types exist, magnetoinductive and magnetoresistive sensors are generally employed for navigation due to their accuracy and small size \cite{Groves2008,Titterton2004}. As with the inertial sensors, three orthogonal magnetometers are usually employed in a strapdown configuration to measure the magnetic field with respect to the body frame \nm{\FB}.

Unfortunately magnetometers do not only measure the Earth magnetic field \nm{\Bvec}, but also that generated by the aircraft permanent magnets and electrical equipment (known as hard iron magnetism), as well as the magnetic field disturbances generated by the aircraft ferrous materials (soft iron magnetism). For that reason, the magnetometers should be placed in a location inside the aircraft that minimizes these errors. On the positive side, magnetometers do not exhibit the bias instability present in inertial sensors, and the error of an individual sensor can be properly modeled by the combination of bias offset and white noise. A triad of magnetometers capable of measuring the magnetic field in three directions adds the same scale factor (nonlinearity) and cross coupling (misalignment) errors as those present in the inertial sensors, together with the transformation between the magnetic axes and the body ones.

Modeling the behavior of a triad of magnetometers is simpler but less precise than that of inertial sensors, as the effect of the fixed hard iron magnetism is indistinguishable from that of the run-to-run bias offset, while the fixed effect of soft iron magnetism is indistinguishable from that of the scale factor and cross coupling error matrix. This has several consequences. First of all is that magnetometers can not be calibrated at the laboratory before being mounted in the aircraft as in the case of inertial sensors (section \ref{subsec:Inertial_Calibration}), but are instead calibrated once attached to the aircraft by a process known as swinging (section \ref{subsec:Swinging}), which is less precise as the aircraft attitude during swinging can not be determined with so much accuracy as it would be in a laboratory setting. Second is that defining a magnetic platform frame to then transform the results into body axes serves no purpose, as the magnetometer readings are only valid, this is, contain the effects of hard and soft iron magnetism, once they are attached to the aircraft, and then they can be directly measured in body axes. And third is that percentage wise the errors induced by the magnetometers are bigger than those of the inertial sensors. The simulation relies on the following model:
\neweq{\BBtilde = \BzeroMAGvec + \BhiMAGvec + \MMAG \, \RBN \, \BN +  \vec e_{\sss W,MAG}^{\sss B}}{eq:Sensor_NonInertial_mag_error}

where \nm{\BBtilde} is the measurement viewed in \nm{\FB}, \nm{\BzeroMAGvec} is the run-to-run bias offset, \nm{\BhiMAGvec} is the fixed hard iron magnetism, \nm{\MMAG} is a fixed matrix combining the effects of soft iron magnetism with the scale factor and cross coupling errors, and \nm{\BN} is the real magnetic field including local anomalies.
\begin{center}
\begin{tabular}{lrc|llllc}
	\hline
	\textbf{MAG Baseline} & \multicolumn{1}{c}{\textbf{Spec}} & \textbf{Unit} & \multicolumn{1}{c}{\textbf{Variable}} & \multicolumn{1}{c}{\textbf{Value}} & \multicolumn{1}{c}{\textbf{Comp.}} & \multicolumn{1}{c}{\textbf{Swinging}} & \textbf{Unit} \\
	\hline
	Output Noise		& 5				& [\nm{nT \cdot sec^{0.5}}]	& \nm{\sigmavMAG}	& \nm{5.00 \cdot 10^{0}}	& \nm{5.00 \cdot 10^{0}}	& \nm{5.00 \cdot 10^{0}}	& [\nm{nT \cdot sec^{0.5}}] \\
	Nonlinearity		& 0.5			& [\%]						& \nm{\sMAG}		& \nm{5.00 \cdot 10^{-3}}   & \nm{7.50 \cdot 10^{-3}}	& \nm{7.50 \cdot 10^{-4}}	& [-] \\
	Misalignment		& \nm{\pm} 0.35	& [\nm{deg}]				& \nm{\mMAG}		& \nm{6.11 \cdot 10^{-3}}	& \nm{9.16 \cdot 10^{-3}}	& \nm{9.16 \cdot 10^{-4}}	& [-] \\
	Bias (1 \nm{\sigma})& \multirow{2}{*}{\nm{\pm} 1500} & \multirow{2}{*}{[\nm{nT}]} 	& \nm{\BhiMAG}		& \multirow{2}{*}{\nm{1.50 \cdot 10^{3}}} & \nm{1.75 \cdot 10^{3}}	& \nm{1.75 \cdot 10^{2}} & [\nm{nT}] \\
	Repeatability		& 								 &								& \nm{\BzeroMAG}	&  & \nm{5.00 \cdot 10^{2}}	& \nm{5.00 \cdot 10^{2}} & [\nm{nT}] \\
	\hline
\end{tabular}
\captionof{table}{Performance of ``Baseline'' magnetometers} \label{tab:Sensors_mag}
\end{center}

The baseline magnetometer features employed in the simulation are shown in table \ref{tab:Sensors_mag} in the same format as those of the inertial sensors, where the white noise has been taken from \cite{Groves2008} and the rest of the parameters correspond to the magnetometers present in the Analog Devices ADIS16488A IMU \cite{ADIS16488A}. Although the value of hard and soft iron magnetism in aircraft is rather small, the author has not been able to obtain trusted values for them. To avoid eliminating sources of error, the author has decided to increase by 50\% the values for bias offset, scale factor, and cross coupling errors found in the literature, as shown in the column named "Compensation". As both result in a similar effect, the author expects that the realism of the results will not be adversely affected. In the case of the bias, the author has assigned most of the error to the fixed hard iron error \nm{\BhiMAG} and the remaining to the run-to-run bias offset \nm{\BzeroMAG}.

A swinging process as that described in section \ref{subsec:Swinging} is assumed to eliminate 90\% of the fixed error contributions, this is, the hard iron magnetism, the scale factor, and the cross coupling error\footnote{The soft iron effect is combined with the scale factor and cross coupling errors.}. This number is inferior to the 95\% reduction achieved by the calibration of inertial sensors because although the lack of magnetometer bias drift facilitates calibration, the determination of the body attitude during calibration is inherently less accurate working with the complete aircraft than just with the IMU in a laboratory setting. The final results can be found in the rightmost column above.

The final model for a triaxial magnetometer thus includes contributions from the hard iron magnetism, bias offset, system noise, soft iron magnetism, scale factor, and cross coupling errors:
\neweq{\BBtilde \lrp{s \, \Deltat} = \BhiMAG \, \NhiMAG + \BzeroMAG \, \NuzeroMAG + \MMAG \, \RBN \, \BNREAL + \dfrac{\sigmavMAG}{\Deltat^{1/2}} \, \NvsMAG}{eq:Sensor_NonInertial_mag_error_final}

where \nm{\NhiMAG}, \nm{\NuzeroMAG} and \nm{\NvsMAG} are uncorrelated normal vectors of size three each composed of three uncorrelated standard normal random variables \nm{N\lrp{0, \, 1}}. The soft iron, scale factor and cross coupling matrix \nm{\vec M_{\sss MAG}} does not vary with time and is computed as follows:
\neweq{\vec M_{\sss MAG} = \begin{bmatrix} \nm{s_{\sss MAG}} & \nm{m_{\sss MAG}} & \nm{m_{\sss MAG}} \\ \nm{m_{\sss MAG}} & \nm{s_{\sss MAG}} & \nm{m_{\sss MAG}} \\ \nm{m_{\sss MAG}} & \nm{m_{\sss MAG}} & \nm{s_{\sss MAG}} \end{bmatrix} \circ \vec{N}_{m,\sss MAG}} {eq:Sensor_NonInertial_summary_scale_factor}

In this expression \nm{\vec{N}_{m,\sss MAG}} contains nine outputs of a standard normal random variable \nm{N\lrp{0, \, 1}}, and the symbol \nm{\circ} represents the Hadamart or element-wise matrix product.

Table \ref{tab:Sensors_NonInertial_mag_error_sources} lists the error sources contained in the magnetometer model represented by (\ref{eq:Sensor_NonInertial_mag_error_final}), noting that soft iron magnetism is also included in both the scale factor and cross coupling errors. The first two columns list the different error sources, while the third column specifies their origin according to the criterion established in the first paragraph of section \ref{subsec:Sensors_Inertial_ErrorSources}. The last two columns list the seeds employed to ensure the results variability for different aircraft (\nm{\seedA}) as well as different flights (\nm{\seedF}):
\begin{center}
\begin{tabular}{lcccc}
	\hline
	\multicolumn{2}{c}{\textbf{Error}} & \textbf{Main Source} & \multicolumn{2}{c}{\textbf{Seeds}} \\
	\hline
	Hard Iron			& \nm{\BhiMAG}		& fixed			& \nm{\seedAMAG} & \nm{\seedA} \\
	Bias Offset			& \nm{\BzeroMAG}	& run-to-run	& \nm{\seedFMAG} & \nm{\seedF} \\
	System Noise		& \nm{\sigmavMAG}	& in-run		& \nm{\seedFMAG} & \nm{\seedF} \\
	Scale Factor		& \nm{\sMAG}		& fixed			& \nm{\seedAMAG} & \nm{\seedA} \\
	Cross Coupling		& \nm{\mMAG}		& fixed			& \nm{\seedAMAG} & \nm{\seedA} \\
	\hline
\end{tabular}
\end{center}
\captionof{table}{Magnetometer error sources}\label{tab:Sensors_NonInertial_mag_error_sources}

Note that \nm{\BhiMAG}, \nm{\BzeroMAG}, \nm{\sigmavMAG}, \nm{\sMAG}, and \nm{\mMAG} are taken from the right column of the specs listed above, and that all errors are modeled as stochastic variables or processes. The stochastic nature of the fixed and run-to-run error contributions to the magnetometer model relies on three realizations of normal distributions for the hard iron magnetism, three for the bias offset, three for the scale factor errors, and six for the cross coupling errors, while the in-run error contributions require three realizations for system noise at every discrete sensor measurement. 

Expression (\ref{eq:Sensor_NonInertial_mag_error_final}) can be rewritten to show the measurements as functions of the magnetometer full error \nm{\EMAG}, which represents all the errors introduced by the magnetometers with the exception of white noise:
\neweq{\BBtilde \lrp{s \, \Deltat} = \BB \lrp{s \, \Deltat} + \EMAG \lrp{s \, \Deltat} + \dfrac{\sigmavMAG}{\Deltat^{1/2}} \, \NvsMAG} {eq:Sensor_NonInertial_mag_error_filter}


\subsection{Global Navigation Satellite System Receiver}\label{subsec:Sensors_NonInertial_GNSS}

A GNSS receiver enables the determination of the aircraft position and absolute velocity based on signals obtained from various constellations of satellites, such as GPS, GLONASS, and Galileo. The position is obtained by triangulation based on the accurate satellite position and time contained within each signal. Instead of derivating the position with respect to time, which introduces noise, GNSS receivers obtain the vehicle absolute velocity by measuring the Doppler shift between the constant satellite frequencies and those measured by the receiver.

It is important to note that because of the heavy processing required to fix a position based on the satellite signals, GNSS receivers are not capable of working at the high frequencies characteristic of inertial and air data sensors, with 1 [hz] being the default value in this article (\nm{\DeltatGNSS = 1 \lrsb{sec}}). The position error of a GNSS receiver can be modeled as the sum of a zero mean white noise process plus slow varying ionospheric effects \cite{Kayton1997} modeled as the sum of the bias offset plus a random walk. This random walk is modeled with a frequency of \nm{1/60 \, \lrsb{hz}} (\nm{\DeltatION = 60 \lrsb{sec}}), and linearly interpolated in between. The ground velocity error is modeled exclusively with a white noise process.
\begin{eqnarray}
\nm{\vec e_{\sss GNSS,POS} \lrp{g \, \DeltatGNSS}} & = & \nm{\xEgdttilde - \xEgdt = \sigmaGNSSPOS \, \NgGNSSPOS + \vec e_{\sss GNSS,ION} \lrp{g \, \DeltatGNSS}} \label{eq:Sensor_NonInertial_GNSS_pos_error} \\
\nm{\vec e_{\sss GNSS,VEL} \lrp{g \, \DeltatGNSS}} & = & \nm{\vNtilde - \vN = \sigmaGNSSVEL \, \NgGNSSVEL} \label{eq:Sensor_NonInertial_GNSS_vel_error} \\
\nm{\vec e_{\sss GNSS,ION} \lrp{g \, \DeltatGNSS}} & = & \nm{\vec e_{\sss GNSS,ION} \lrp{i \, \DeltatION} +} \nonumber \\
 & & \nm{\dfrac{r}{f_{\sss ION}} \Big(\vec e_{\sss GNSS,ION} \big(\lrp{i+1} \DeltatION\big) - \vec e_{\sss GNSS,ION} \lrp{i \, \DeltatION}\Big)} \label{eq:Sensor_NonInertial_GNSS_ion_error} \\
\nm{g} & = & \nm{f_{\sss ION} \cdot i + r \ \ \ \ \ \ \ \ \ 0 \le r < f_{\sss ION}} \label{eq:Sensor_NonInertial_GNSS_ion_error2} \\
\nm{\vec e_{\sss GNSS,ION} \lrp{i \, \DeltatION}} & = & \nm{\BzeroGNSSION \, \NzeroGNSSION + \sigmaGNSSION \, \sum_{j=1}^i \NjGNSSION} \label{eq:Sensor_NonInertial_GNSS_ion_error3} \\
\nm{f_{\sss ION}} & = & \nm{\DeltatION / \DeltatGNSS = 60}\label{eq:Sensor_NonInertial_GNSS_ion_error4}
\end{eqnarray}

where \nm{\sigmaGNSSPOS}, \nm{\sigmaGNSSION}, \nm{\BzeroGNSSION}, and \nm{\sigmaGNSSVEL} are taken from the table below, and \nm{\NgGNSSPOS}, \nm{\NgGNSSVEL}, \nm{\NzeroGNSSION}, and \nm{\NjGNSSION} and  are uncorrelated normal vectors of size three each composed of three uncorrelated standard normal random variables \nm{N\lrp{0, \, 1}}. Also note that as both g and \nm{f_{\sss ION}} are integers, the quotient remainder theorem guarantees that there exist unique integers i and r that comply with (\ref{eq:Sensor_NonInertial_GNSS_ion_error2}) \cite{Pinter1990}.
\begin{center}
\begin{tabular}{lrc|lrc}
	\hline
	\textbf{GNCC} & \multicolumn{1}{c}{\textbf{Spec}} & \textbf{Unit} & \multicolumn{1}{c}{\textbf{Variable}} & \multicolumn{1}{c}{\textbf{Value}} & \textbf{Unit} \\
  	\hline
	Horizontal position accuracy (CEP 50\%)			& 2.50	& [\nm{m}]		& \nm{\sigmaGNSSPOSHOR} & \nm{2.12 \cdot 10^{0}}	& [\nm{m}] \\
	Vertical position accuracy (CEP 50\%)	  		& N/A	& 				& \nm{\sigmaGNSSPOSVER} & \nm{4.25 \cdot 10^{0}}	& [\nm{m}] \\
	Ionospheric random walk \nm{1/60 \, \lrsb{hz}}	& N/A	& 				& \nm{\sigmaGNSSION}	& \nm{1.60 \cdot 10^{-1}}	& [\nm{m}] \\
	Ionospheric bias offset                         & N/A   &               & \nm{\BzeroGNSSION}    & \nm{8.00 \cdot 10^{0}}   & [\nm{m}] \\
	Velocity accuracy (50\%) 						& 0.05	& [\nm{m/sec}]	& \nm{\sigmaGNSSVEL}	& \nm{7.41 \cdot 10^{-2}}	& [\nm{m/sec}] \\
  	\hline
\end{tabular}
\captionof{table}{Performances of GNSS receiver} \label{tab:Sensors_gnss}
\end{center}

The horizontal position accuracy in the table above corresponds to the U-blox NEO-M8 receiver data sheet \cite{NEOM8}, where CEP stands for circular error probability. As CEP is equivalent to 1.18 standard deviations \cite{NOVATELGPS}, it enables the obtainment of \nm{\sigmaGNSSPOSHOR}. The author has not found any reference for GNSS vertical accuracy, but has determined through conversations with several knowledgeable individuals that it is at least 50\% higher than the horizontal one. A conservative value for \nm{\sigmaGNSSPOSVER} of twice that of \nm{\sigmaGNSSPOSHOR} has been selected. The ionospheric effects have also been obtained from these conversations. The velocity accuracy also originates at the U-blox NEO-M8 receiver data sheet \cite{NEOM8}. Assuming that it corresponds to a per axis error of \nm{\pm 0.05 \, [m/sec]} instead of CEP, and knowing that the 50\% mark of a normal distribution lies at 0.67448 standard deviations, it is possible to obtain \nm{\sigma_{\sss GNSS,VEL}}.

Table \ref{tab:Sensors_NonInertial_GNSS_error_sources} lists the error sources contained in the GNSS receiver model represented by (\ref{eq:Sensor_NonInertial_GNSS_pos_error}, \ref{eq:Sensor_NonInertial_GNSS_vel_error}). The first two columns list the different error sources, while the third column specifies their origin according to the criterion established in the first paragraph of section \ref{subsec:Sensors_Inertial_ErrorSources}. The last two columns list the seeds employed to ensure the results variability for different flights (\nm{\seedF}):
\begin{center}
\begin{tabular}{lcccc}
	\hline
	\multicolumn{2}{c}{\textbf{Error}} & \textbf{Main Source} & \multicolumn{2}{c}{\textbf{Seeds}} \\
	\hline
	Bias Offset			& \nm{\BzeroGNSSION}										& run-to-run	& \nm{\seedFGNSS} & \nm{\seedF} \\
	System Noise		& \nm{\sigmaGNSSPOS, \, \sigmaGNSSVEL, \, \sigmaGNSSION}	& in-run		& \nm{\seedFGNSS} & \nm{\seedF} \\
	\hline
\end{tabular}
\end{center}
\captionof{table}{GNSS receiver error sources}\label{tab:Sensors_NonInertial_GNSS_error_sources}

Note that all errors are modeled as stochastic variables or processes. Three realizations of a normal distribution are required for the run-to-run error contributions, while the in-run error contributions require three realizations each for position and velocity at every discrete sensor measurement, plus an extra three when corresponding for the ionospheric error.


\subsection{Air Data System} \label{subsec:Sensors_NonInertial_ADS}

The mission of the air data system is to measure the aircraft pressure altitude \nm{\Hp} \cite{ISA, INSA} by means of the atmospheric pressure \nm{p}, the outside air temperature T, the airspeed \nm{\vtas}, and the angles of attack and sideslip that provide the orientation of the aircraft structure with respect to the airflow.

A barometer or static pressure sensor, generally part of the Pitot tube as explained below \cite{Titterton2004}, measures atmospheric pressure, which can be directly translated into pressure altitude \cite{ISA, INSA}. The \nm{\sigmaOSP} value shown in table \ref{tab:Sensors_air} below, where \texttt{OSP} stands for outside static pressure, comes from the \nm{\pm \, 10 \, \lrsb{m}} altitude error listed in the specifications of the Aeroprobe air data system \cite{Aeroprobe}, which translates into \nm{\pm \, 100 \, \lrsb{pa}} at a pressure altitude of \nm{1500 \, \lrsb{m}}. Although not present in the documentation, the author has decided to also include a bias offset \nm{\BzeroOSP} for added realism, so the barometer error is modeled as a combination of bias offset and random noise:
\neweq{e_{\sss OSP} \lrp{s \, \DeltatSENSED} = e_{\sss OSP} \lrp{s \, \Deltat} = \widetilde{p}\lrp{s \, \Deltat} - p\lrp{s \, \Deltat} = \BzeroOSP \, \NzeroOSP +  \sigmaOSP \, \NsOSP}{eq:Sensor_NonInertial_OSP}

where \nm{\NzeroOSP} and \nm{\NsOSP} are uncorrelated standard normal random variables \nm{N\lrp{0, \, 1}}.

Air data systems are also equipped with a thermometer to measure the external air temperature T. As in the case of the barometer above, the specifications of the Analog Devices ADT7420 temperature sensor \cite{ADT7420} only include system noise, but the author has decided to also include a bias offset for added realism. The model contained in (\ref{eq:Sensor_NonInertial_OAT}) hence relies on the bias offset \nm{\BzeroOAT} and system noise \nm{\sigmaOAT} shown in table \ref{tab:Sensors_air}, where \texttt{OAT} stands for outside air temperature, plus two uncorrelated standard normal random variables \nm{N\lrp{0, \, 1}} (\nm{\NzeroOAT} and \nm{\NsOAT}):
\neweq{e_{\sss OAT} \lrp{s \, \DeltatSENSED} = e_{\sss OAT} \lrp{s \, \Deltat} = \widetilde{T}\lrp{s \, \Deltat} - T\lrp{s \, \Deltat} = \BzeroOAT \, \NzeroOAT + \sigmaOAT \, \NsOAT}{eq:Sensor_NonInertial_OAT}
\begin{center}
\begin{tabular}{lrc|lrc}
	\hline
	\textbf{OSP-OAT Baseline} & \multicolumn{1}{c}{\textbf{Spec}} & \textbf{Unit} & \multicolumn{1}{c}{\textbf{Variable}} & \multicolumn{1}{c}{\textbf{Value}} & \textbf{Unit} \\
	\hline
	Altitude Error			         & \nm{\pm \, 10}	& [\nm{m}]		   & \nm{\sigmaOSP}	& \nm{1.00 \cdot 10^{+2}}	& [\nm{pa}] \\
	                    	         & 					&				   & \nm{\BzeroOSP}	& \nm{1.00 \cdot 10^{+2}}	& [\nm{pa}] \\
	Temperature Error (\nm{3 \sigma}) & \nm{\pm \, 0.15} & [\nm{^{\circ}K}] & \nm{\sigmaOAT}	& \nm{5.00 \cdot 10^{-2}}   & [\nm{^{\circ}K}] \\
	                    	         & 					&				   & \nm{\BzeroOAT}	& \nm{5.00 \cdot 10^{-2}}	& [\nm{^{\circ}K}] \\
	\hline
\end{tabular}
\captionof{table}{Performance of ``Baseline'' atmospheric sensors} \label{tab:Sensors_air}
\end{center}

A \emph{Pitot probe} is a tube with no outlet pointing directly into the undisturbed air stream, where the values of the air variables (temperature, pressure, and density) at its dead end resemble the total or stagnation variables of the atmosphere prior to its deceleration inside the Pitot \cite{Eshelby2000}. \emph{Total atmospheric variables} (temperature \nm{T_t}, pressure \nm{p_t}, and density \nm{\rho_t}) are those obtained if a moving fluid with static atmospheric variables (T, p, \nm{\rho}) and speed \nm{v} decelerates until it has no speed through a process that is stationary, has no viscosity nor gravity (gravitation plus inertial) accelerations, is adiabatic, and presents fixed boundaries for the analyzed control volume \cite{Batchelor2000,Oates1989}. Such a process maintains the fluid total enthalpy, as well as its entropy, and hence complies with the Bernoulli equation \cite{Batchelor2000}:
\neweq{\dfrac{d}{dt}\lrp{\dfrac{\kappa}{\kappa - 1} \dfrac{p}{\rho} + \dfrac{1}{2}\, v^2} = 0}{eq:Sensor_NonInertial_Bernoulli}

where \nm{\kappa = 1.4} is the air adiabatic index. In addition to the static pressure and temperature sensors discussed above, a Pitot tube is also equipped with a dynamic pressure sensor located at the dead end to measure the air flow total pressure \nm{p_t}. The air data system then estimates the aircraft true airspeed based on the following expression, which results from applying (\ref{eq:Sensor_NonInertial_Bernoulli}) at the Pitot dead end as well as at the static ports:
\neweq{\vtas = \sqrt{\frac{2 \, \kappa}{\kappa - 1}\frac{p}{\rho}\lrsb{\lrp{\frac{p_t - p}{p} + 1}^{\textstyle \frac{\kappa - 1}{\kappa}}-1}}}{eq:Sensor_NonInertial_tas}

The errors induced when measuring the true airspeed this way can also be modeled by a combination of bias offset and system noise, where \nm{\sigmaTAS} is the error standard deviation taken from table \ref{tab:Sensors_vtasb}, \nm{\BzeroTAS} has been added by the author for increased realism, and \nm{\NzeroTAS} and \nm{\NsTAS} are two uncorrelated standard normal random variables \nm{N\lrp{0, \, 1}}:
\neweq{e_{\sss TAS} \lrp{s \, \DeltatSENSED} = e_{\sss TAS} \lrp{s \, \Deltat} = \vtastilde\lrp{s \, \Deltat} - \vtas\lrp{s \, \Deltat} = \BzeroTAS \, \NzeroTAS + \sigmaTAS \, \NsTAS}{eq:Sensor_NonInertial_TAS}

The Aeroprobe air data system specifications \cite{Aeroprobe} list a maximum airspeed error of \nm{1 \, \lrsb{m/sec}}, which can be interpreted as \nm{3 \, \sigma}, and hence results in the \nm{\sigmaTAS} value shown below:
\begin{center}
\begin{tabular}{lrc|lrc}
	\hline
	\textbf{TAS-AOA-AOS Baseline} & \multicolumn{1}{c}{\textbf{Spec}} & \textbf{Unit} & \multicolumn{1}{c}{\textbf{Variable}} & \multicolumn{1}{c}{\textbf{Value}} & \textbf{Unit} \\
	\hline
	Airspeed Error (max)	         & \nm{1}			& [\nm{m/sec}]	   & \nm{\sigmaTAS}	& \nm{3.33 \cdot 10^{-1}}	& [\nm{m/sec}] \\
	                    	         & 					&				   & \nm{\BzeroTAS}	& \nm{3.33 \cdot 10^{-1}}	& [\nm{m/sec}] \\
	Flow Angle Error (max)	         & \nm{\pm \, 1.0}	& [\nm{deg}]	   & \nm{\sigmaAOA}	& \nm{3.33 \cdot 10^{-1}}	& [\nm{deg}] \\
	                    	         & 					&				   & \nm{\BzeroAOA}	& \nm{3.33 \cdot 10^{-1}}	& [\nm{deg}] \\
	Flow Angle Error (max)	         & \nm{\pm \, 1.0}	& [\nm{deg}]	   & \nm{\sigmaAOS}	& \nm{3.33 \cdot 10^{-1}}	& [\nm{deg}] \\
							         &					& 				   & \nm{\BzeroAOS}	& \nm{3.33 \cdot 10^{-1}}	& [\nm{deg}] \\
	\hline
\end{tabular}
\captionof{table}{Performance of ``Baseline'' Pitot tube and air vanes} \label{tab:Sensors_vtasb}
\end{center}

The air data system is also capable of measuring the direction of the air stream with respect to the aircraft, represented by the angles of attack and sideslip. To do so, it can be equipped with two air vanes that align themselves with the unperturbed air stream or with a more complex multi hole Pitot probe. The latter is the case of the Aeroprobe air data system \cite{Aeroprobe} employed as the baseline in this article, which measures both angles with a maximum error of \nm{\pm \, 1.0 \, \lrsb{deg}}. If interpreted as \nm{3 \, \sigma}, this results in standard deviations \nm{\sigmaAOA} and \nm{\sigmaAOS} of \nm{0.33 \, \lrsb{deg}}, where \texttt{AOA} and \texttt{AOS} stand for angles of attack and sideslip, respectively. Although not present in the documentation, the author has decided to also include bias offsets \nm{\BzeroAOA} and \nm{\BzeroAOS} to provide more realism to the sensor models, specially in the case of air vanes. The final expressions for the angles of attack and sideslip measurement errors are the following:
\begin{eqnarray}
\nm{e_{\sss AOA} \lrp{s \, \DeltatSENSED} = e_{\sss AOA} \lrp{s \, \Deltat}} & = & \nm{\widetilde\alpha\lrp{s \, \Deltat} - \alpha\lrp{s \, \Deltat} = \BzeroAOA \, \NzeroAOA + \sigmaAOA \, \NsAOA}\label{eq:Sensor_NonInertial_AOA} \\
\nm{e_{\sss AOS} \lrp{s \, \DeltatSENSED} = e_{\sss AOS} \lrp{s \, \Deltat}} & = & \nm{\widetilde\beta \lrp{s \, \Deltat} - \beta \lrp{s \, \Deltat} = \BzeroAOS \, \NzeroAOS + \sigmaAOS \, \NsAOS}\label{eq:Sensor_NonInertial_AOS}
\end{eqnarray}

where \nm{\NzeroAOA}, \nm{\NsAOA}, \nm{\NzeroAOS}, and \nm{\NsAOS} are uncorrelated standard normal random variables \nm{N\lrp{0, \, 1}}.

Table \ref{tab:Sensors_NonInertial_air_data_sources} lists the error sources contained in the air data sensor model represented by (\ref{eq:Sensor_NonInertial_OSP}, \ref{eq:Sensor_NonInertial_OAT}, \ref{eq:Sensor_NonInertial_TAS}, \ref{eq:Sensor_NonInertial_AOA}, \ref{eq:Sensor_NonInertial_AOS}). The specific errors are contained in the first two columns, while their origin (according to the criterion established in the first paragraph of section \ref{subsec:Sensors_Inertial_ErrorSources}) is listed in the third column. The last two columns list the seeds employed to ensure the results variability for different flights (\nm{\seedF}):
\begin{center}
\begin{tabular}{lcccc}
	\hline
	\multicolumn{2}{c}{\textbf{Error}} & \textbf{Main Source} & \multicolumn{2}{c}{\textbf{Seeds}} \\
	\hline
	Bias Offset		& \nm{\BzeroOSP, \, \BzeroOAT, \, \BzeroTAS, \, \BzeroAOA, \, \BzeroAOS}	& run-to-run & \nm{\seedFOSP, \seedFOAT} & \multirow{2}{*}{\nm{\seedF}} 	\\
	System Noise	& \nm{\sigmaOSP, \, \sigmaOAT, \, \sigmaTAS, \, \sigmaAOA, \, \sigmaAOS}	& in-run	 & \nm{\seedFTAS, \seedFAOA, \seedFAOS}   	\\
	\hline
\end{tabular}
\end{center}
\captionof{table}{Air data sensor error sources}\label{tab:Sensors_NonInertial_air_data_sources}

Note that all errors are modeled as stochastic variables or processes. The stochastic nature of the run-to-run error contributions to the models relies on five realizations of normal distributions for the bias offsets, while the in-run error contributions require five realizations for the system noises at every discrete sensor measurement.


\section{Camera} \label{sec:Sensors_camera}

Image generation is a power and data intensive process that can not work at the high frequencies characteristic of inertial and air data sensors. A frequency of 10 [hz] is employed as the default in this article (\nm{\DeltatIMG = 0.1 \lrsb{sec}}), although there are cameras available capable of working significantly faster. The camera is considered rigidly attached to the aircraft structure, and it is assumed that the shutter speed is sufficiently high that all images are equally sharp, and that the image generation process is instantaneous. In addition, the camera \texttt{ISO} setting remains constant during the flight, and all generated images are noise free. The simulation also assumes that the visible spectrum radiation reaching all patches of the Earth surface remains constant, and the terrain is considered Lambertian \cite{Soatto2001}, so its appearance at any given time does not vary with the viewing direction. The combined use of these assumptions implies that a given terrain object is represented with the same luminosity in all images, even as its relative pose (position and attitude) with respect to the camera varies. 
\begin{center}
\begin{tabular}{lcrc}
	\hline
	\textbf{Parameter} & \textbf{Symbol} & \textbf{Value} & \textbf{Unit} \\
	\hline
	Focal length						& f            & 19.0			       & [mm]   \\
	Image width                         & \nm{\Sh}   & 768                   & [px]    \\
	Image height                        & \nm{\Sv}   & 1024                  & [px]    \\
	Pixel size                          & \nm{\sPX}    & \nm{17 \cdot 10^{-3}} & [mm/px] \\
	Principal point horizontal location & \nm{\cIMGi}  & 384.5                 & [px]    \\
	Principal point vertical location   & \nm{\cIMGii} & 511.5                 & [px]    \\
	Horizontal field of view            & \nm{\fovh}   & 37.923               & [deg]   \\
	Vertical field of view              & \nm{\fovv}   & 49.226               & [deg]   \\
	\hline
\end{tabular}
\captionof{table}{Camera parameters} \label{tab:Sensors_cam}
\end{center}

Geometrically, the simulation adopts a perspective projection or pinhole camera model \cite{Soatto2001}, which in addition is perfectly calibrated and hence shows no distortion. Table \ref{tab:Sensors_cam} contains the default camera parameters employed in the simulation, which can be modified by the user if so desired.


\subsection{Mounting of Camera}\label{subsec:Sensors_Camera_Mounting}

The digital camera can be located anywhere on the aircraft structure as long as its view of the terrain is unobstructed by other platform elements. It is desirable that the lever arm or distance between the camera optical center and the aircraft center of mass is as small as possible to reduce the negative effects of any camera alignment error. With respect to its orientation, the camera should be facing down to show a balanced view of the ground during level flight, but minor deviations are not problematic. 

As in the case of the IMU platform, the simulation considers that the camera location is deterministic but its orientation stochastic. The expressions below are hence analogous to those employed in section \ref{subsec:Sensors_Inertial_Mounting}, where each specific camera Euler angle is obtained as the product of the standard deviations (\nm{\sigmapsiC}, \nm{\sigmathetaC}, \nm{\sigmaxiC}) contained in table \ref{tab:Sensor_Camera_mounting} by a single realization of a standard normal random variable \nm{N\lrp{0, \, 1}} (\nm{\NpsiC}, \nm{\NthetaC}, and \nm{\NxiC}):
\begin{eqnarray}
\nm{\TBCB}      & = & \nm{\vec f\lrp{m} = \TBCBfull + \dfrac{m_{full} - m}{m_{full} - m_{empty}} \; \lrp{\TBCBempty - \TBCBfull}} \label{eq:Sensors_Camera_Mounting_Tbcb} \\
\nm{\phiBC}     & = & \nm{\lrsb{90 \lrsb{deg} + \sigmapsiC \, \NpsiC, \ \sigmathetaC \, \NthetaC, \ \sigmaxiC \NxiC}^T} \label{eq:Sensors_Camera_Mounting_eulerBC}
\end{eqnarray}

The problem however is not the real translation and rotation between the \nm{\FB} and \nm{\FC} frames given by the previous equations, but the accuracy with which they are known to the navigation system. The determination of the camera position \nm{\TBCBest} and rotation \nm{\phiBCest = \lrsb{\psiCest, \ \thetaCest, \ \xiCest}^T} is discussed in section \ref{subsec:Camera_frame}. As in previous cases, stochastic models are considered for both the translation \nm{\TBCBest} and rotation \nm{\phiBCest}, changing their values from one simulation run to another:
\begin{eqnarray}
\nm{\TBCBest} & = & \nm{\TBCB + \lrsb{\sigmaTBCBest \, \NTBCBiest, \ \sigmaTBCBest \, \NTBCBiiest, \ \sigmaTBCBest \NTBCBiiiest}^T} \label{eq:Sensors_Camera_Mounting_TBCBest} \\
\nm{\phiBCest} & = & \nm{\phiBC + \lrsb{\sigmaphiBCest \, \NpsiCest, \ \sigmaphiBCest \, \NthetaCest, \ \sigmaphiBCest \NxiCest}^T} \label{eq:Sensors_Camera_Mounting_eulerBCest}
\end{eqnarray}

where the standard deviations \nm{\sigmaTBCBest} and \nm{ \sigmaphiBCest} are shown in table \ref{tab:Sensor_Camera_mounting}, and \nm{\NpsiCest}, \nm{\NthetaCest}, \nm{\NxiCest}, \nm{\NTBCBiest}, \nm{\NTBCBiiest}, \nm{\NTBCBiiiest} are six realizations of a standard normal random variable \nm{N\lrp{0, \, 1}}.

Table \ref{tab:Sensor_Camera_mounting} lists the default values employed in the simulation, which can be adjusted by the user:
\begin{center} 
\begin{tabular}{lccc}
	\hline
	\textbf{Concept} & \textbf{Variable} & \textbf{Value} & \textbf{Unit} \\
	\hline
	True Yaw Error 	   	             & \nm{\sigmapsiC}     & \nm{0.1}  & [\nm{deg}]  \\
	True Pitch Error                 & \nm{\sigmathetaC}   & \nm{0.1}  & [\nm{deg}]  \\
	True Bank Error                  & \nm{\sigmaxiC}      & \nm{0.1}  & [\nm{deg}]  \\
	Camera Position Estimation Error & \nm{\sigmaTBCBest}  & \nm{0.002} & [\nm{m}] \\
	Camera Angular Estimation Error  & \nm{\sigmaphiBCest} & \nm{0.01} & [\nm{deg}]  \\
	\hline
\end{tabular}
\end{center}
\captionof{table}{Camera mounting accuracy values}\label{tab:Sensor_Camera_mounting}

The translation \nm{\TBCB} between the origins of the \nm{\FB} and \nm{\FC} frames can be considered quasi stationary as it slowly varies based on the aircraft mass (\ref{eq:Sensors_Camera_Mounting_Tbcb}), and the relative position of their axes \nm{\phiBC} remains constant because the camera is rigidly attached to the aircraft structure (\ref{eq:Sensors_Camera_Mounting_eulerBC}). 


\subsection{Earth Viewer}\label{subsec:Sensors_camera_earth_viewer}

\begin{figure}[h]
\centering
\includegraphics[width=7.5cm]{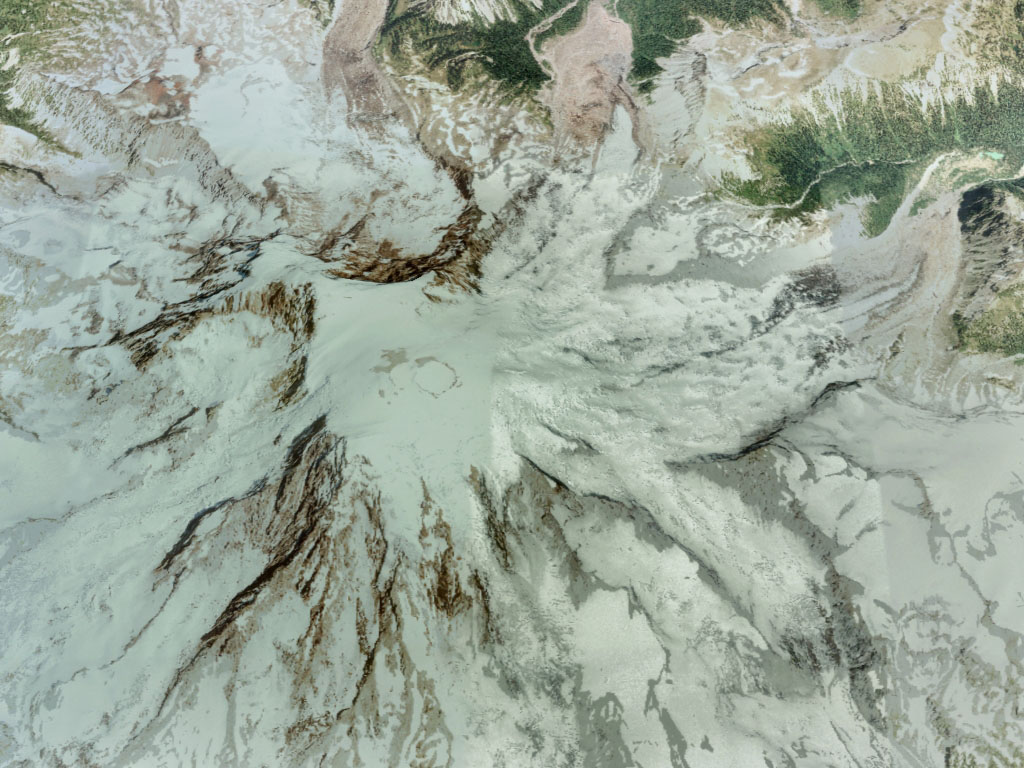}
\hskip 10pt
\includegraphics[width=7.5cm]{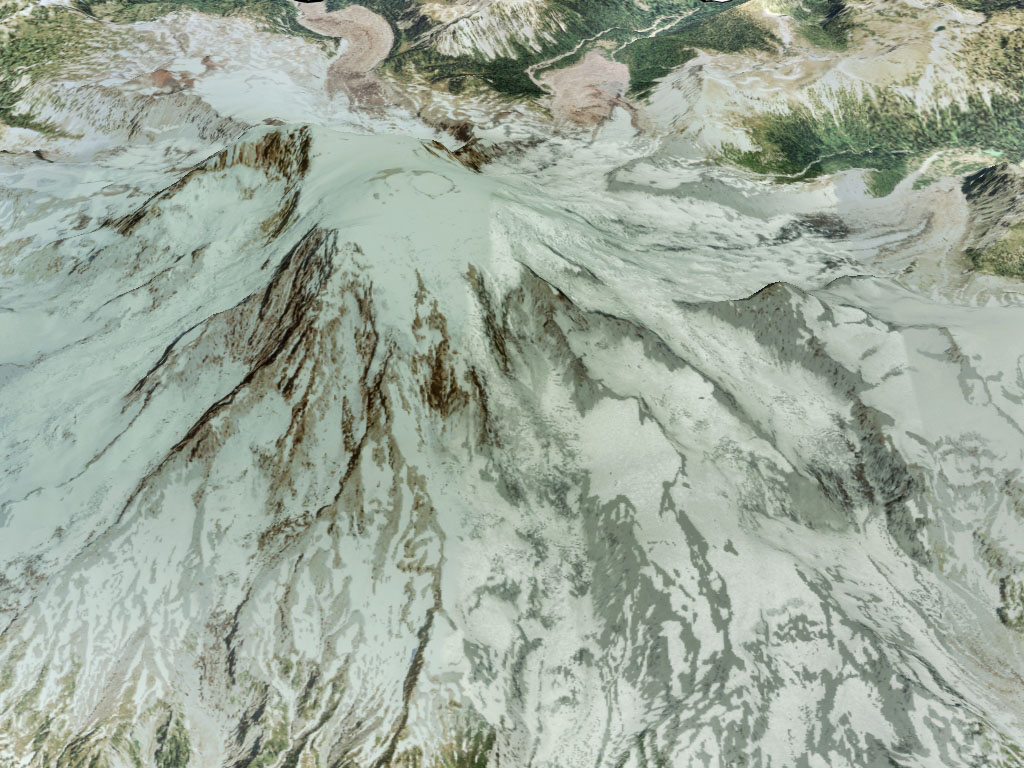}
\caption{Example of Earth Viewer images}
\label{fig:EarthViewer_photo}
\end{figure}

The camera model differs from all other sensor models described in this chapter in that it does not return a sensed variable \nm{\xvectilde} consisting of its real value \nm{\xvec} plus a sensor error \nm{\vec E}, but instead generates a digital image simulating what a real camera would record based on the aircraft position and attitude as given by the actual or real trajectory \nm{\xvec = \xTRUTH}. When provided with the camera pose with respect to the Earth at equally time spaced intervals, the simulation is capable of generating images that resemble the view of the Earth surface that the camera would record if located at that particular pose. To do so, it relies on three software libraries:
\begin{itemize}

\item \texttt{OpenSceneGraph} \cite{OpenSceneGraph} is an open source high performance 3D graphics toolkit written in \texttt{C++} and \texttt{OpenGL}, used by application developers in fields such as visual simulation, games, virtual reality, scientific visualization and modeling. The library enables the representation of objects in a scene by means of a graph data structure, which allows grouping objects that share some properties to automatically manage rendering properties such as the level of detail necessary to faithfully draw the scene, but without considering the unnecessary detail that slows down the graphics hardware drawing the scene.

\item \texttt{osgEarth} \cite{osgEarth} is a dynamic and scalable 3D Earth surface rendering toolkit that relies on \texttt{OpenSceneGraph}, and is based on publicly available \emph{orthoimages} of the area flown by the aircraft. Orthoimages consist of aerial or satellite imagery geometrically corrected such that the scale is uniform; they can be used to measure true distances as they are accurate representations of the Earth surface, having been adjusted for topographic relief, lens distortion, and camera tilt. When coupled with a \emph{terrain elevation model}, \texttt{osgEarth} is capable of generating realistic images based on the camera position as well as its yaw and pitch, but does not accept the camera roll (in other words, the \texttt{osgEarth} images are always aligned with the horizon).

\item \texttt{Earth Viewer} is a modification to \texttt{osgEarth} implemented by the author so it is also capable of accepting the bank angle of the camera with respect to the \texttt{NED} axes. \texttt{Earth Viewer} is capable of generating realistic Earth images as long as the camera height over the terrain is significantly higher than the vertical relief present in the image. As an example, figure \ref{fig:EarthViewer_photo} shows two different views of a volcano in which the dome of the mountain, having very steep slopes, is properly rendered.
\end{itemize}


\section{Calibration Procedures}\label{sec:Calibration}

This section describes various calibration processes required for the determination of the fixed and run-to-run error contributions to the accelerometers, gyroscopes, magnetometers, and onboard camera. These procedures only need to be executed once and do not need to be repeated unless the sensors are replaced or their position inside the aircraft modified\footnote{In addition, the swinging process of section \ref{subsec:Swinging} needs to be performed every time new equipment is installed inside the aircraft, as this may modify the hard and soft iron magnetism and hence the magnetometer readings.}.

The calibration procedures include the laboratory calibration of the accelerometers and gyroscopes described in section \ref{subsec:Inertial_Calibration}, the determination of the pose between the platform and body frames explained in section \ref{subsec:Platform_frame}, the magnetometer calibration or swinging described in section \ref{subsec:Swinging}, and the determination of the pose between the camera and body frames explained in section \ref{subsec:Camera_frame}. Their main objective is the determination of the fixed contributions to the sensor error models\footnote{Refer to section \ref{subsec:Sensors_Inertial_ErrorSources} for the different types of sensor error contributions, including fixed, run-to-run, and in-run.}, this is, the scale factor and cross coupling errors of both inertial sensors and magnetometers (\nm{\MACCest, \, \MGYRest, \, \MMAGest})\footnote{Note that \nm{\MMAGest} also includes the soft iron magnetism.}, the magnetometers hard iron magnetism \nm{\BhiMAGvecest}, the body to platform transformation (\nm{\TBPBest, \ \phiBPest}), and the body to camera transformation (\nm{\TBCBest, \ \phiBCest}). These procedures also provide estimations for the run-to-run error contributions (\nm{\BzeroACCest, \ \BzeroGYRest, \ \BzeroMAGvecest}), but these need to be discarded as they change every time the aircraft systems are switched on.


\subsection{Inertial Sensors Calibration}\label{subsec:Inertial_Calibration}

\emph{Calibration} is the process of comparing instrument outputs with known references to determine coefficients that force the outputs to agree with the references over a range of output values \cite{Chatfield1997}. The IMU inertial sensors need to be calibrated to eliminate the fixed errors originated from manufacturing, and also to determine their temperature sensitivity \cite{Rogers2007}. The calibration process requires significant material and time resources, but it greatly reduces the measurement errors. While high grade IMUs are always factory calibrated, low cost ones generally are not, so it is necessary to calibrate the IMU at the laboratory before mounting it on the aircraft \cite{Groves2008}.

The calibration process is executed at a location where the position \nm{\xEgdt} and the gravity\footnote{Gravity includes both gravitation and centrifugal accelerations.} vector \nm{\gc} have been previously determined with great precision \cite{Chatfield1997}. It relies on a three axis table, which enables rotating the IMU with known angular velocities into a set of predetermined precisely controlled orientations \cite{Rogers2007,Tidaldi2014}. Accelerometer and gyroscope measurements are then compared to reference values (gravity for the accelerometers, torquing rate plus Earth angular velocity for the gyroscopes), and the differences employed to generate corrections \cite{Chatfield1997}.

During calibration, the amount of time that the IMU is maintained stationary at each attitude, as well as the time required to rotate it between two positions, are trade offs based on two opposing influences. On one side, longer periods of time are preferred as the negative influence of the system noise in the measurements tends to even out over time, while on the other, shorter times imply smaller variations of the bias drift over the measurement interval.

It is worth noting that as the calibration is performed before the IMU is installed on the aircraft, it relies on the platform frame \nm{\FP} and the models contained in sections \ref{subsec:Sensors_Accelerometer_Triad_ErrorModel} and  \ref{subsec:Sensors_Gyroscope_Triad_ErrorModel}. Although it is possible to use a calibration strategy based on selecting platform orientations that isolate sensor input onto a single axis (for example, gravity will only be sensed by the accelerometer that is placed vertically with respect to the Earth surface) to then apply least squares techniques, in real life it is better to employ state estimation techniques\footnote{The estimation filter not only relies on known gravity and angular velocity, but also the fact that the IMU is stationary and hence its velocity is zero.} to obtain estimates of the inertial sensors scale factors, cross coupling errors, and bias offsets \cite{Rogers2007,Groves2008}. The process is repeated at different temperatures so the IMU processor can later apply the correction based on the IMU sensor temperature \cite{Groves2008}.
\begin{center}
\begin{tabular}{ccc}
	\hline
	\textbf{Estimation} & \textbf{\#} & \textbf{Coefficients} \\
	\hline
	\nm{\MACCest}		& 6 & \nm{\lrb{\sACCXiest, \ \mACCXijest}}	\\
	\nm{\MGYRest}		& 9 & \nm{\lrb{\sGYRXiest, \ \mGYRXijest}}	\\ 
	\nm{\BzeroACCest}	& 3 & \nm{\BzeroACC \ \NuzeroACCiest}		\\
	\nm{\BzeroGYRest}	& 3 & \nm{\BzeroGYR \ \NuzeroGYRiest}		\\
	\hline
\end{tabular}
\end{center}
\captionof{table}{Results of calibration process}\label{tab:PreFlight_Inertial_calibration}

The twenty-one coefficients estimated in the calibration process are listed in table \ref{tab:PreFlight_Inertial_calibration}. Once the coefficients have been estimated, they can be introduced into the IMU processor so it automatically performs the corrections contained in (\ref{eq:PreFlight_Inertial_acc_error_calib_correction}) and (\ref{eq:PreFlight_Inertial_gyr_error_calib_correction}):
\begin{eqnarray}
\nm{\fIPPtildetilde} & = & \nm{{\MACCest}^{-1} \ \fIPPtilde - \BzeroACCest}\label{eq:PreFlight_Inertial_acc_error_calib_correction} \\
\nm{\wIPPtildetilde} & = & \nm{{\MGYRest}^{-1} \ \wIPPtilde - \BzeroGYRest}\label{eq:PreFlight_Inertial_gyr_error_calib_correction}
\end{eqnarray}

This article assumes that the bias offset is exclusively a run-to-run source of error that varies every time the IMU is switched on, so the \nm{\BzeroACCest} and \nm{\BzeroGYRest} coefficients obtained by calibration are discarded as they have no relation to the offsets that occur during flight. Modeling the calibration process results implies reducing the scale factor and cross couplings errors found on the inertial sensors specifications by an arbitrary amount of \nm{95\%}. This reduction is already included in the baseline tables contained in section \ref{subsec:Sensors_Inertial_Selected_gyr_acc}. To summarize, instead of applying (\ref{eq:PreFlight_Inertial_acc_error_calib_correction}, \ref{eq:PreFlight_Inertial_gyr_error_calib_correction}) to the measurements obtained by (\ref{eq:Sensor_Inertial_gyr_error_final}, \ref{eq:Sensor_Inertial_acc_error_final}), the simulation directly employs (\ref{eq:Sensor_Inertial_gyr_error_final}, \ref{eq:Sensor_Inertial_acc_error_final}) with reduced \nm{\MGYR} and \nm{\MACC} taken from the tables of sections \ref{subsec:Sensors_Inertial_Selected_gyr_acc}.


\subsection{Determination of the Platform Frame Pose}\label{subsec:Platform_frame}

The true relative pose between the body and platform frames (\nm{\FB}, \nm{\FP}), given by \nm{\TBPB} and \nm{\phiBP}, as well as their estimated values \nm{\TBPBest} and \nm{\phiBPest}, play a key role in the readings generated by the inertial sensors, as explained in section \ref{subsec:Sensors_Inertial_ErrorModel}. Let's see how they can be estimated.

Considering that the position of the aircraft center of mass is known (in both full and empty tank configurations), the true displacement \nm{\TBPB} can be determined with near exactitude based on the IMU attachment point to the aircraft, resulting in the very small error \nm{\sigmaTBPBest} assigned in table \ref{tab:Sensor_Inertial_mounting} for the estimation of \nm{\TBPBest} in (\ref{eq:Sensors_Inertial_Mounting_TBPBest}).

With regards to the attitude \nm{\phiBP}, after mounting the IMU platform so two of its axes are approximately aligned with the forward and down directions of an approximate aircraft plane of symmetry (with no particular need for accuracy), measure the angular deviation \nm{\phiBP} by means of the self-alignment \cite{Groves2008}, resulting in the small error \nm{\sigmaphiBPest} assigned in table \ref{tab:Sensor_Inertial_mounting} for the estimation of \nm{\phiBPest} in (\ref{eq:Sensors_Inertial_Mounting_eulerBPest}).


\subsection{Swinging or Magnetometer Calibration}\label{subsec:Swinging}

Magnetometer calibration is inherently more complex than that of the inertial sensors as it must be performed with the sensors already mounted on the aircraft, as otherwise it would not capture the fixed contributions of the hard iron and soft iron magnetisms (section \ref{subsec:Sensors_NonInertial_Magnetometers}). The calibration process, known as \emph{swinging}, relies on obtaining magnetometer readings while the aircraft is positioned at different attitudes that encompass a wide array of heading, pitch, and roll values \cite{Groves2008}, and is executed at a location where the magnetic field is precisely known. 

The accuracy of the results is very dependent of the precision with which the different aircraft attitudes can be determined during swinging. This can be done with self-alignment procedures \cite{Groves2008} or with the use of expensive static instruments. In any case, attitude accuracy is always going to be inferior to that obtained with a three axis table during inertial sensor calibration. Once the magnetic field readings are obtained, they are compared to the real magnetic field values, and expression (\ref{eq:Sensor_NonInertial_mag_error_final}) employed with least squares techniques to obtain estimations of the bias (sum of hard iron magnetism \nm{\BhiMAGvec} and offset \nm{\BzeroMAGvec}), and the scale factor and cross coupling matrix \nm{\MMAG}, which also includes the soft iron magnetism. The process can be repeated several times to isolate the influence of hard iron magnetism (a fixed effect that does not change) from the offset, which is a run-to-run error source that changes every time the magnetometers are turned on.
\begin{center}
\begin{tabular}{lcc}
	\hline
	\textbf{Estimation} & \textbf{\#} & \textbf{Coefficients} \\
	\hline
	\nm{\MMAGest}			& 9 & \nm{\lrb{\sMAGXiest, \ \mMAGXijest}}	\\ 
	\nm{\BhiMAGvecest}		& 3 & \nm{\BhiMAGXiest} \\
	\nm{\BzeroMAGvecest}	& 3 & \nm{\BzeroMAG \ \NuzeroMAGiest} \\
	\hline
\end{tabular}
\end{center}
\captionof{table}{Results of swinging process}\label{tab:PreFlight_NonInertial_swinging}

The fifteen coefficients estimated in the swinging process are listed in table \ref{tab:PreFlight_NonInertial_swinging}. Once the coefficients have been estimated, they can be introduced into the processor so it automatically performs the corrections shown in (\ref{eq:PreFlight_NonInertial_mag_error_calib_correction}):
\neweq{\BBtildetilde = {\MMAGest}^{-1} \ \BBtilde - \BhiMAGvecest - \BzeroMAGvecest}{eq:PreFlight_NonInertial_mag_error_calib_correction}

This articles assumes that the bias offset \nm{\BzeroMAGvec} is exclusively a run-to-run source of error that varies every time the IMU is switched on, so bias offset coefficients obtained by swinging are discarded as they have no relation to the offsets that occur during flight. Modeling the swinging process results implies reducing the hard iron bias \nm{\BhiMAGvec} and scale factor and cross couplings errors \nm{\MMAG} found on the sensors specifications by an arbitrary amount of \nm{90\%}. This reduction is already included in the specifications table of section \ref{subsec:Sensors_NonInertial_Magnetometers}. To summarize, instead of applying (\ref{eq:PreFlight_NonInertial_mag_error_calib_correction}) to the measurements obtained by (\ref{eq:Sensor_NonInertial_mag_error_final}), the simulation directly employs (\ref{eq:Sensor_NonInertial_mag_error_final}) with reduced \nm{\BhiMAGvec} and \nm{\MMAG} taken from the table of section \ref{subsec:Sensors_NonInertial_Magnetometers} .


\subsection{Determination of the Camera Frame Pose}\label{subsec:Camera_frame}

The images generated by the onboard camera, and simulated by means of the \texttt{Earth Viewer} application introduced in section \ref{subsec:Sensors_camera_earth_viewer}, do not only depend on the relative pose between the body and the Earth, but also on that of the camera with respect to the aircraft structure, represented by the rotation \nm{\phiBC} and displacement \nm{\TBCB} generated when mounting the camera as described in section \ref{subsec:Sensors_Camera_Mounting}. Visual navigation algorithms however rely on the navigation system best estimate of this pose, this is, \nm{\phiBCest} and \nm{\TBCBest}, which need to be estimated once the already calibrated camera has been mounted on the aircraft. The two-phase process requires a chess board such as that employed in calibration \cite{Soatto2001, Kaehler2016}. 

The first phase uses an optimization procedure quite similar to that used in calibration to determine the relative pose between the camera frame \nm{\FC} and the one rigidly attached to the chessboard. Instead of using the location of each chess box corner in different images, this process relies on a single photo and imposes that all chess boxes are square and have the same size, which is enough to obtain a solution up to an unknown scale. The size of the chess boxes provides the scale required to unambiguously solve the identification problem with high precision.

The second step is to obtain the pose between the chessboard and body frames. This is a straightforward geometric optimization problem that relies on distance measurements between chessboard points and aircraft structure points whose coordinates in \nm{\FB} frame are known. The resulting accuracy depends on the accuracy with which these distances can be measured, so special equipment may be required given the importance of the final estimations for the success of the visual navigation algorithms.

Overall this is a very robust and accurate process if properly executed, which enables the author to employ the very small errors \nm{\sigmaTBCBest} \nm{\sigmaphiBCest} assigned in table \ref{tab:Sensor_Camera_mounting} for the stochastic estimation of \nm{\TBCBest} and \nm{\phiBCest} in each simulation run by means of (\ref{eq:Sensors_Camera_Mounting_TBCBest}) and (\ref{eq:Sensors_Camera_Mounting_eulerBCest}).

\appendix 

\section{Motion of Multiple Rigid Bodies} \label{cha:Composition}

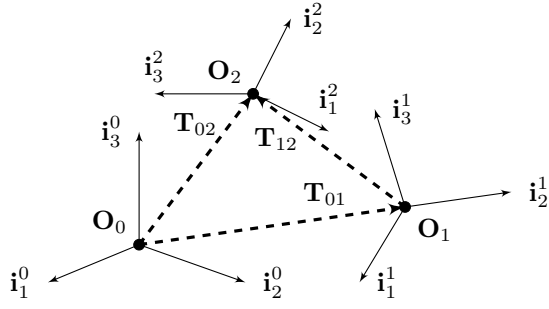
\begin{figure}[h]
\centering
\begin{tikzpicture}[auto, node distance=2cm,>=latex']
	\filldraw [black] (+0.0, +0.0) circle [radius=2pt] node [above left=1pt] {\nm{\vec O_0}};
    \draw [->] (+0.0,+0.0) -- (-1.2,-0.5) node [left=3pt]  {\nm{\vec i_1^0}};
	\draw [->] (+0.0,+0.0) -- (+1.4,-0.5) node [right=3pt] {\nm{\vec i_2^0}};
	\draw [->] (+0.0,+0.0) -- (+0.0,+1.5) node [left=3pt]  {\nm{\vec i_3^0}};
	
	\filldraw [black, xshift=3.5cm, yshift=0.5cm] (+0.0,+0.0) circle [radius=2pt] node [below right=1pt] {\nm{\vec O_1}};
    \draw [xshift=3.5cm, yshift=0.5cm] [->] (+0.0,+0.0) -- (-0.6,-1.0) node [right=3pt] {\nm{\vec i_1^1}};
	\draw [xshift=3.5cm, yshift=0.5cm] [->] (+0.0,+0.0) -- (+1.4,+0.2) node [right=3pt] {\nm{\vec i_2^1}};
	\draw [xshift=3.5cm, yshift=0.5cm] [->] (+0.0,+0.0) -- (-0.4,+1.3) node [right=3pt] {\nm{\vec i_3^1}};
	
	\filldraw [black, xshift=1.5cm, yshift=2.0cm] (+0.0,+0.0) circle [radius=2pt] node [above left=0.5pt] {\nm{\vec O_2}};
    \draw [xshift=1.5cm, yshift=2.0cm] [->] (+0.0,+0.0) -- (+1.0,-0.5) node [above=3pt] {\nm{\vec i_1^2}};
	\draw [xshift=1.5cm, yshift=2.0cm] [->] (+0.0,+0.0) -- (+0.5,+1.0) node [right=1pt] {\nm{\vec i_2^2}};
	\draw [xshift=1.5cm, yshift=2.0cm] [->] (+0.0,+0.0) -- (-1.3,+0.0) node [above=1pt] {\nm{\vec i_3^2}};
	
	\draw [dashed, very thick] [->] (+0.0,+0.0) -- (+3.5,+0.5) node [pos=0.7, above=1pt] {\nm{\Tzeroone}};
	\draw [dashed, very thick] [->] (+0.0,+0.0) -- (+1.5,+2.0) node [pos=0.8, left=1pt]  {\nm{\Tzerotwo}};
	\draw [dashed, very thick] [->] (+3.5,+0.5) -- (+1.5,+2.0) node [pos=0.6, left=2pt]  {\nm{\Tonetwo}};
\end{tikzpicture}
\caption{Reference system for combination of movements}
\label{fig:SO3_ref_systems}
\end{figure}

The equations employed in this article make use of positions, velocities, and accelerations (both linear and angular) that refer to different reference systems or rigid bodies, which are in continuous motion (translation and rotation) among themselves. Their relationships are obtained in this appendix based on the three reference systems shown in figure \ref{fig:SO3_ref_systems}: an inertial reference system \nm{F_0 \{\vec 0_0, \, \vec i_1^0, \, \vec i_2^0, \, \vec i_3^0\}}, and two non inertial reference systems \nm{F_1 \{\vec 0_1, \, \vec i_1^1, \, \vec i_2^1, \, \vec i_3^1\}} and \nm{F_2 \{\vec 0_2, \, \vec i_1^2, \, \vec i_2^2, \, \vec i_3^2\}}, where \nm{\lrp{\Tzeroone, \ \vzeroone, \ \azeroone}} are the position, linear velocity, and linear acceleration of the origin of \nm{F_1} with respect to \nm{F_0}, \nm{\lrp{\Tzerotwo, \ \vzerotwo, \ \azerotwo}} those of the origin of \nm{F_2} with respect to \nm{F_0}, and \nm{\lrp{\Tonetwo, \ \vonetwo, \ \aonetwo}} those of the origin of \nm{F_2} with respect to \nm{F_1}. Similarly, \nm{\lrp{\wzeroone, \ \alphazeroone}} are the angular velocity and angular acceleration of \nm{F_1} with respect to \nm{F_0}, \nm{\lrp{\wzerotwo, \ \alphazerotwo}} those of \nm{F_2} with respect to \nm{F_0}, and \nm{\lrp{\wonetwo, \ \alphaonetwo}} those of \nm{F_2} with respect to \nm{F_1}. Let's also consider that \nm{\Rzeroone}, \nm{\Rzerotwo}, and \nm{\Ronetwo} are the rotation matrices among the three different rigid bodies.


\subsection{Composition of Position}\label{subsec:MOT_Composition_Position}

The relationship between the linear position vectors \nm{\Tzerotwo}, \nm{\Tonetwo}, and \nm{\Tzeroone} can be established by vectorial arithmetics when expressed in the same reference frame, or by coordinate transformation \cite{Shuster1993} when not so:
\neweq{\Tzerotwozero = \Tonetwozero + \Tzeroonezero = \Rzeroone \; \Tonetwoone + \Tzeroonezero}{eq:MOT_Comp_Pos1}


\subsection{Composition of Linear Velocity}\label{subsec:MOT_Composition_LinearVelocity}

The derivation with time of (\ref{eq:MOT_Comp_Pos1}) results in:
\neweq{\Tzerotwozerodot = \Rzeroonedot \; \Tonetwoone +\Rzeroone \; \Tonetwoonedot + \Tzeroonezerodot}{eq:MOT_Comp_LinearVel1}

The use of the relationship between the rotation matrix and its time derivative \cite{Shuster1993} results in\footnote{Note that the wide hat \nm{< \widehat{\cdot} >} refers to the skew-symmetric form of a vector.}:
\neweq{\Tzerotwozerodot =\Rzeroone \; \wzerooneoneskew \; \Tonetwoone +\Rzeroone \; \Tonetwoonedot + \Tzeroonezerodot}{eq:MOT_Comp_LinearVel2}

Reordering and replacing the position time derivatives with their respective velocities results in the relationship between the linear velocity vectors \nm{\vzerotwo}, \nm{\vonetwo}, and \nm{\vzeroone} expressed in the inertial frame \nm{F_0}:
\begin{eqnarray}
\nm{\vzerotwozero} & = & \nm{\Rzeroone \; \vonetwoone + \vzeroonezero +\Rzeroone \; \wzerooneoneskew \; \Tonetwoone} \label {eq:MOT_Comp_LinearVel3} \\
\nm{\vzerotwozero} & = & \nm{\vonetwozero + \vzeroonezero + \wzeroonezeroskew \; \Tonetwozero} \label{eq:MOT_Comp_LinearVel4} 
\end{eqnarray}


\subsection{Composition of Linear Acceleration}\label{subsec:MOT_Composition_LinearAcceleration}

The derivation with time of (\ref{eq:MOT_Comp_LinearVel3}) results in:
\neweq{\vzerotwozerodot = \Rzeroonedot \; \vonetwoone +\Rzeroone \; \vonetwoonedot + \vzeroonezerodot + \Rzeroonedot \; \wzerooneoneskew \; \Tonetwoone +\Rzeroone \; \wzerooneoneskewdot \; \Tonetwoone +\Rzeroone \; \wzerooneoneskew \; \Tonetwoonedot} {eq:MOT_Comp_LinearAcc1}

Replacing the rotation matrix time derivative results in:
\neweq{\vzerotwozerodot =\Rzeroone \; \wzerooneoneskew \; \vonetwoone +\Rzeroone \; \vonetwoonedot + \vzeroonezerodot +\Rzeroone \; \wzerooneoneskew \; \wzerooneoneskew \; \Tonetwoone + \Rzeroone \; \wzerooneoneskewdot \; \Tonetwoone + \Rzeroone \; \wzerooneoneskew \; \Tonetwoonedot} {eq:MOT_Comp_LinearAcc2}

Reordering, replacing the position, linear velocity, and angular velocity time derivatives with their respective velocities, linear accelerations and angular accelerations, results in the relationship between the linear acceleration vectors \nm{\azerotwo}, \nm{\aonetwo}, and \nm{\azeroone} expressed in the inertial frame \nm{F_0}:
\begin{eqnarray}
\nm{\azerotwozero} & = & \nm{\Rzeroone \; \aonetwoone + \lrp{ \azeroonezero + \Rzeroone \; \alphazerooneoneskew \; \Tonetwoone + \Rzeroone \; \wzerooneoneskew \; \wzerooneoneskew \; \Tonetwoone } +  2 \; \Rzeroone \; \wzerooneoneskew \; \vonetwoone} \label{eq:MOT_Comp_LinearAcc3} \\
\nm{\azerotwozero} & = & \nm{\aonetwozero + \lrp{ \azeroonezero + \alphazeroonezeroskew \; \Tonetwozero + \wzeroonezeroskew \; \wzeroonezeroskew \; \Tonetwozero } +  2 \; \wzeroonezeroskew \; \vonetwozero} \label{eq:MOT_Comp_LinearAcc4}
\end{eqnarray}

The term on the left hand side is called absolute acceleration, while the three right hand side terms are usually named relative, transport, and Coriolis accelerations, respectively.


\subsection{Composition of Angular Velocity}\label{subsec:MOT_Composition_AngularVelocity}

The relationship among the different frames angular velocities is given by the rotation matrix composition rule \cite{Shuster1993}, which can be derivated with respect to time:
\neweq{\Rzerotwo = \Rzeroone \, \Ronetwo \ \rightarrow \ \Rzerotwodot = \Rzeroonedot \, \Ronetwo + \Rzeroone \, \Ronetwodot} {eq:MOT_Comp_AngularVel1}

Replacing the rotation matrix time derivatives results in:
\neweq{\Rzerotwo \; \wzerotwotwoskew = \wzeroonezeroskew \; \Rzeroone \; \Ronetwo + \Rzeroone \; \Ronetwo \; \wonetwotwoskew = \wzeroonezeroskew \; \Rzerotwo + \Rzerotwo \; \wonetwotwoskew = \Rzerotwo \; \wzeroonetwoskew + \Rzerotwo \; \wonetwotwoskew} {eq:MOT_Comp_AngularVel2}

The relationship among the angular velocity vectors \nm{\wzerotwo}, \nm{\wonetwo}, and \nm{\wzeroone} is hence the following:
\neweq{\wzerotwozero = \wonetwozero + \wzeroonezero = \Rzeroone \; \wonetwoone + \wzeroonezero}{eq:MOT_Comp_AngularVel3}


\subsection{Composition of Angular Acceleration}\label{subsec:MOT_Composition_AngularAcceleration}

The derivation with time of (\ref{eq:MOT_Comp_AngularVel3}) results in:
\neweq{\wzerotwozerodot = \Rzeroonedot \; \wonetwoone + \Rzeroone \; \wonetwoonedot + \wzeroonezerodot}{eq:MOT_Comp_AngularAcc1}

Replacing the rotation matrix time derivatives results in:
\neweq{\wzerotwozerodot = \Rzeroone \; \wzerooneoneskew \; \wonetwoone + \Rzeroone \; \wonetwoonedot + \wzeroonezerodot}{eq:MOT_Comp_AngularAcc2}

Reordering, replacing the angular velocity time derivatives with their respective angular accelerations, results in the relationship between the angular acceleration vectors \nm{\alphazerotwo}, \nm{\alphaonetwo}, and \nm{\alphazeroone} expressed in the inertial frame \nm{F_0}:
\begin{eqnarray}
\nm{\alphazerotwozero} & = & \nm{\Rzeroone \; \alphaonetwoone + \alphazeroonezero + \Rzeroone \; \wzerooneoneskew \; \wonetwoone} \label {eq:MOT_Comp_AngularAcc3} \\
\nm{\alphazerotwozero} & = & \nm{\alphaonetwozero + \alphazeroonezero + \wzeroonezeroskew \; \wonetwozero} \label {eq:MOT_Comp_AngularAcc4} 
\end{eqnarray}


\subsection{Summary of Compositions}\label{subsec:MOT_Composition_Summary}

The final expressions of the compositions above (\ref{eq:MOT_Comp_Pos1}, \ref{eq:MOT_Comp_LinearVel4}, \ref{eq:MOT_Comp_LinearAcc4}, \ref{eq:MOT_Comp_AngularVel3}, and \ref{eq:MOT_Comp_AngularAcc4}) are all expressed in the inertial frame \nm{F_0}, but they are also valid in any other frame as long as all its components are converted into that frame\footnote{Note that it is not the same to compute a time derivative (velocity, acceleration, or angular acceleration) in the inertial frame and then convert it into a different frame, than to directly compute the derivative in a non inertial frame.}:
\begin{eqnarray}
\nm{\Tzerotwo} & = & \nm{\Tonetwo + \Tzeroone} \label{eq:MOT_Comp_PosFinal} \\
\nm{\vzerotwo} & = & \nm{\vonetwo + \vzeroone + \wzerooneskew \; \Tonetwo} \label{eq:MOT_Comp_LinearVelFinal} \\ 
\nm{\azerotwo} & = & \nm{\aonetwo + \lrp{ \azeroone + \alphazerooneskew \; \Tonetwo + \wzerooneskew \; \wzerooneskew \; \Tonetwo } +  2 \; \wzerooneskew \; \vonetwo} \label{eq:MOT_Comp_LinearAccFinal} \\
\nm{\wzerotwo} & = & \nm{\wonetwo + \wzeroone}\label{eq:MOT_Comp_AngularVelFinal} \\
\nm{\alphazerotwo} & = & \nm{\alphaonetwo + \alphazeroone + \wzerooneskew \; \wonetwo} \label {eq:MOT_Comp_AngularAccFinal}
\end{eqnarray}

\bibliographystyle{ieeetr}   
\bibliography{sensors}

\begin{thebibliography}{10}

\bibitem{Gallo2021_sensors}
E.~Gallo, ``{High Fidelity Model of the Sensors and Camera onboard a Low SWaP
  Fixed Wing UAV}.''
  \url{https://github.com/edugallogithub/sensor_camera_model}, 2021.
\newblock C++ open source code.

\bibitem{SIMULATION}
E.~Gallo, ``{Stochastic High Fidelity Simulation and Scenarios for Testing of
  Fixed Wing Autonomous GNSS-Denied Navigation Algorithms},'' 2021.
\newblock \texttt{arXiv:2102.00883 [cs.RO]},
  \url{https://arxiv.org/abs/2102.00883}.

\bibitem{Farrell2008}
J.~A. Farrell, {\em {Aided Navigation, GPS with High Rate Sensors}}.
\newblock McGraw-Hill, Electronic Engineering Series, 2008.

\bibitem{Etkin1972}
B.~Etkin, {\em {Dynamics of Atmospheric Flight}}.
\newblock {John Wiley \& Sons}, 1972.

\bibitem{Groves2008}
P.~D. Groves, {\em {Principles of GNSS, Inertial, and Multisensor Integrated
  Navigation Systems}}.
\newblock Artech House, GNSS Technology and Application Series, 2008.

\bibitem{Hibbeler2015}
R.~C. Hibbeler, {\em {Engineering Mechanics: Statics and Dynamics}}.
\newblock Pearson, \fourteenth~ed., 2015.

\bibitem{Titterton2004}
D.~Titterton and J.~Weston, {\em Strapdown Inertial Navigation Technology}.
\newblock Radar, Sonar and Navigation Series, The Institution of Engineering
  and Technology, \second~ed., 2004.

\bibitem{Grewal2010}
M.~Grewal and A.~Andrews, ``{How Good Is Your Gyro?},'' tech. rep., {IEEE
  Control Systems Magazine}, 2010.

\bibitem{Chatfield1997}
A.~B. Chatfield, {\em {Fundamentals of High Accuracy Inertial Navigation}}.
\newblock American Institute of Aeronautics and Astronautics, Progress in
  Astronautics and Aeronautics, 1997.

\bibitem{Trusov2011}
A.~A. Trusov, ``{Allan Variance Analysis of Random Noise Modes in
  Gyroscopes},'' tech. rep., University of California, 2011.

\bibitem{KVH2014}
``{Guide to Comparing Gyro and IMU Technologies: Micro-Electro-Mechanical
  Systems and Fiber Optic Gyros},'' 2014.
\newblock KVH Fiber Optic Gyro.

\bibitem{Chow2011}
R.~Chow, ``{Evaluating Inertial Measurement Units},'' tech. rep., Epson
  Electronics America, 2011.

\bibitem{IEEE1998}
``{IEEE Standard Specification Format Guide and Test Procedure for Single-Axis
  Interferometric Fiber Optic Gyros},'' 1998.

\bibitem{Crassidis2006}
J.~L. Crassidis, ``{Sigma-Point Kalman Filtering for Integrated GPS and
  Inertial Navigation},'' {\em {IEEE Transactions on Aerospace and Electronic
  Systems}}, 2006.

\bibitem{Woodman2007}
O.~J. Woodman, ``{An Introduction to Inertial Navigation},'' tech. rep.,
  University of Cambridge Computer Laboratory, 2007.

\bibitem{Stockwell}
W.~Stockwell, ``{Angle Random Walk},'' tech. rep., Crossbow Technology Inc.

\bibitem{Renaut2013}
F.~Renaut, ``{MEMS Inertial Sensors Technology},'' tech. rep., Swiss Federal
  Institude of Technology, 2013.

\bibitem{Farrenkopf1974}
R.~L. Farrenkopf, ``{Analytic Steady-State Accuracy Solutions for Two Common
  Spacecraft Attitude Estimators},'' {\em Guidance and Control}, 1974.

\bibitem{Rogers2007}
R.~M. Rogers, {\em {Applied Mathematics in Integrated Navigation Systems}}.
\newblock AIAA Education Series, American Institute of Aeronautics and
  Astronautics, \third~ed., 2007.

\bibitem{Frishman1971}
F.~Frishman, ``{On the Arithmetic Means and Variances of Products and Ratios of
  Random Variables},'' tech. rep., {Army Research Office}, 1971.

\bibitem{Shuster1993}
M.~D. Shuster, ``{A Survey of Attitude Representations},'' {\em Journal of the
  Astronautical Sciences}, 1993.

\bibitem{ADIS16488A}
``{Analog Devices ADIS16488A Ten Degrees of Freedom Inertial Sensor Data
  Sheet},'' 2015.
\newblock
  \url{http://www.analog.com/media/en/technical-documentation/data-sheets/ADIS16488A.pdf}.

\bibitem{Kayton1997}
M.~Kayton and W.~R. Fried, {\em {Avionics Navigation Systems}}.
\newblock John Wiley \& Sons, \second~ed., 1997.

\bibitem{Pinter1990}
C.~C. Pinter, {\em {A Book of Abstract Algebra}}.
\newblock Dover Publications, \second~ed., 1990.

\bibitem{NEOM8}
``{U-Blox NEO-M8 Data Sheet}.''
\newblock
  \url{https://www.u-blox.com/sites/default/files/NEO-M8-FW3_DataSheet_%28UBX-15031086%29.pdf}.

\bibitem{NOVATELGPS}
``{GPS Position Accuracy Measures},'' 2003.
\newblock \url{http://www.gisresources.com/wp-content/uploads/2014/03/}
  \texttt{gps\_book.pdf}.

\bibitem{ISA}
``{Manual of the ICAO International Standard Atmosphere},'' tech. rep.,
  International Civil Aviation Organization, 2000.
\newblock ICAO DOC-7488/3, \third edition.

\bibitem{INSA}
E.~Gallo, ``{Quasi Static Atmospheric Model for Aircraft Trajectory Prediction
  and Flight Simulation},'' 2021.
\newblock \texttt{arXiv:2101.10744v1 [eess.SY]},
  \url{https://arxiv.org/abs/2101.10744}.

\bibitem{Aeroprobe}
``{Aeroprobe Corporation Air Data Systems Data Sheet}.''
\newblock \url{www.aeroprobe.com}.

\bibitem{ADT7420}
``{Analog Devices ADT7420 Temperature Sensor},'' 2012.
\newblock
  \url{http://www.analog.com/media/en/technical-documentation/data-sheets/ADT7420.pdf}.

\bibitem{Eshelby2000}
M.~E. Eshelby, {\em {Aircraft Performance, Theory and Practice}}.
\newblock Arnold, 2000.

\bibitem{Batchelor2000}
G.~Batchelor, {\em {An Introduction to Fluid Dynamics}}.
\newblock {Cambridge Mathematical Library}, 2000.

\bibitem{Oates1989}
G.~C. Oates, {\em {Aircraft Propulsion Systems Technology and Design}}.
\newblock AIAA Education Series, American Institute of Aeronautics and
  Astronautics, 1989.

\bibitem{Soatto2001}
Y.~Ma, S.~Soatto, J.~Kosecka, and S.~S. Sastry, {\em {An Invitation to 3-D
  Vision, From Images to Geometric Models}}.
\newblock Imaging, Vision, and Graphics, Springer, 2001.

\bibitem{OpenSceneGraph}
{\em {Open Scene Graph}}.
\newblock \url{http://openscenegraph.org}.

\bibitem{osgEarth}
{\em {osgEarth}}.
\newblock \url{http://osgearth.org}.

\bibitem{Tidaldi2014}
D.~Tedaldi, A.~Pretto, and E.~Menegatti, ``{A Robust and Easy to Implement
  Method for IMU Calibration without External Equipments Required},'' in {\em
  {IEEE International Conference on Robotics and Automation}}, IEEE, 2014.

\bibitem{Kaehler2016}
A.~Kaehler and G.~Bradski, {\em Learning OpenCV 3: Computer Vision in C++ with
  the OpenCV Library}.
\newblock O'Reilly Media, Inc, 2016.

\end{thebibliography}

\end{document}